\documentclass[ejs]{imsart}

\RequirePackage[OT1]{fontenc}
\RequirePackage{amsthm,amsmath}
\RequirePackage[numbers]{natbib}
\RequirePackage[colorlinks,citecolor=blue,urlcolor=blue]{hyperref}

\usepackage{amssymb}
\usepackage{bm}
\usepackage{graphicx,subfigure}
\usepackage[graphicx]{realboxes}
\usepackage{xcolor}
\usepackage{booktabs}
\usepackage{subfigure}
\usepackage{multirow}
\usepackage{cancel}
\usepackage{mathtools}
\usepackage{adjustbox}
\usepackage{cleveref}

\def\Supp{\textsc{Supp}}
\newcommand{\bu}{\bm{u}}
\newcommand{\bv}{\bm{v}}
\newcommand{\bw}{\bm{w}}
\newcommand{\ba}{\bm{a}}
\newcommand{\bW}{\bm{W}}

\DeclareMathOperator*{\argmin}{\arg\!\min}
\newtheorem{assumption}{Assumption}
\newtheorem{corollary}{Corollary}
\newtheorem{example}{Example}
\newtheorem{lemma}{Lemma}
\newtheorem{proposition}{Proposition}
\newtheorem{remark}{Remark}
\crefname{assumption}{assumption}{assumptions}

\startlocaldefs
\numberwithin{equation}{section}
\theoremstyle{plain}
\newtheorem{theorem}{Theorem}
\endlocaldefs

\begin{document}

\begin{frontmatter}
\title{Statistical Learning\\from Biased Training Samples}
\runtitle{Statistical Learning from Biased Training Samples}

\begin{aug}
\author{\fnms{Stephan} \snm{Cl\'{e}men\c{c}on}\ead[label=e1]{stephan.clemencon@telecom-paris.fr}}
\address{LTCI, T\'{e}l\'{e}com Paris, Institut Polytechnique de Paris, France\\
\printead{e1}}

\author{\fnms{Pierre} \snm{Laforgue}%\thanksref{t1}
\ead[label=e2]{pierre.laforgue@unimi.it}}
\address{Uinversit\`{a} degli Studi di Milano, Milan, Italy\\
\printead{e2}}

%\thankstext{t1}{Work done when Pierre Laforgue was in PhD at T\'{e}l\'{e}com Paris}

\runauthor{S. Cl\'{e}men\c{c}on and P. Laforgue}
\end{aug}

\begin{abstract}
With the deluge of digitized information in the Big Data era, massive datasets are becoming increasingly available for learning \mbox{predictive} models.
However, in many practical situations, the poor control of the data acquisition processes may naturally jeopardize the outputs of \mbox{machine} learning algorithms, and selection bias issues are now the subject of much attention in the literature.
The present article investigates how to extend Empirical Risk Minimization, the principal paradigm in statistical \mbox{learning}, when training observations are generated from biased models, i.e., from distributions that are different from that in the test/prediction stage, and absolutely continuous with respect to the latter.
Precisely, we show how to build a ``nearly debiased'' training statistical population from biased samples and the related biasing functions, following in the footsteps of the approach originally proposed in \cite{Vardi85}.
Furthermore, we study from a nonasymptotic perspective the performance of minimizers of an empirical version of the risk computed from the statistical population thus created.
Remarkably, the learning rate achieved by this procedure is of the same order as that attained in absence of selection bias.
Beyond the theoretical guarantees, we also present experimental results supporting the relevance of the algorithmic approach promoted in this paper.
\end{abstract}



\tableofcontents

\end{frontmatter}

\section{Introduction}
In the standard setting of binary classification, the flagship problem in \mbox{statistical} learning, $Z=(X,Y)$ is a random pair defined on a probability space with \mbox{unknown} probability distribution $P$.
The random vector $X$, valued in $\mathcal{X}\subset \mathbb{R}^d$, models some information supposedly useful to predict the random binary label $Y$, taking its values in $\{-1,+1\}$.
The objective is to build a Borelian predictive function, i.e., a classifier, $g:\mathcal{X}\rightarrow \{-1,+1\}$ that minimizes the error \mbox{probability}, i.e., the risk, of the decision: $L_P(g)=\mathbb{P}\{Y\neq g(X)\}$.
It is well-known that the optimal solution is given by the Bayes classifier $g^*(x)=2\,\mathbb{I}\{\eta(x)\geq 1/2\}-1$, where $\eta(X)=\mathbb{P}\{Y=1 \mid X\}$ denotes the posterior probability, with \mbox{minimum} risk $L_P^{*}=\mathbb{E}[\min\{\eta(X),\; 1-\eta(X)\}]$.
In practice however, $P$ (and consequently $\eta$) is usually unknown, and one generally resorts to a training dataset $\mathcal{D}_n=\{(X_1,Y_1),\; \ldots,\; (X_n,Y_n)\}$, composed of $n\geq 1$ independent copies of $(X,Y)$.
Empirical Risk Minimization (ERM in short, see e.g., \cite{DGL96}) consists in solving the minimization problem $\min_{g\in \mathcal{G}}\widehat{L}_n(g)$, where $\widehat{L}_n(g)$ is a statistical estimator of the risk $L_P(g)$, generally obtained by replacing $P$ in $L_P$ with the empirical distribution of the $(X_i,Y_i)$'s, and $\mathcal{G}$ is a class of predictive rules hopefully rich enough to contain an accurate approximant of $g^*$.
In this case, the empirical risk is the statistical average $\widehat{L}_n(g)=(1/n)\sum_{i=1}^n\mathbb{I}\{ Y_i\neq g(X_i)\}$, denoting by $\mathbb{I}\{\mathcal{E}\}$ the indicator function of any event $\mathcal{E}$.
Under various assumptions controlling the complexity of the class $\mathcal{G}$ over which the learning task is achieved (e.g., finite {\sc VC} dimension, metric entropy, or Rademacher complexity), the performance of empirical risk \mbox{minimizers} (i.e., solutions to the ERM problem), measured through the excess of risk $g\mapsto L_P(g)-L_P^*$, can be classically studied by means of concentration inequalities for empirical processes, see e.g., \cite{Boucheron2005}.
Although very informative in i.i.d. settings, these generalization results nonetheless crucially rely upon the assumption that training observations are sampled from the true test distribution, which is often violated in practice.
Motivated by the poor control of the data acquisition process in many applications (see e.g., \cite{vM16}), the purpose of the present article is to investigate ERM in the presence of sample \mbox{selection} bias, that is to say in the situation where the samples at disposal for learning a predictive rule $g$ are not distributed as $P$, which can be viewed as a very specific case of Transfer Learning, see \cite{BenDavid10}.
As recently highlighted by \cite{BCZSK16}, \cite{ZWYOC17} or \cite{BHSDR19} among others, representativeness issues do not vanish simply under the effect of the size of the training set.
Hence, ignoring selection bias issues may dramatically jeopardize the accuracy of the outputs of machine learning algorithms.
The method we propose stands out from previous approaches for two main reasons: (1) it encompasses a wide range of biasing scenarios, and (2) it applies to biased training distributions that may not dominate the test distribution.

Selection bias can be due to a wide variety of causes (e.g., the use of a survey scheme to collect observations, censorship, truncation, see for instance \cite{heckman1990varieties} or \cite{vella1998estimating}), and the study of its impact on inference methods, as well as techniques to remedy it, have a very long history in statistics.
Depending on the nature of the mechanism causing the sample selection bias, and on that of the statistical information available to the learner, particular cases have been considered in the machine learning literature, for which specific approaches have been developed.
For instance, the case where some errors occur among the labels of the training data is studied in \cite{Lugosi92}, while in \cite{Papa16} ERM is extended to the framework of survey training data (when inclusion probabilities are known).
In \cite{van_belle_learning_2011} and \cite{ACP19}, authors consider statistical learning of regression models in the context of right censored training observations.
Recently, a very special case of sample selection bias, referred to as \textit{covariate shift}, has been the subject of a good deal of attention (though it had been already considered by \cite{manski1977estimation} in a simplified version).
In this case, addressing the sample selection bias issue is made much easier by the hypothesis stipulating that, in supervised problems, only the marginal input distribution may possibly change, the conditional distribution of the output $Y$ given the input observation $X$ being the same in the learning and predictive stages.
One may refer to \cite{shimodaira2000improving}, \cite{sugiyama2005input} and \cite{huang2007correcting}, or to the monographs  \cite{quionero2009dataset} and \cite{sugiyama2012machine}.
In contrast to the aforementioned settings, the procedure exposed here allows for much more complex biasing mechanisms.
Specifically, survey schemes and censorship scenarios can be seen as particular instances of our framework, see \Cref{ex:classif,ex:regression}.
Moreover, bias may apply to covariates, labels, or both at the same time without altering the guarantees.
We emphasize that despite focus has been put on supervised learning for the sake of clarity, the presented debiasing approach remains valid for unsupervised tasks, as long as they build upon ERM, see \Cref{ex:clustering,ex:clustering_2}.

Methods dedicated to correct sample bias usually boil down to reweighting the training observations with appropriate weights, based on the Importance Sampling approach, or according to the Inverse Probability Weighting technique (IPW in abbreviated form, see e.g., \cite{dubin1989selection} or \cite{winship1992models} in the context of linear regression models), rather than using uniform weights.
For instance, these weights are the inverses of the first order inclusion probabilities in the case where data are acquired by means of a survey plan, see \cite{Clem_Bert2017} and \cite{Papa16}, or the inverses of estimates of the probability of not being censored when data suffer from random censorship, see \cite{ACP19} and the references therein.
Side information about the cause of the selection may also be used to derive explicit forms for the appropriate weights, see e.g., \cite{zadrozny2004learning}, \cite{rosset2005method} in a semi-supervised framework, \cite{dudik2006correcting} in the context of maximum entropy density estimation, or \cite{lin2002support} for the adaptation of the SVM algorithm to certain selection bias situations.
More generally, if the Radon-Nikodym derivative of the test distribution $P$ w.r.t. the training distribution $Q$ (supposedly dominating $P$) is known, one may simply reweight each training observation $z$ by $(dP/dQ)(z)$ in order to get an unbiased estimate of the true risk.
However, this method may be inapplicable, as soon as $P$ is not absolutely continuous w.r.t. $Q$.
To bypass this limitation, several techniques have been developed, based for instance on the discrepancy distance between $P$ and $Q$, see \cite{mansour2009domain} and references therein, or on their R\'{e}nyi divergence, see e.g., \cite{cortes2010learning}.
We point out that statistical learning based on biased samples can be viewed as a very specific case of \textit{transfer learning}, see \cite{storkey2009training}, but also e.g., \cite{Survey1} and \cite{Weiss2016}.
Several recent works in this area also provide theoretical analyses for particular machine learning tasks without requiring the absolute continuity condition, at the cost of additional restrictive assumptions however.
Hence, a no-free-lunch theorem for multitask learning is established in \cite{hanneke2020no}, as well as a method to aggregate the datasets if the task distributions have small discrepancies with respect to the target distribution, see the transfer exponent condition therein.
In \cite{chua2021fine,tripuraneni2021provable} the authors assume that the tasks share an (approximately) common data representation, while \cite{cai2021transfer} analyzes the specific posterior drift model, i.e., it is assumed that the distributions of the covariates remain the same.
Finally, \cite{reeve2021adaptive} studies transfer learning for binary classification under several assumptions on the transfer mechanism, the marginal distributions and their smoothness.
We highlight that none of these assumptions is made in the present paper.

The perspective embraced in the present paper is quite different.
We consider multiple biased training distributions, none of them being assumed to dominate $P$. In particular, the variance of the Radon-Nikodym derivatives $(dP/dQ)(Z)$ are not supposed to be bounded, in contrast to \cite{cortes2010learning}.
Instead, we leverage training samples drawn from these biased distributions and show how to combine them in order to construct an unbiased estimate of the target distribution $P$, under mild identifiability hypotheses.
The debiasing weights are defined as solutions to a nontrivial system of equations, and do not enjoy  any simple closed-form \mbox{expressions} in general, in contrast to those used in the context of survey schemes or censorship models.
Precisely, we focus on the case where statistical learning is based on training data sampled from \textit{biased sampling models}, as \mbox{originally} introduced in \cite{Vardi85} in the context of asymptotic nonparametric estimation of \mbox{cumulative} distribution functions, see also \cite{GVW88}.
This very general selection bias framework accounts for many situations encountered in practice, covering for \mbox{instance} the (far from uncommon) situation where the samples available to learn a binary classifier $g(x)$ are sampled from conditional distributions of $(X,Y)$ given that $X$ lies in specific subsets of the input space $\mathcal{X}$ (assuming that the union of these subsets is equal to $X$'s support).
In this setting, we extend ERM to the case of biased training data with nonasymptotic guarantees about their generalization ability.
We propose to build an unbiased empirical estimator of the test distribution $P$ by solving a generally nontrivial system of equations, which we use to compute a ``nearly unbiased'' risk estimate.
We then establish a tail probability bound for the maximal deviations between the risk functional and the estimate thus constructed.
Based on this result, we finally prove that minimizers of the ``debiased empirical risk'' achieve learning rate bounds that are of the same order as those attained by empirical risk minimizers in absence of any bias mechanism.
If our approach builds on the distribution estimation procedure for biased sampling models introduced in \cite{GVW88}, note that the latter work is restricted to the asymptotic study of cumulative distribution functions.
In contrast, we provide the first ---to the best of our knowledge--- nonasymptotic guarantees for this approach, in a much more general framework.
This allows us to devise an extension of the ERM paradigm to biased training datasets with provable finite-sample  guarantees, as required in the statistical learning literature.
%
%We highlight that nonasymptotic bounds are crucial in machine learning, where models are learned on the basis of datasets with a finite number of observations.
%
For the sake of completeness, the notion of biased sampling model is recalled at length in \Cref{subsec:biased_models} and the slightly stronger assumptions needed to carry out a nonasymptotic analysis are detailed and discussed in \Cref{subsec:learning_bias}.
We also present results from various numerical experiments, based on synthetic and real data, that provide strong empirical evidence of the relevance of the approach we propose.
If the fact that knowledge of the biasing functions is required can be seen at first glance as a limitation of the framework developed, one should have in mind that absolutely no learning strategy with statistical guarantees can be designed in absence of any understanding of the biasing mechanism at work.
Moreover, it is  actually far from uncommon in practice that the latter is known (e.g., one may know the types of images that are more easily collected, or the profiles of individuals who most likely answer a questionnaire).
Yet, the situation where the biasing mechanism is only approximately known is of considerable interest in practice, and investigating to which extent the statistical guarantees established in this paper are preserved will be the subject of future research.

The rest of the article is structured as follows.
In \Cref{sec:background}, basics on biased sampling models are briefly recalled, and the framework for statistical learning based on biased training samples is described at length, as well as the algorithmic approach extending the ERM methodology to this setting.
In \Cref{sec:main}, the main theoretical results of this paper, guaranteeing the generalization capacity of ERM under selection bias, are stated.
Illustrative experiments are displayed in \Cref{sec:num}, while technical details are deferred to the Appendix section.

\section{Background and Preliminaries}
\label{sec:background}
We first recall in \Cref{subsec:biased_models} the \emph{biased sampling models} framework developed in \cite{Vardi85} and \cite{GVW88} for asymptotic estimation of cumulative distribution functions.
Next, we present in \Cref{subsec:learning_bias} our approach to generalize ERM to the case where training data samples are drawn from such models.
Here and throughout, we denote by $\delta_a$ the Dirac mass at any point $a$, by $\| U\|_\mathrm{sup}$ the essential supremum of any real-valued random variable (r.v.) $U$, and by $\Supp(P)$ the support of any probability distribution $P$.
Vectors are denoted by bold characters, e.g., $\bv \in \mathbb{R}^K = (v_1, \ldots, v_K)$ for $K \in \mathbb{N}$.
The Euclidean and sup norms are denoted by $\|\cdot\|_2$ and $\|\cdot\|_\infty$, such that $\|\bv\|_2^2 = \sum_{k=1}^K v_k^2$, and $\|\bv\|_\infty = \max_{k \le K} |v_k|$.

%%%%%%%%%%%%%%%%%%%%%%%%%%%%
%                          %
%  BIASED SAMPLING MODELS  %
%                          %
%%%%%%%%%%%%%%%%%%%%%%%%%%%%

\subsection{Biased Sampling Models - The Statistical Framework}
\label{subsec:biased_models}

Let $Z$ be a random vector, taking its values in $\mathcal{Z}\subset\mathbb{R}^q$, where $q \in \mathbb{N}$, with unknown probability distribution $P$.
If independent copies $Z_1,\; \ldots,\; Z_n$ of $Z$ were at disposal, a natural estimator of $P$ would be the raw empirical distribution $(1/n)\sum_{i=1}^n \delta_{Z_i}$.
In \textit{biased sampling models}, as defined in \cite{Vardi85}, one cannot rely on such observations.
Instead, statistical inference must be based on $K\geq 1$ independent biased i.i.d. samples $\mathcal{D}_k=\{Z_{k,1},\; \ldots,\;  Z_{k,n_k} \}$, of size $n_k\geq 1$.
We denote by $n= \sum_{k=1}^K n_k$ the size of the pooled sample, and by $\hat{\lambda}_k = n_k /n$ the proportion of each sample among the total population.
For $k \le K$, the distribution $P_k$ of the $Z_{k,i}$ is assumed to be absolutely continuous w.r.t. the test distribution $P$, and related to it through a known nonnegative biasing function $\omega_k$ such that
\begin{equation}\label{eq:lik_ratio}
\forall k \le K,~\forall z\in \mathcal{Z}, \qquad \frac{dP_k}{dP}(z)=\frac{\omega_k(z)}{\Omega_k}\,,
\end{equation}
where $\Omega_k=\mathbb{E}_P[\omega_k(Z)]=\int\omega_k(z)dP(z)$.
We emphasize that, just like $P$, the $\Omega_k$'s are unknown.
Note that in the case of interest where $\omega_k(Z) = \mathbb{I}\{Z \in \mathcal{Z}_k\}$ for $\mathcal{Z}_k\subset \mathcal{Z}$, see \Cref{ex:classif} below for instance, it is much easier to know ---or guess--- the biasing functions $\omega_k$ (or equivalently the subsets $\mathcal{Z}_k$ in which the observations lie), rather than having access to the $\Omega_k$'s.
Estimating the $\Omega_k$'s is incidentally at the core of our debiasing procedure, see e.g., \Cref{prop:dev_omega}.
We further emphasize that, unlike the $\Omega_k$'s, knowing the biasing functions $\omega_k$ does not provide any information about the target distribution $P$.
In particular, knowing a stratum $\mathcal{Z}_k$ which the observations belong to does not imply in any way that one has access to the conditional distribution $P_k=P(\cdot \mid Z\in \mathcal{Z}_k)$.

The statistical framework defined by Equations \eqref{eq:lik_ratio} has been considered in \cite{GVW88} for nonparametric estimation of a univariate cumulative distribution function (cdf).
Under mild assumptions, it is shown therein that a consistent and asymptotically normal estimator of $P$ can be constructed from the biased samples $\mathcal{D}_k$ and the knowledge of the biasing functions $\omega_k$, for $k \le K$.
The first fundamental assumption, referred to as Assumption $S$ in \cite{GVW88}, guarantees identifiability.
It can be formulated as follows.

\begin{assumption}\label{hyp:ident}
The union of the supports of the biased distributions $P_k$ is equal to the support of distribution $P$:
\begin{equation*}
\bigcup_{k=1}^K\Big\{z\in \mathcal{Z}:\; \omega_k(z)>0  \Big\} = \Supp(P)\,.
\end{equation*}
\end{assumption}

Of course, we have by definition $\bigcup_{k=1}^K \Supp(P_k) \subset \Supp(P)$.
If this inclusion is strict, some parts of $\Supp(P)$ shall never be covered by observations sampled from the $P_k$.
As may be the case, one may only hope to estimate the restriction of $P$ to $\bigcup_{k=1}^K \Supp(P_k)$, and estimation on the entire support is impossible in absence of prior knowledge.
From now on, Assumption~\ref{hyp:ident} is thus supposed to be satisfied.
One should pay attention to the fact that Assumption~\ref{hyp:ident} does not require that the support of a single biased distribution $P_k$ entirely covers that of the target distribution $P$.
In particular, each likelihood ratio~\eqref{eq:lik_ratio} may vanish on a certain measurable subset  weighted by $P$ here, i.e., for all $k \le K$, one may have: $\mathbb{P}\{\omega_k(Z)=0\}>0$.
As discussed in the introduction, this significantly differs from the biased learning framework developed in other works, see \cite{ICMA2020} and the references therein, where the biased distributions are generally assumed to dominate the test distribution as in the usual Importance Sampling setting.
Note also that \Cref{hyp:ident} prevents biasing functions to vanish all at the same time.
This condition is key to invert the likelihood ratio \eqref{eq:invert} and be able to recover the distribution $P$ statistically based on samples drawn from the $P_k$'s.
When $K=1$, this means $\omega_1(Z) > 0$ and IPW debiasing is then immediate of course: one simply weights each observation by means of $1/\omega_1$.
In this work, focus is naturally on situations where $K \ge 2$. As shall be seen, the difficulty caused by the possibly vanishing biasing functions can be bypassed by combining appropriately the biased datasets, so as to compute nearly debiasing weights through the resolution of a system of equations, see \eqref{sys:emp_omega}.
The generic setting described by \Cref{hyp:ident} encompasses many estimation/learning problems, ranging from stratified sampling to censorship and clustering, see Examples \ref{ex:classif}, \ref{ex:regression}, and \ref{ex:clustering}.

The second assumption required is standard in a multi-sample setting.
It stipulates that the sample sizes $n_k$ all tend to infinity as $n\rightarrow \infty$, in a way such that the fractions $\hat{\lambda}_k$ converge towards fixed values $\lambda_k>0$.

\begin{assumption}\label{hyp:lambda_cvg}
There exist $(\lambda_1, \ldots, \lambda_K) \in (0, 1)^K$ satisfying $\sum_{k=1}^K \lambda_k = 1$ such that for all $k \le K$ it holds $\hat{\lambda}_k \rightarrow \lambda_k$ as $n\rightarrow +\infty$.
\end{assumption}

Ignoring the bias selection issue, one may compute the empirical distribution based on the pooled sample
\begin{equation*}%\label{eq:raw_emp}
\widehat{P}_n=\frac{1}{n}\sum_{k=1}^K\sum_{i=1}^{n_k}\delta_{Z_{k,i}}=\sum_{k=1}^K \hat{\lambda}_k \widehat{P}_k\,,
\end{equation*}
where $\widehat{P}_k=(1/n_k)\sum_{i\leq n_k}\delta_{Z_{k,i}}$ is the raw empirical distribution based on the (biased) sample $\mathcal{D}_k$, for $k \le K$.
This discrete random measure is a natural estimator of  the linear convex combination of the $P_k$ given by $\bar{P}=\sum_k\lambda_k P_k$.
Since $\bar{P}$ is different from $P$ in general, it is then easy to see why minimizing the raw empirical risk over the pooled sample may lead to decision rules that generalize poorly.
However, observe that $\bar{P}$ is absolutely continuous w.r.t. $P$, with likelihood ratio
\begin{equation*}
\frac{d\bar{P}}{dP}(z)=\sum_{k=1}^K \frac{\lambda_k\omega_k(z)}{\Omega_k}\,.
\end{equation*}
Under Assumption \ref{hyp:ident}, the latter is strictly positive on the whole support of $Z$, and we have:
\begin{equation}\label{eq:invert}
dP(z) = \left(\sum_{k=1}^K \frac{\lambda_k\omega_k(z)}{\Omega_k}\right)^{-1} d\bar{P}(z)\,.
\end{equation}
Hence, if estimates $\widehat{\Omega}_k$ of the unknown expectations $\mathbb{E}_P[\omega_k(Z)]$ were at our disposal, one could immediately form a plug-in estimator of $P$ by replacing $\bar{P}$, the $\Omega_k$ and the $\lambda_k$ in Equation \eqref{eq:invert} with their statistical versions, namely $\widehat{P}_n$, the $\widehat{\Omega}_k$ and the $\hat{\lambda}_k$
\begin{equation}\label{eq:plug_in}
d\widetilde{P}_n(z) = \left( \sum_{k=1}^K \frac{\hat{\lambda}_k \omega_k(z)}{\widehat{\Omega}_k} \right)^{-1} d\widehat{P}_n(z)\,.
\end{equation}
In order to estimate the vector $\bm{\Omega}=(\Omega_1,\; \ldots,\; \Omega_K)$, note that \Cref{eq:invert} immediately implies that $\bm{\Omega}$ is a solution (in $\bm{W} \in \mathbb{R}^K$) to the system of equations
\begin{equation}\label{sys:true_omega}
\mathbf{1}=\big(\Gamma_1(\bm{W}),\; \ldots,\; \Gamma_K(\bm{W})\big)\,,
\end{equation}
where $\mathbf{1}$ means the $K$-dimensional vector with all components equal to $1$, and for any $k \le K$, and all $\bm{W}=(W_1,\; \ldots,\; W_K)\in (\mathbb{R}_+)^{K}$, the notation
\begin{equation}\label{eq:gamma}
\Gamma_k(\bm{W})=\frac{1}{W_k}\int\frac{\omega_k(z)}{\sum_{l=1}^K\frac{\lambda_l \omega_l(z)}{W_l}}d\bar{P}(z)\,.
\end{equation}
A natural way to approximately recover $\bm{\Omega}$ thus consists in solving a statistical version of Equation \eqref{sys:true_omega}, namely
\begin{equation}\label{sys:emp_omega}
\mathbf{1}=\left( \widehat{\Gamma}_1(\bm{W}),\; \ldots,\;  \widehat{\Gamma}_K(\bm{W}) \right)\,,
\end{equation}
where the $\widehat{\Gamma}_l(\bm{W})$ are built by replacing $\lambda_l$ and $\bar{P}$ in Equation \eqref{eq:gamma} with $\hat{\lambda}_l$ and $\widehat{P}_n$ respectively.
It is important to notice that the $\widehat{\Gamma}_k$ (just like the $\Gamma_k$) are homogeneous of degree 0.
Hence, it is only possible to solve Systems \eqref{sys:true_omega} and \eqref{sys:emp_omega} up to a multiplicative factor.
Hopefully, $\bm{\Omega}$ can be recovered from any solution $\bW^*$ to System \eqref{sys:true_omega}.
Indeed, for all $k \le K$ it holds:
\begin{equation}\label{eq:recover_omega}
\Omega_k = \frac{W^*_k}{\int \left(\sum_{l=1}^K \frac{\lambda_l \omega_l(z)}{W^*_l} \right)^{-1} d\bar{P}(z)}\,,
\end{equation}
refer to \Cref{apx:recover_omega} for technical details.
Similarly, for any solution $\widehat{\bW}_n$ to System \eqref{sys:emp_omega} and any $k \le K$, we define:
\begin{equation}\label{eq:recover_omega_emp}
\widehat{\Omega}_{n, k} = \frac{\widehat{W}_{n, k}}{\int \left(\sum_{l=1}^K \frac{\hat{\lambda}_l \omega_l(z)}{\widehat{W}_{n, l}} \right)^{-1} d\widehat{P}_n(z)}\,.
\end{equation}
Plugging estimators \eqref{eq:recover_omega_emp} into \Cref{eq:plug_in}, the debiased estimate $\widetilde{P}_n$ is
\begin{equation}\label{eq:est_dist}
\widetilde{P}_n=\sum_{k=1}^K\sum_{i=1}^{n_k}\left(  \frac{\left( \sum_{l=1}^K\frac{\hat{\lambda}_l \omega_l(Z_{k,i})}{\widehat{W}_{n, l}} \right)^{-1}}{\sum_{m=1}^K\sum_{j=1}^{n_m}\left( \sum_{l'=1}^K\frac{\hat{\lambda}_{l'} \omega_{l'}(Z_{m,j})}{\widehat{W}_{n, l'}} \right)^{-1}} \right)\delta_{Z_{k,i}}\,.
\end{equation}
The next assumption now aims at ensuring that the solution to System \eqref{sys:emp_omega} is asymptotically unique.
The mapping from distribution $P$ to the family of biased distributions $(P_k)_{k \le K}$, is then one-to-one.
It is expressed as a graph connectivity hypothesis, cf. Assumption $C$ in \cite{GVW88}.

\begin{assumption}\label{hyp:connect_gvw}
Let $G$ be the (undirected) graph with vertices in $\{1, \ldots, K\}$, and edges between vertices $k$ and $l$ ($k \ne l$) if and only if
\begin{equation*}
\int \mathbb{I}\{\omega_k(z) > 0 \} \cdot \mathbb{I}\{\omega_l(z) > 0\}dP(z) > 0.
\end{equation*}
The graph $G$ is connected. 
\end{assumption}

In the one-dimensional case ($q=1$), and under \Cref{hyp:ident,hyp:lambda_cvg,hyp:connect_gvw}, the limit behavior (i.e., consistency, asymptotic normality) of the univariate cdf estimator of \Cref{eq:est_dist}, namely $z\in\mathbb{R}\mapsto \widetilde{P}_n(]-\infty,\; z])$, has been investigated in \cite{GVW88}.
It is the purpose of the subsequent analysis to show that this approach can be successfully applied to statistical learning via ERM in the presence of selection bias, by deriving nonasymptotic guarantees under slightly stronger assumptions.
Incidentally, the arguments which this analysis relies upon permit to establish an exponential tail bound for the cdf estimator mentioned above, extending the Dvoretzky-Kiefer-Wolfowitz inequality, and completing the results of \cite{GVW88}, see \Cref{thm:dkw}.
In the next subsection, the notion of biased sampling model is used in order to develop a framework for statistical learning based on biased training examples.
Before showing rigorously in the next subsection how the ideas previously sketched permit to extend the ERM principle to this framework, a few remarks are in order.

\begin{remark}[{\sc Covariate shift}]
Let $Z = (X, Y)$ be a random pair taking its values in $\mathcal{X} \times \mathcal{Y}$ with distribution $P$ and defining a supervised predictive problem, where $X$ models some input information, useful to predict the output r.v. $Y$.
In the very specific so-called \emph{covariate shift} situation, for each sampling distribution $P_k$ involved in the biasing model, the conditional distribution of $Y$ given $X$ is the same and is thus independent from $k$. However, the $X$-marginals are not necessarily the same and can be possibly supported on different subsets $\mathcal{X}_k \subset \mathcal{X}$.
Note that, in the dedicated covariate shift literature,
% overlapping of the $\mathcal{X}_k$, enforced by 
\Cref{hyp:connect_gvw} is not stipulated in general, insofar as solving the predictive problem statistically only requires to recover the conditional distribution. However, this assumption is of course necessary to emulate the whole distribution $P$ and accomplish other tasks, unsupervised for instance, even in such a specific context, see \Cref{ex:clustering}.
%
%This generality allows to cover much more scenarios with biased sampling models, including for instance unsupervised settings, see \Cref{ex:clustering}.}
%
\end{remark}

\begin{remark}[{\sc Truncation, missing values}]
We point out that, because of  \Cref{hyp:connect_gvw}, the biased sampling models analyzed here do not cover the case of truncated observations, nor certain settings of missing variables.
The latter may instead be treated by different methods, such as (multiple) imputation techniques, see e.g., \cite{rubin2004multiple}.
\end{remark}

We now exhibit a simple example supporting the need for a general approach and showing in particular that solving System \eqref{sys:emp_omega} cannot be avoided in general.

\begin{example}[{\sc Multivariate length biased samples}]
\label{ex:length_bias}
The bias sampling model where the probability of sampling an observation is proportional to its length is referred to as \emph{length bias}.
Its use is motivated by various applications, such as estimating the distribution of the number of children with a rare anomaly in families with proneness to engender such children \citep{haldane1938estimation}, or correcting visibility bias during wildlife population estimation from aerial data \citep{cook1974model} for instance.
Refer to e.g., \cite{patil1977weighted} for an overview of its applications.
In the univariate case, it corresponds to $\omega(z) = z$, with $z\in \mathbb{R}_+$.
When the learner can access two samples (one unbiased, one length biased), an approach to recover the nonparametric maximum likelihood estimator of $P$ is proposed in \cite{vardi1982nonparametric}.
Precisely, let $\mathcal{D}_1 = \{Z_{1, 1}, \ldots, Z_{1, n_1}\}$ be an i.i.d. sample drawn from $P$, and $\mathcal{D}_2 = \{Z_{2, 1}, \ldots, Z_{2, n_2}\}$ be an i.i.d. sample drawn from $P_2$, the length biased version of $P$ such that $dP_2(z) = zdP(z)/\int_{\mathbb{R}_+} z dP(z)$.
Let $Z_1 < \ldots < Z_n$ be the observations of the pooled sample $\mathcal{D}_1 \cup \mathcal{D}_2$ sorted in increasing order (we assume that ties cannot occur for simplicity), and let $\xi_i = \mathbb{I}\{Z_i \in \mathcal{D}_1\}$ indicate whether $Z_i$ comes from $\mathcal{D}_1$ or not for $i=1,\; \ldots,\; n$.
It is immediate to see that the cdf $P$ that maximizes
\begin{equation}\label{eq:likelihood}
\prod_{i=1}^n \big(dP(Z_i)\big)^{\xi_i} \left(\frac{Z_i\,dP(Z_i)}{\int_{\mathbb{R}_+} z\,dP(z) }\right)^{1 - \xi_i}
\end{equation}
has positive jumps only at the $Z_i$'s, so that estimating the jumps $dP(Z_i)$ in \eqref{eq:likelihood} is sufficient.
Simple computations show that
\begin{equation}\label{eq:p}
dP(Z_i) = \frac{\hat{\mu}}{n_2 Z_i + n_1 \hat{\mu}}\,, \quad \text{where $\hat{\mu}$ satisfies} \quad \sum_{i=1}^n \frac{Z_i}{n_2 Z_i + n_1 \hat{\mu}} = 1\,.
\end{equation}
Hence, maximizing \eqref{eq:likelihood} requires to solve the equation on the right hand side of \eqref{eq:p}. The approach can be straightforwardly extended to the multivariate case. Assume that the random variables observed are now valued in $\mathbb{R}_+^K$ and consider $K+1$ datasets such that $\mathcal{D}_0$ is composed of i.i.d. realizations drawn from $P$ and $\mathcal{D}_k$ is drawn from $P_k$ such that $dP_k(z) = z^{(k)}dP(z)/\int_{\mathbb{R}^+} z^{(k)}dP(z)$, where $z^{(k)}$ denotes the $k$-th coordinate of $z=(z^{(1)},\; \ldots,\; z^{(K)})$.
In other words, all datasets (except $\mathcal{D}_0$) are length biased according to different dimensions.
Equipped with the notations $\{Z_1, \ldots, Z_n\} = \mathcal{D}_0 \cup \ldots \cup \mathcal{D}_K$ and $\xi_{i, k} = \mathbb{I}\{Z_i \in \mathcal{D}_k\}$ for $i=1,\; \ldots,\; n$ and $k=1,\; \ldots,\; K$, the quantity that must be maximized writes
\[
\prod_{i=1}^n \big(dP(Z_i)\big)^{\xi_{i, 0}} \prod_{k=1}^K\left(\frac{Z_i^{(k)}\,dP(Z_i)}{\int_{\mathbb{R}_+} z^{(k)}\,dP(z) }\right)^{\xi_{i, k}}\,.
\]
As in the scalar case, we have:
$
dP(Z_i) = 1/\left(n_0 + \sum_{k=1}^K n_k Z_i^{(k)}/\hat{\mu}_k\right)
$, where the $\hat{\mu}_k$'s satisfy
\[
\hat{\mu}_l = \sum_{i=1}^n\, \frac{Z_i^{(l)}}{n_0 + \sum_{k=1}^K \frac{n_k Z_i^{(k)}}{\hat{\mu}_k}}\,,\quad 1\leq l \leq K.
\]
Thus, we need to solve a system of $K$ equations in order to recover the $\hat{\mu}_k$'s, which is actually a specific case of  \eqref{sys:emp_omega}.
Hence, except in certain simplistic situations, see e.g., \Cref{sec:simple_example}, solving System~\eqref{sys:emp_omega} cannot be avoided to debias biased samples in general.
Although the approach based on bias sampling models encompasses IPW, we highlight that it is much more general than the latter.
\end{example}

%%%%%%%%%%%%%%%%%%%%%%%%%%%%%%%%%%
%                                %
%  LEARNING FROM BIASED SAMPLES  %
%                                %
%%%%%%%%%%%%%%%%%%%%%%%%%%%%%%%%%%S

\subsection{Learning from Biased Samples - Extending the ERM Approach}
\label{subsec:learning_bias}

Recall that $Z$ is a random vector valued in $\mathcal{Z}\subset\mathbb{R}^q$, $q\geq 1$, with probability distribution $P$.
Let $\Theta$ be a decision space and consider some loss function $\psi:\mathbb{R}^q\times \Theta\rightarrow \mathbb{R}_+$, that is $P$-integrable for any decision rule $\theta\in \Theta$.
The goal pursued here is to solve the risk minimization problem
\begin{equation}\label{pbm:risk_min}
\min_{\theta\in \Theta} \quad L_P(\theta) = \mathbb{E}_P[\psi(Z,\theta)]
\end{equation}
where $L_P$ is called the risk function.
As recalled in introduction, if independent copies $Z_1,\; \ldots,\; Z_n$ of $Z$ are available, the unknown risk $L_P$ is classically replaced with $\widehat{L}_n = L_{\widehat{P}_n}$, where $\widehat{P}_n=(1/n)\sum_{i=1}^n\delta_{Z_i}$ is the empirical distribution.
Here, the training data is composed of $K$ biased samples $\mathcal{D}_k$, as defined in \Cref{subsec:biased_models}.
Note that it is a strict generalization of the standard ERM setting, insofar as the latter can be recovered as the special case $K=1$ and $\omega_1 \equiv 1$.
This general framework encompasses a wide variety of situations encountered in practice, as illustrated by the following examples.

\begin{example}[{\sc Binary classification under stratified sampling}]
\label{ex:classif}
We place ourselves in the context of binary classification, i.e., we have $Z=(X,Y)$, $\mathcal{Z}=\mathcal{X}\times \{-1,\; +1  \}$, $q=d+1$, $\Theta=\mathcal{G}$, and $\psi((X,Y),g)=\mathbb{I}\{Y\neq g(X)  \}$.
Consider $K\geq 1$ subsets $\mathcal{X}_1,\; \ldots,\; \mathcal{X}_K$  of the input space $\mathcal{X}$, such that $\mu(\mathcal{X}_k)>0$ for all $k \le K$, $\mu$ denoting $X$'s marginal distribution.
The case where only labeled examples with input observations in $\mathcal{X}_k$ can be collected to form sample $\mathcal{D}_k$ corresponds to the situation where $\omega_k(Z)=\mathbb{I}\{X\in \mathcal{X}_k  \}$.
In this case, the $P_k$ are the conditional distributions of $Y$ given that $X\in\mathcal{X}_k$.
\end{example}

In \Cref{ex:classif}, selection bias is due to stratified sampling schemes, where observations are sampled in strata of interest (the $\mathcal{X}_k$'s namely).
Note that in this case the $\mathcal{X}_k$'s, and therefore the $\omega_k$'s, are controlled and known by the learner.
This setting also covers many practical situations, where the training dataset is constructed by the aggregation of different sources, and naturally applies to other learning tasks.
In such scenarios, having access to the $\omega_k$ is natural, as the learner may know the conditions in which the data have been collected, e.g., the part of the world in which the photos have been taken, or the profile of a user answering the questionary.

\begin{example}[{\sc Regression under right censorship}]
\label{ex:regression}
Let the distribution-free regression framework where $T$ is a bounded random duration (i.e., a nonnegative r.v. such that $\| T\|_\mathrm{sup}<+\infty$), and $X$ is a random vector valued in $\mathcal{X}\subset\mathbb{R}^d$, defined on the same probability space, and supposedly useful to predict $T$.
The goal is to learn a regression function $h:\mathcal{X}\rightarrow \mathbb{R}$ in a class $\mathcal{H}$ of bounded functions with minimum quadratic risk.
This corresponds to $Z=(X,T)$, $\mathcal{Z}=\mathcal{X}\times \mathbb{R}_+$, $q=d+1$, $\Theta=\mathcal{H}$, and $\psi((X,T), h)=(T-h(X))^2$.
Let $K\geq 1$, and $0<\tau_1<\ldots<\tau_{K-1}<\tau_K=\| T \|_\mathrm{sup}$.
Consider the case where, the samples $\mathcal{D}_k$ are composed of censored observations with a deterministic right censorship, i.e., of copies of $(X, \min\{T,\; \tau_k\})$.
This is equivalent to the case $\omega_k(Z)=\mathbb{I}\{T\leq \tau_k \}$, and the $P_k$ are the conditional distributions of $(X,T)$ given that $T\leq \tau_k$.
\end{example}

\Cref{ex:regression} typically arises in longitudinal experiments (e.g., medical trials, customer behavior evaluations) that must be stopped at some point (by lack of means for instance).
Instead of discarding every observation for which the event of interest has not occurred yet, one may register the time at which the experiment has been stopped (the $\tau_k$) and leverage this information to debias the population.

\begin{example}[{\sc Clustering}]
\label{ex:clustering}
Consider an unsupervised variant of \Cref{ex:classif}, where $Z=X \in \mathbb{R}^q$ has distribution $P$ and $\omega_k(X) = \mathbb{I}\{X \in \mathcal{X}_k\}$ where $\mathcal{X}_k\subset \mathcal{X}$ for $k=1,\; \ldots,\; K$ is the observable strata of the population of interest.
A popular approach to clustering consists in assuming that distribution $P$ is an unknown mixture of $\mathcal{K}\geq 2$ Gaussian distributions with means $\mu_1,\; \ldots,\; \mu_{\mathcal{K}}$ in $\mathbb{R}^q$ and same covariance matrix $\Sigma$, see \textit{e.g.} \cite[Chapter~14]{Hastie}: $P=\sum_{m=1}^{\mathcal{K}}\pi_m\,\mathcal{N}(\mu_m,\; \Sigma)$, with $(\pi_1, \; \ldots,\;  \pi_\mathcal{K}) \in [0, 1]^\mathcal{K}$ \textit{s.t.} $\sum_{m=1}^{\mathcal{K}}\pi_m=1$.
Denoting by $\theta$ the parameter encoding the Gaussian mixture model and $p_{\theta}(x)$ its likelihood, the Expectation-Maximization algorithm (EM algorithm, see e.g., \cite[Section~8.5]{Hastie}) computes the optimal $\theta$ by maximizing the (log-)likelihood over the observed datapoints.
The latter can be seen as minimizing an empirical version of problem \eqref{pbm:risk_min} with $\psi(X, \theta) =-\log p_\theta(X)$, and is therefore another particular case in which our debiasing approach applies.
\end{example}

% Note that using the debiased population is critical in \Cref{ex:clustering}.
% %
% Indeed, in the case of unbalanced datasets $\mathcal{D}_k$, a naive approach would simply overfit on the regions from which are issued the datasets with the greatest number of observations.
% %
% Instead, leveraging the information furnished by the $\omega_k$ allows to appropriately reweight the data to better reflect the clusters induced by $P$.

As illustrated by Examples \ref{ex:classif}, \ref{ex:regression}, and \ref{ex:clustering},  the framework developed in this paper applies to a wide variety of statistical learning problems, indifferently supervised or unsupervised, sampling bias being determined by the covariates and/or the output.
We also point out that the vast majority of statistical techniques for correcting sampling/selection bias relies on some Inverse Probability Weighting \mbox{approaches}, e.g., Beran, Kaplan-Meier methods, Horvitz-Thompson techniques, propensity score matching.
The sole difference in these variations is the form of the biasing functions and their arguments.
The main advantage of the general framework developed here consists in encompassing all these situations, diverse in appearance only.
It allows to derive generalization guarantees for any risk minimization problem in the presence of selection bias.
The price to pay for our unifying framework is that the debiasing weights cannot be computed trivially but requires the solving of System \eqref{sys:emp_omega}.
We recall that \Cref{ex:length_bias} shows that this step is unavoidable in general.
In this context, we prove in \Cref{sec:main} that, under mild assumptions, minimizing $L_{\widetilde{P}_n}$, where $\widetilde{P}_n$ is defined in \eqref{eq:est_dist}, allows to attain learning rates that are of the same order, $O_{\mathbb{P}}(1/\sqrt{n})$ namely, as those achieved in absence of any selection bias, i.e., when $\omega_k \equiv 1$ for all $k \le K$.
The minimization of the functional $L_{\widetilde{P}_n}$ boils down to a weighted ERM procedure, with debiasing weights depending on the solution to System \eqref{sys:emp_omega}.
The learning procedure can thus be implemented in three steps as summarized in Figure \ref{fig:db_erm_procedure}.

\begin{figure}[!t]
\begin{center}
\fbox{\begin{minipage}[l]{0.9\textwidth}{
\smallskip
{\small

\begin{itemize}
\item {\bf Input.} Samples $\mathcal{D}_k=\{Z_{k,i},\; i \le n_k \}$, coefficients $\hat{\lambda}_k = n_k/n$, and biasing functions $\omega_k$ for $k \le K$.
\item {\bf Debiasing the raw empirical distribution.} Form the raw empirical distribution based on the pooled sample
$$
\widehat{P}_n=\frac{1}{n}\sum_{k=1}^K\sum_{i=1}^{n_k}\delta_{Z_{k,i}},
$$
\begin{itemize}
\item[(i)] for $k \le K$, compute the functions given by: $\forall \bm{W} \in (\mathbb{R}_+)^K$,
$$
\widehat{\Gamma}_k(\bm{W})=\frac{1}{W_k}\int \frac{\omega_k(z)}{\sum_{l=1}^K \frac{\hat{\lambda}_l\omega_l(z)}{W_l}}d\widehat{P}_n(z);
$$
\item[(ii)] solve \eqref{sys:emp_omega}, i.e., find $\widehat{\bm{W}}_n=(\widehat{W}_{n,1},\; \ldots,\; \widehat{W}_{n,K}) \in (\mathbb{R}_+)^K$ satisfying
\begin{equation}\label{eq:uni_cond}
\max_{1\leq k\leq K} ~ \widehat{W}_{n,k} / \hat{\lambda}_k =1;
\end{equation}
\item[(iii)] for $k \le K$ and $i \le n_k$, compute the weights
$$
\pi_{k,i}=\frac{ \left(\sum_{l=1}^K( \hat{\lambda}_l/\widehat{W}_{n, l})\omega_l(Z_{k,i}) \right)^{-1}}{\sum_{m=1}^K\sum_{j=1}^{n_m}\left( \sum_{l'=1}^K(\hat{\lambda}_{l'}/\widehat{W}_{n, l'})\omega_{l'}(Z_{m,j}) \right)^{-1}},
$$
so as to form the ``debiased'' distribution estimator given by
$$
\widetilde{P}_n=\sum_{k=1}^K\sum_{i=1}^{n_k}\pi_{k,i}\delta_{Z_{k,i}}.$$
\end{itemize}
\item {\bf ERM.} Solve the ERM problem $\min_{\theta\in \Theta}\widetilde{L}_{n}(\theta)$, to produce the solution $\tilde{\theta}_n$, with $\widetilde{L}_{n}(\theta)$ given by
\begin{equation}\label{eq:debiased_risk}
\widetilde{L}_n(\theta)\overset{\mathrm{def}}{=}L_{\widetilde{P}_n}(\theta)=\sum_{k=1}^K\sum_{i=1}^{n_k}\pi_{k,i}\psi(Z_{k,i},\; \theta).
\end{equation}
\end{itemize}}
}
\vspace{-0.1cm}
\end{minipage}\hspace{0.6cm}}
\end{center}
\vspace{-0.2cm}
\caption{\sc ERM based on biased training samples}
\label{fig:db_erm_procedure}
\end{figure}

\begin{enumerate}
\item First, we use the raw empirical distribution $\widehat{P}_n$ to form System \eqref{sys:emp_omega} by computing the estimates $\widehat{\Gamma}_k$ of the $\Gamma_k$;
\item Next, we solve the latter system to build the ``nearly debiased'' estimate $\widetilde{P}_n$ of distribution $P$;
\item Finally, we obtain the decision rule by solving the statistical version of Problem \eqref{pbm:risk_min}, in which $P$ is replaced with $\widetilde{P}_n$.
\end{enumerate}

Discussing how to perform in practice the minimization of the nearly debiased empirical risk estimate of \Cref{eq:debiased_risk}, or of a smooth/penalized version of it, is beyond the scope of the present paper.
However, observe that most machine learning libraries offer the option to reweight the training observations involved in the learning stage in a simple plug-in fashion (e.g., the \texttt{sample\_weight} option for scikit-learn \cite{scikit}).
We also highlight the generality of the above approach, insofar as it may be straightforwardly combined with any ERM-like learning algorithm, for a wide range of biasing scenarios.
We point out however that this generality goes along with the solving of System~\eqref{sys:emp_omega}.
This step cannot be avoided in general, and yields nontrivial solutions, except for simplistic cases such as that discussed in \Cref{sec:simple_example}.
Finally, note that the computational cost induced by the debiasing procedure is low, the unique difference with standard methods lying in the computation of the weights involved in the risk functional, which can be tackled efficiently by means of a Gradient Descent strategy.
Indeed, as can be seen in the proof of \Cref{prop:unique_bounded}, solving System~\eqref{sys:true_omega} is equivalent to minimize the strongly convex function $\bar{D}$, defined for all $\bu = (u_1, \ldots, u_K) \in \mathbb{R}^K$ by
\[
\bar{D}(\bu) = \int \log\left[ \sum_{l=1}^K e^{u_l} \omega_l(z)\right]d\bar{P}(z) - \sum_{l=1}^K \lambda_l u_l\,,
\]
and with Hessian matrix $\bar{D}'' \in \mathbb{R}^{K \times K}$ such that
\begin{equation}\label{eq:hessian}
\left[\bar{D}''(\bu)\right]_{k, k'} = \int \left[\frac{e^{u_k} \omega_k(z) \delta_{kk'}}{\sum_{l=1}^K e^{u_l} \omega_l(z)} - \frac{e^{u_k} \omega_k(z) e^{u_{k'}} \omega_{k'}(z)}{\left(\sum_{l=1}^K e^{u_l} \omega_l(z)\right)^2}\right]d\bar{P}(z)\,.
\end{equation}
By characterizing the curvature of $\bar{D}$, the eigenvalues of $\bar{D}''$ thus influence the convergence of the solution to System \eqref{sys:true_omega}.
Bounding away from $0$ the second smallest eigenvalue of $\bar{D}''$ is actually required in the subsequent nonasymptotic analysis, see \Cref{asm:eigenvalue} for more details.
We conclude this section with two remarks, on the normalization \eqref{eq:uni_cond} and about the possibility to use a sampling approach instead of the reweighting, and a numerical illustration of the benefits of the approach presented here.

\begin{remark}
{\sc (On normalization \eqref{eq:uni_cond})}
As highlighted in \Cref{subsec:biased_models}, recall that System \eqref{sys:emp_omega} is homogeneous of degree $0$.
Hence, normalization \eqref{eq:uni_cond} is just a way to select one $\widehat{\bW}_n$ among all possible solutions.
In \cite{GVW88} for instance, the normalization $\widehat{W}_{n, K} = 1$ is used instead.
Normalization \eqref{eq:uni_cond} happens to be more suited to our nonasymptotic analysis.
In particular, it ensures that $\widehat{\bW}_n$ is unique and bounded away from $0$ with high probability, see \Cref{prop:unique_bounded}.
\end{remark}

% \begin{remark}\label{rmk:solving_sys}
% %
% {\sc (On solving System \eqref{sys:emp_omega})}
% %
% As revealed by the proof of the second claim of \Cref{prop:unique_bounded}, solving System \eqref{sys:emp_omega} is equivalent to finding the root(s) of the gradient of a convex function.
% %
% This in turn can be tackled by means of classical optimization solvers, or by the implementation of a Robbins-Monro algorithm for instance.
%
% \red{Additionally, it has been shown in \cite{Vardi85} that the strong connectivity of the directed graph $\widehat{G}_n$ defined in \Cref{eq:graph_def} is a necessary and sufficient condition for the convex function minimized to be strictly convex, and therefore for the solution to be unique. The strong connectivity assumption may be arduous to check, compared to computing the system's solutions. Thus, one is rather encouraged to compute the debiasing weights without prior verification. If the graph is not connected, the minimized function is simply convex, solutions are not unique but still exist. One of them is found by the algorithm, and one can proceed just like in the strongly convex scenario.}
%
%Otherwise, a criterion such as the gradient norm may alert on the non-convergence, and other bias functions should be chosen, or debiasing abandoned for standard ERM.
%
% \end{remark}

\begin{remark}
{\sc (Plug-in \textit{vs} Sampling)}
From a practical perspective, \mbox{modifying} the objective function using the weights computed at step $(iii)$ in the above scheme is not the only option.
An alternative to learn the predictive rule would be to sample observations from the distribution \eqref{eq:est_dist}, given the original data.
This would generate a new (unique and nearly debiased) dataset, from which any ERM-based learning algorithm can be run in a standard fashion.
\end{remark}

\begin{example}{\sc (Clustering, bis)}\label{ex:clustering_2}
%In \Cref{fig:clusters}, we show the benefit of using our approach on a $2$-dimensional clustering problem.
%
%The distribution 
Let distribution $P$ be a mixture of two \mbox{$2$-d} Gaussian distributions, centered at $(X_1, X_2) = (-1.5, -2)$ and $(X_1, X_2) = (1.5, 2)$ respectively.
On the left of \Cref{fig:clusters}, an unbiased dataset is displayed, on which the vanilla EM algorithm finds the centroids accurately.
The second dataset (on the right) is actually composed of $3$ samples: that on the left, for which $\omega_1(X) = \mathbb{I}\{X_1 \le -0.5\}$, that in the middle, for which $\omega_2(X) = \mathbb{I}\{-0.7 \le X_1 \le 0.7\}$, and that on the right for which $\omega_3(X) = \mathbb{I}\{0.5 \le X_1\}$.
The dataset in the middle is of size $n_2 = 300$ observations, while the left/right ones are composed of $n_1 = n_3 = 30$ observations.
As expected, the centroids found by means of the vanilla version of the EM algorithm are heavily shifted towards the center, whereas the debiased variant of the EM algorithm is able to leverage the bias functions information so as to recover nearly the correct centroids.

\begin{figure*}[!t]
\begin{center}
\begin{minipage}[c]{0.49\textwidth}
\centering
\includegraphics[height=4.65cm]{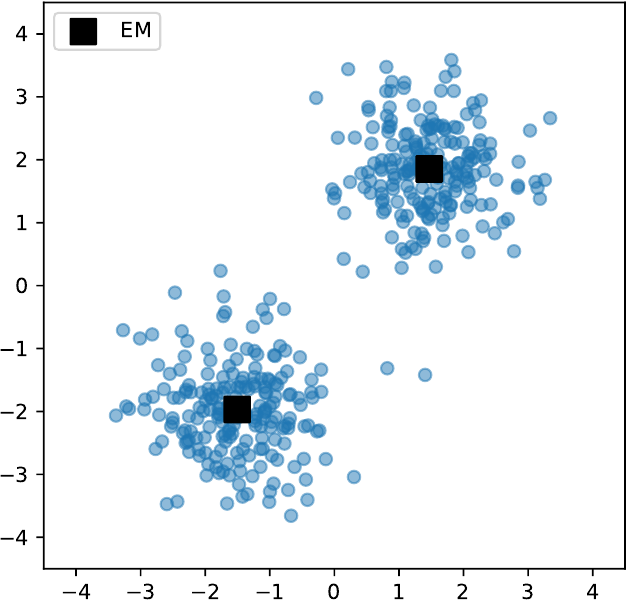}
\end{minipage}
\hfill
\begin{minipage}[c]{0.49\textwidth}
\centering
\includegraphics[height=4.65cm]{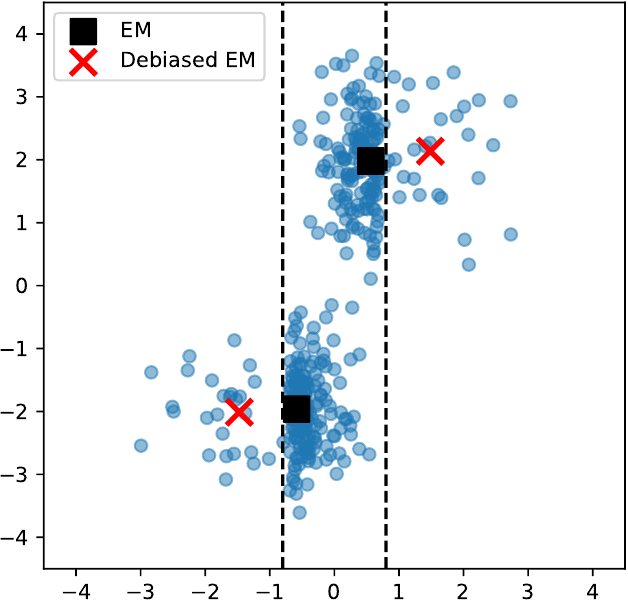}
\end{minipage}
\caption{{\sc (Clustering by vanilla and debiased EM algorithms)}.
In absence of sampling bias (left), the vanilla EM algorithm finds the right centroids.
When the data are biased, over-represented towards the center region (right), the centroids obtained through the vanilla EM algorithm are also attracted towards the center region.
In contrast, the debiased EM algorithm produces nearly the same centroids than those obtained from the unbiased dataset.}
\label{fig:clusters}
\end{center}
\vspace{-0.5cm}
\end{figure*}

\end{example}

\section{Empirical Risk Minimization in Biased Sampling Models}
\label{sec:main}
In this section, we provide theoretical guarantees for the extension of ERM to biased training samples we have introduced in \Cref{subsec:learning_bias}.
Unsurprisingly, the subsequent nonasymptotic analysis requires slightly more stringent assumptions than those involved in the asymptotic study carried out in \cite{GVW88}, and listed in \Cref{subsec:biased_models}.
In particular, \Cref{hyp:rates} strengthens \Cref{hyp:lambda_cvg} in order to control the fluctuations of the sample sizes, so as to establish finite-sample learning rate bounds.
In the same spirit, additional parameters are introduced to guarantee that crucial quantities are bounded away from critical values.
Hence, expectations involved in \Cref{hyp:connect_gvw} are supposed to be greater than $\kappa > 0$, while the minimal positive value of the $\omega_k$ is lower bounded by $\varepsilon > 0$ (see \Cref{hyp:bounded_omega}).
Notice that this lower bound does not prevent the $\omega_k$ to vanish, preserving the generality of the approach.

\begin{assumption}\label{hyp:rates}
There exist $(\lambda_1, \ldots, \lambda_K) \in (0, 1)^K$ satisfying $\sum_{k=1}^K \lambda_k = 1$, and $C_\lambda, \underline{\lambda} > 0$ such that for all $k \le K$ and $n \ge K$ it holds
\begin{equation}\label{eq:size_fluct}
\underline{\lambda} \le \lambda_k\,, \qquad \underline{\lambda} \le \hat{\lambda}_k\,, \qquad \text{and} \qquad \big| \hat{\lambda}_k - \lambda_k \big| \leq \frac{C_\lambda}{\sqrt{n}}\,.
\end{equation}
\end{assumption}

Observe that the control of the order of magnitude of the sample sizes and that of their fluctuations in \Cref{hyp:rates} cannot be avoided, since the goal here is to establish nonasymptotic (learning) rate bounds, see \Cref{lem:decompo} in particular.
%
%Indeed, it is necessary to control the order of magnitude of the fluctuations of the sample sizes, in particular after the decomposition of \Cref{lem:decompo}.
%
%Controlling the fluctuations is however not enough to bound the $\hat{\lambda}_k$ away from $0$, hence the necessity of the first part too.}

\begin{remark}
We point out that, in the situation where the vector of sample sizes $(n_1,\; \ldots,\; n_K)$ is random, distributed as a multinomial of size $n$ with parameters $(\lambda_1,\; \ldots,\; \lambda_k)$, the last bounds in \Cref{eq:size_fluct} simultaneously hold true for any $k$ and an appropriate constant $C_\lambda$ with overwhelming probability.
Indeed, using Hoeffding's inequality (see \cite{Hoeffding1963}) combined with the union bound for instance, one obtains that, for any $\delta\in (0,1)$, all these conditions are fulfilled with probability larger than $1-\delta$ with $C_\lambda=\sqrt{\log(K/\delta)/2}$, and that $\underline{\lambda} \ge \min_k \lambda_k-C_\lambda/\sqrt{n}$, provided that $n>C_\lambda^2/\min_k \lambda_k$.
Note that for simplicity, we restrict our analysis to the situation where the sample sizes are deterministic, the random case being a straightforward extension.
\end{remark}

\begin{assumption}\label{hyp:connect}
For $\kappa > 0$, define $G_\kappa$ the (undirected) graph with vertices in $\{1, \ldots, K\}$, and edge between $k$ and $l$ $(k\ne l)$ if and only if
\begin{equation*}
\int \mathbb{I}\{\omega_k(z) > 0 \} \cdot \mathbb{I}\{\omega_l(z) > 0\}dP(z) \ge \kappa.
\end{equation*}
There exists $\kappa > 0$ such that $G_\kappa$ is connected.
\end{assumption}

From an algebraic viewpoint, one may classically check whether \Cref{hyp:connect} is fulfilled or not by means of a breadth-first search algorithm, or by examining the spectrum of the Laplacian matrix of $G_\kappa$ for instance, see e.g., \cite{GR01}.
Note that such a verification would require to have access to $P$, which is unknown in general.
However, we highlight that in practice the connectivity property that must be checked concerns $\widehat{G}_n$, the empirical counterpart of $G$ defined in \Cref{eq:graph_def}, which only depends on the observed empirical distributions $\widehat{P}_k$.

\begin{assumption}\label{hyp:bounded_omega}
There exists $\varepsilon > 0$ such that
\begin{equation*}
\forall z\in \mathcal{Z},~\forall k \le K, \qquad \varepsilon\cdot\mathbb{I}\{\omega_k(z) > 0\} \le \omega_k(z) \le 1.
\end{equation*}
In particular this implies $\omega_k(z_i) \ge \varepsilon$ for all $z_i \in \mathcal{D}_k$, and $\Omega_k \le 1$ for all $k \le K$.
\end{assumption}

\smallskip\noindent
\begin{minipage}[l]{0.62\textwidth}
Note that parameters $\kappa$ and $\varepsilon$ allow to quantify the overlap between two biasing functions, in a way that extends the simple overlap/non overlap condition of \cite{GVW88} (recovered here by $\mathbb{I}\{\kappa \varepsilon \ne 0 \}$). The results derived below typically hold true with probability $1 - e^{-(\kappa \varepsilon)^2n}$, see e.g., \Cref{prop:unique_bounded}, confirming that no learning is possible without overlap, but also providing the new insight that performances improve with the overlapping.
\end{minipage}
\hfill
\begin{minipage}[c]{0.33\textwidth}
\centering
\includegraphics[width=\textwidth]{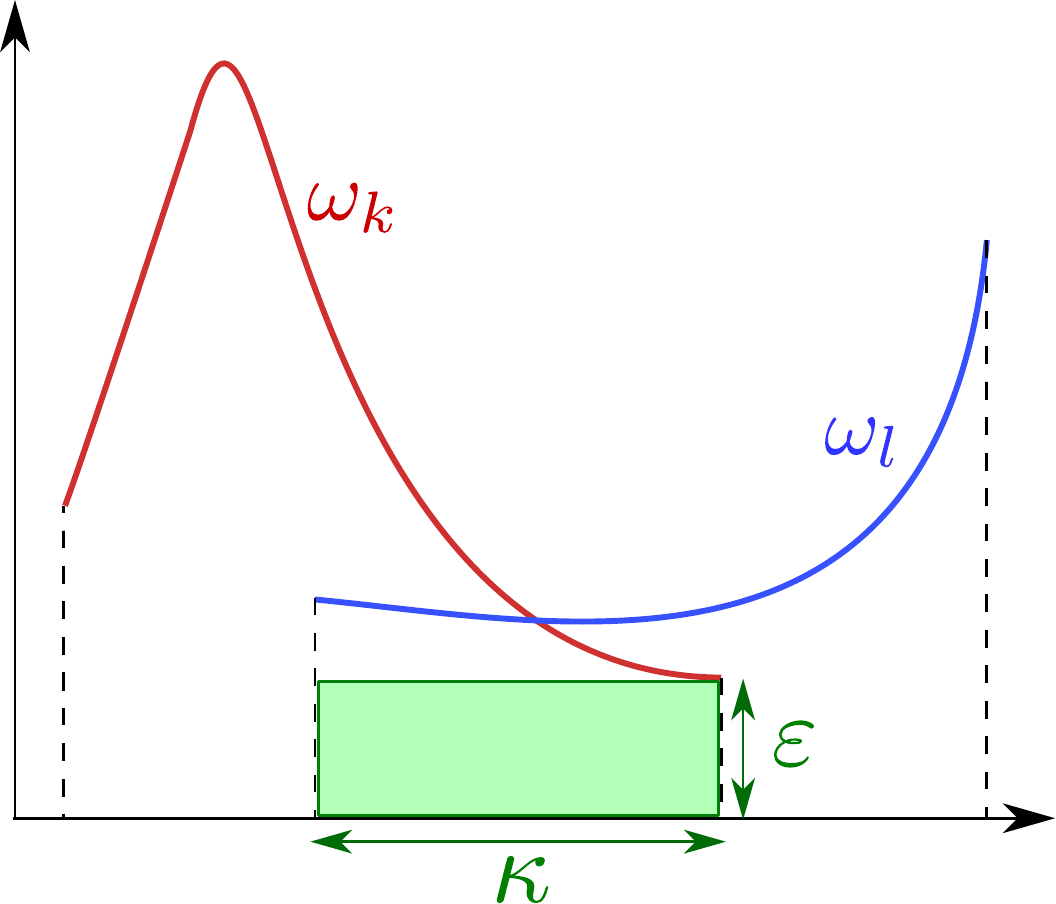}
{\footnotesize {\sc Fig 3.} {\it Overlap control.}}
\end{minipage}
\smallskip
\setcounter{figure}{3}

\begin{remark}
We point out that, in \Cref{ex:classif}, \Cref{hyp:ident} simply means that $\mathcal{X}=\mathcal{X}_1\cup \cdots\cup \mathcal{X}_K$.
\Cref{hyp:bounded_omega} is fulfilled with $\varepsilon =1$, and \Cref{hyp:connect} can be checked in a simple manner, insofar as we have: $\forall 1\leq k\neq l\leq K$, $e_{k,l}=1 \Leftrightarrow \mu\left(   \mathcal{X}_k\cap \mathcal{X}_l   \right)\geq \kappa$.
In \Cref{ex:regression}, \Cref{hyp:ident} is directly fulfilled, just like \Cref{hyp:bounded_omega} with $\varepsilon = 1$.
\end{remark}

As discussed at the end of Section \ref{subsec:learning_bias}, we also introduce an assumption on the second smallest eigenvalue of the Hessian matrix $\bar{D}''$ defined in \Cref{eq:hessian}.
\begin{assumption}\label{asm:eigenvalue}
Let $U = \log(K/\varepsilon)\sum_{t=1}^{K-1}2^t(\underline{\lambda}\kappa \varepsilon)^{-t}$, $\mathcal{U} = [0, U]^K \subset \mathbb{R}^K$, and $\sigma > 0$.
See \Cref{prop:strong_cvx} in \Cref{apx:proof_dev_W} for more insights about the first two values.
For all $\bu \in \mathcal{U}$, $\sigma_2(\bar{D}''(\bu)) \ge \sigma$, where $\sigma_2(A)$ denotes the second smallest eigenvalue of a matrix $A$.
\end{assumption}

As revealed by the proof of \Cref{prop:dev_W}, \Cref{asm:eigenvalue} is required to control the deviation $\|\hat{\bm{u}}_n - \bm{u}^*\|_2$ in terms of $\big\|\widehat{D}'_n(\bm{u}^*) - \bar{D}'_n(\bm{u}^*)\big\|_2$, using the curvature of $D$ in the non-flat parts of the optimization landscape.
Note that \Cref{asm:eigenvalue} is not needed in the asymptotic analysis as even the smallest possible curvature (and we know it is strictly positive by the proof of Proposition~1 in \cite{GVW88}) is still sufficient when $n$ goes to infinity.
On the opposite, to establish finite sample bounds, we have to bound away from zero the second smallest eigenvalue of $\bar{D}''(\bm{u})$, uniformly over $\mathcal{U}$, in an explicit manner.
Although such a lower bound is always attained, as $\bar{D}''$ is continuous on the compact set $\mathcal{U}$, its dependence with respect to the problem instance (i.e., the distribution $P$, the biasing functions $\omega_k$, the sample proportions $\lambda_k$) is non-trivial.
For this reason, we rather opted for explicitly introducing a parameter $\sigma$ to materialize this lower bound.
The results subsequently derived then depend on $\sigma$ in a more interpretable fashion.

Equipped with these assumptions, we now carry out a detailed rate bound analysis.
The first step, described in \Cref{subsec:existence}, consists in showing that with overwhelming probability the solution to System \eqref{sys:emp_omega} exists, is unique, and belongs to a compact set bounded away from $0$ (\Cref{prop:unique_bounded}).
This crucial property then allows to derive nonasymptotic concentration bounds for $\widehat{\bW}_n$ (\Cref{prop:dev_W}) and next for $\widehat{\bm{\Omega}}_n$ (\Cref{prop:dev_omega}).
The generalization results are finally stated in \Cref{subsec:generalization}, under a standard complexity assumption.
The guarantees for the minimizers of the debiased risk version are established in \Cref{thm:main}, and a corollary about the excess risk is discussed.
When the concept class is composed of indicator functions of subsets, a tighter analysis is presented (\Cref{thm:dkw}), which provides an extension of the Dvoretsky-Kiefer-Wolfowitz inequality under biased sampling models.

%%%%%%%%%%%%%%%%%%%%%
%                   %
%  SYSTEM SOLUTION  %
%                   %
%%%%%%%%%%%%%%%%%%%%%

\subsection{Existence, Uniqueness, and Concentration of the Solution}
\label{subsec:existence}

As detailed in \Cref{subsec:learning_bias}, our debiasing ERM procedure critically relies on solving System \eqref{sys:emp_omega}.
It is shown in \cite{GVW88} (Theorem 1.1 therein) that the latter admits a unique solution if and only if a directed and statistical (i.e., with $\widehat{P}_k$ instead of $P_k$) version  of graph $G$ in \Cref{hyp:connect_gvw}, denoted by $\widehat{G}_n$ thereafter, is strongly connected.
From a limit perspective, the strong law of large numbers suffices to guarantee that, with probability $1$, the edges of $\widehat{G}_n$ are asymptotically the same as those of $G$.
\Cref{hyp:connect_gvw} then allows to conclude that $\widehat{G}_n$ is strongly connected and that System \eqref{sys:emp_omega} admits a unique solution.
This result is stated as Corollary 1.1 in \cite{GVW88}.
The proposition below refines this assertion from a nonasymptotic angle.
It shows that existence and uniqueness actually occur with overwhelming probability.
Uniqueness is of course understood up to the homogeneity property.
To avoid any ambiguity, $\widehat{\bW}_n$ now refers to the solution to System \eqref{sys:emp_omega} satisfying $\max_{k \le K} \widehat{W}_{n, k} / \hat{\lambda}_k = 1$, see \Cref{eq:uni_cond}.
Similarly, $\bW^*$ is assumed to verify $\max_{k \le k} W^*_k / \lambda_k = 1$.
\Cref{prop:unique_bounded} also shows that both $\widehat{\bW}_n$ and $\bW^*$ belong to a compact set bounded away from $0$.
This property is key in the subsequent analysis, as $\widehat{\bm{W}}_n$ is often present in denominators, see e.g., \Cref{eq:est_dist}.
Note that to keep notation simple, we use generic constants in the statements of the results, that may have different values from one proposition to the other.
For completeness, we provide their exact values in the technical proofs of the Appendix section.
Importantly, they only depend on parameters $K, C_\lambda, \underline{\lambda}, \kappa, \varepsilon, \sigma$ introduced in Assumptions \ref{hyp:rates}, \ref{hyp:connect}, \ref{hyp:bounded_omega}, and \ref{asm:eigenvalue}.
Although $K$ is treated as a constant here, note that our results remain meaningful as long as $K = o(n)$.
If $K$ grows linearly with $n$, it is immediate to see that the dataset sizes $n_k$ are then necessarily bounded, making a consistent recovery of the $\widehat{P}_k$'s impossible, and the debiasing procedure bound to fail.

\begin{proposition}\label{prop:unique_bounded}
Suppose that Assumptions \ref{hyp:rates}, \ref{hyp:connect}, and \ref{hyp:bounded_omega} are satisfied.
Then, there exist $M, c, \rho > 0$, depending only on $K, \underline{\lambda}, \kappa, \varepsilon$, such that for all $n \ge \log(M) / c$, it holds with probability at least $1 - M\exp(-c n)$:
\begin{itemize}
\item the solution $\widehat{\bm{W}}_n$ to System \eqref{sys:emp_omega} exists and is unique,
\item for all $k \le K, \quad \rho \le \widehat{W}_{n, k} \le 1, \quad \text{and} \quad \rho \le W^*_k \le 1$.
\end{itemize}
\end{proposition}

\noindent The rationale behind the proof is similar to that used to establish Corollary~1.1 in \cite{GVW88}.
Rather than simply establishing that the edges of $\widehat{G}_n$ asymptotically match those of $G$, we bound the probability that they differ from those of $G_\kappa$, defined in \Cref{hyp:connect}.

\begin{proof}
First, define the directed graph $\widehat{G}_n$ with vertices in $\{1, \ldots, K\}$ and edge $k\rightarrow l$ if and only if
\begin{equation}\label{eq:graph_def}
\int \mathbb{I}\{\omega_k(z) >0 \} d\widehat{P}_l(z)>0.
\end{equation}
The graph $\widehat{G}_n$ is said to be strongly connected if, for any pair of vertices $(k,l)$, there exist a directed path from $k$ to $l$ and a directed path from $l$ to $k$.
It is proved in \cite{Vardi85} (see also Theorem 1.1 in \cite{GVW88}) that this is a necessary and sufficient condition for System \eqref{sys:emp_omega} to have a unique solution.
To show that $\widehat{G}_n$ is strongly connected, we prove that the left hand side in \Cref{eq:graph_def} is sufficiently close to (a weighted version of) the link condition in \Cref{hyp:connect} with overwhelming probability.
Let $(k, l)$ be an edge in $G_\kappa$.
Using \Cref{hyp:connect,hyp:bounded_omega} we get:
\begin{align*}
\int \mathbb{I}\{\omega_k(z) > 0\}dP_l(z) &= \int \mathbb{I}\{\omega_k(z) > 0\} \frac{\omega_l(z)}{\Omega_l}dP(z),\\
&\ge \varepsilon \int \mathbb{I}\{\omega_k(z) > 0\} \cdot \mathbb{I}\{\omega_l(z)>0\}dP(z),\\
&\ge \kappa \varepsilon.
\end{align*}
Now, observe that the left-hand side in \Cref{eq:graph_def} is the empirical version of the above term.
By Hoeffding's inequality, for every $t > 0$ it holds:
\begin{align*}
\mathbb{P}\left\{ \int \mathbb{I}\{\omega_k(z) > 0\}d\widehat{P}_l(z) - \int \mathbb{I}\{\omega_k(z) > 0\}dP_l(z)\le -t \right\} &\le \exp(-2n_lt^2),\\
&\le \exp(-2\underline{\lambda}nt^2).
\end{align*}
In particular, setting $\delta = \exp\left(-\frac{\underline{\lambda}(\kappa \varepsilon)^2n}{2}\right)$ it holds with probability at least $1 - \delta$:
\begin{equation}\label{eq:sufficient_edge}
\int \mathbb{I}\{\omega_k(z) > 0\}d\widehat{P}_l(z) \ge \int \mathbb{I}\{\omega_k(z) > 0\}dP_l(z) - \frac{\kappa \varepsilon}{2} \ge \frac{\kappa \varepsilon}{2},
\end{equation}
so that $k \rightarrow l$ in $\widehat{G}_n$.
The exact same reasoning can be applied after having switched $k$ and $l$.
The union bound then gives that with probability at least $1 - 2\delta$ it holds: $k \rightarrow l$ and $l \rightarrow k$ in $\widehat{G}_n$.
Now, $G_\kappa = (V, E)$ being connected, we know that there exists a set of edges $E_\text{min} \subset E$ of cardinal $K-1$ such that $G_\text{min} = (V, E_\text{min})$ is connected.
Applying the above method to every edge in $E_\text{min}$, we get that with probability at least $1 - 2(K-1)\delta$, every pair $(k, l)$ linked in $G_\text{min}$ is linked both ways in $\widehat{G}_n$.
Since $G_\text{min}$ is connected, this means that $\widehat{G}_n$ is strongly connected.
The proof is concluded by setting $M = 2(K-1)$, and $c = \underline{\lambda}(\kappa \varepsilon)^2/2$.
As a lengthy technical analysis is required to identify $\rho$, the proof of the second claim of \Cref{prop:unique_bounded} is postponed to \Cref{apx:proof_unique_bounded}.
\end{proof}

\noindent The identification of a compact set, bounded away from $0$ and containing $\widehat{\bW}_n$ and $\bW^*$ with high probability, is essential to carry out a nonasymptotic analysis.
In particular, it permits to derive the following exponential concentration bound.
\begin{proposition}\label{prop:dev_W}
Suppose that Assumptions \ref{hyp:rates}, \ref{hyp:connect}, \ref{hyp:bounded_omega}, and \ref{asm:eigenvalue} are satisfied.
Then, there exist $M, M', c, c', \gamma, n_0 > 0$, depending only on $K, C_\lambda, \underline{\lambda}, \kappa, \varepsilon, \sigma$, such that for all $t > 0$ and $n \ge n_0$ it holds:
\begin{equation*}
\mathbb{P}\left\{\big\|\widehat{\bm{W}}_n - \bm{W}^*\big\|_2 > \frac{\gamma}{\sqrt{n}} + t\right\} \le Me^{-cn} + M'e^{-c'nt^2}.
\end{equation*}
\end{proposition}
\noindent Note that the sup norm $\|\widehat{\bW}_n - \bW^*\|_\infty$ is upper bounded by the Euclidean norm $\|\widehat{\bW}_n - \bW^*\|_2$, so that the inequality in \Cref{prop:dev_W} is simultaneously satisfied by $|\widehat{W}_{n, k} - W^*_k|$ for all $k \le K$.
The proof uses the same reparameterization as for the second claim of \Cref{prop:unique_bounded}.
It involves some notion of curvature related to System~\eqref{sys:true_omega}, that characterizes the hardness of the problem.
The compactness derived in \Cref{prop:unique_bounded} is then used to uniformly lower bound this curvature, see \Cref{asm:eigenvalue}.
Technical details are provided in \Cref{apx:proof_dev_W}.

Using \Cref{eq:recover_omega,eq:recover_omega_emp}, one can finally control the deviation of $\widehat{\bm{\Omega}}_n$ with respect to $\bm{\Omega}$, as revealed by the following proposition, whose proof is detailed in \Cref{apx:proof_dev_omega}.
\begin{proposition}\label{prop:dev_omega}
Suppose that Assumptions \ref{hyp:rates}, \ref{hyp:connect}, \ref{hyp:bounded_omega}, and \ref{asm:eigenvalue} are satisfied.
Then, there exist $M, M', c, c', \gamma, n_0$, depending only on $K, C_\lambda, \underline{\lambda}, \kappa, \varepsilon, \sigma$, such that for all $t > 0$ and $n \ge n_0$ it holds:
\begin{equation*}
\mathbb{P}\left\{\big\|\widehat{\bm{\Omega}}_n - \bm{\Omega}\big\|_\infty > \frac{\gamma}{\sqrt{n}} + t\right\} \le Me^{-cn} + M'e^{-c'nt^2}.
\end{equation*}
\end{proposition}

Hence, we have shown that, using the procedure described in \Cref{subsec:learning_bias}, we can compute an estimate $\widehat{\bm{\Omega}}_n$ of $\bm{\Omega}$ with good nonasymptotic concentration properties.
The last step consists in analyzing the performance of the minimizers of the debiased risk \eqref{eq:debiased_risk} when $\widetilde{P}_n$ is built using \Cref{eq:plug_in} and $\widehat{\bm{\Omega}}_n$.

%%%%%%%%%%%%%%%%%%%%
%                  %
%  GENERALIZATION  %
%                  %
%%%%%%%%%%%%%%%%%%%%

\subsection{Generalization Ability of Minimizers of the Debiased Risk}
\label{subsec:generalization}

As a first go, we introduce the following standard complexity assumption on the class $\mathcal{F} = \mathcal{F}_\Theta=\{\psi(\cdot,\theta)\colon\theta\in \Theta \}$, see e.g., Equation (2.14.6) in \cite{VanderVaart1996}.
\begin{assumption}\label{hyp:complex}
The collection of functions $\mathcal{F}_\Theta=\{\psi(\cdot,\theta)\colon\theta\in \Theta  \}$ satisfies $|\psi(z, \theta)| \le 1$ for all $z, \theta$, and is a uniform Donsker class (relative to $L_2$) with polynomial uniform covering numbers, i.e., there exist constants $C_\Theta>0$ and $r\geq 1$ such that for all $\zeta>0$
\[
\sup_{Q}\mathcal{N}(\zeta,\; \mathcal{F}_\Theta,\; L_2(Q) )\leq (C_\Theta/\zeta)^r
\]
where the supremum is taken over the set of probability measures $Q$ on $\mathcal{Z}$, and $\mathcal{N}(\zeta, \mathcal{F}_\Theta, L_2(Q) )$ is the minimum number of $L_2(Q)$ balls of radius $\zeta$ needed to cover $\mathcal{F}_\Theta$.
\end{assumption}

\begin{remark}\label{rmk:complex}
The above hypothesis is a classic complexity assumption.
Of course, the subsequent rate bound analysis can be straightforwardly extended to settings involving alternative complexity conditions, such as e.g., finite {\sc VC} dimension, Rademacher averages.
Recall that a collection of functions $\mathcal{F}_\Theta$ of finite {\sc VC} dimension $V<+\infty$, and with envelope function $F \equiv 1$, satisfies \Cref{hyp:complex} with $r=2V-2$, and $C_\Theta$ depending only on $V$, see e.g., Theorem 2.6.7 in \cite{VanderVaart1996}.
%
% Recall that, in the binary classification for instance, if the collection of classifiers $\mathcal{G}$ considered is of finite {\sc VC} dimension $V<+\infty$, then the collection of functions $\{(x,y)\in\mathcal{X}\times\{-1,\; +1\}\mapsto \mathbb{I}\{ Y\neq g(X)\},\; \; g\in \mathcal{G} \}$ satisfies \Cref{hyp:complex} with $C_\Theta = CV(4e)^{V}$, and $r=V-1$, $C$ being a universal constant, see e.g., Theorem 2.6.4 in \cite{VanderVaart1996}.
%
\end{remark}

The main argument of the subsequent analysis then consists in showing that the uniform deviation between the nearly debiased risk and the true risk
\begin{equation}\label{eq:max_dev}
\sup_{\theta\in \Theta}\left|\widetilde{L}_n(\theta) - L(\theta)\right|
\end{equation}
is small with high probability.
This is however far from being as straightforward as in the unbiased situation, since the set of random variables $\{\widetilde{L}_n(\theta) - L(\theta)\}_{\theta\in \Theta}$ is not an empirical process (i.e., a collection of i.i.d. averages), and the standard concentration inequalities therefore do not apply.
Indeed, we recall that $\widetilde{L}_n(\theta)$ depends on $\widehat{\bm{\Omega}}_n$, which is obtained from the solution to System \eqref{sys:emp_omega}, and for which no closed analytical form is available in general, see \Cref{subsec:learning_bias}.
To bypass this difficulty, we decompose the excess of risk $|\widetilde{L}_n(\theta) - L(\theta)|$ as follows.
Let
\[\hat{h}_{n, \theta}(z) = \psi(z,\theta) \left( \sum_{k=1}^K \frac{\hat{\lambda}_k \omega_k(z)}{\widehat{\Omega}_{n, k}} \right)^{-1}, \text{~~and~~} h_\theta(z) = \psi(z,\theta) \left( \sum_{k=1}^K \frac{\lambda_k \omega_k(z)}{\Omega_k} \right)^{-1}.
\]
We have (see \Cref{lem:decompo} for details):

\begin{align}
&\left|\widetilde{L}_n(\theta) - L(\theta)\right|\nonumber\\
&= \left|\int \psi(z,\theta) \left( \sum_{k=1}^K \frac{\hat{\lambda}_k \omega_k(z)}{\widehat{\Omega}_{n, k}} \right)^{-1} d\widehat{P}_n(z) - \int \psi(z,\theta)\left(\sum_{k=1}^K \frac{\lambda_k \omega_k(z)}{\Omega_k} \right)^{-1} d\bar{P}(z)\right|\nonumber\\
&= \left| \int \hat{h}_{n, \theta}(z)d\widehat{P}_n(z) - \int h_\theta(z)d\bar{P}(z)\right|\nonumber\\
&\le \big\|\hat{h}_{n, \theta} - h_\theta \big\|_\infty \hspace{0.1cm}+\hspace{0.1cm} \|h_\theta\|_\infty \sum_{k=1}^K \left|\hat{\lambda}_k - \lambda_k\right| \hspace{0.1cm}+\hspace{0.1cm} \sum_{k=1}^K \hat{\lambda}_k \left| \int h_\theta d\widehat{P}_k - \int h_\theta dP_k\right|\label{eq:decompo_text}
\end{align}

\noindent As it is assumed that $|\psi(z, \theta)| \le 1$, the first term of the right hand side of~\eqref{eq:decompo_text} actually depends on $\|\widehat{\bm{\Omega}}_n - \bm{\Omega}\|_\infty$ only, and can thus be bounded uniformly over $\Theta$ using \Cref{prop:dev_omega}.
Similarly, the second term depends on the $|\hat{\lambda}_k - \lambda_k|$, and can be uniformly upper bounded using \Cref{hyp:rates}.
Finally,  the last term writes as the sum of empirical processes, indexed by $\Theta$, for which standard arguments apply.
%
% Note that we recover that the global speed is governed by the samples with the fewest observation, which is a standard feature of multi-sample settings.
%
This leads to the following theorem, whose complete proof can be found in \Cref{apx:proof_thm}.

\begin{theorem}\label{thm:main}
Suppose that Assumptions \ref{hyp:rates}, \ref{hyp:connect}, \ref{hyp:bounded_omega}, \ref{asm:eigenvalue}, and \ref{hyp:complex} are satisfied.
Then, there exist $M, M', M'', c, c', c'', \gamma, n_0$, depending only on $K, C_\lambda, \underline{\lambda}, \kappa, \varepsilon, \sigma, C_\Theta, r$, such that for all $t > 0$ and $n \ge n_0$ it holds:
\begin{equation*}
\mathbb{P}\left\{\sup_{\theta \in \Theta} \left|\widetilde{L}_n(\theta) - L(\theta)\right| > \frac{\gamma}{\sqrt{n}} + t\right\} \le Me^{-cn} + M'e^{-c'nt^2} + (\sqrt{n}t)^r M''e^{-c''nt^2}.
\end{equation*}
\end{theorem}

\noindent An important corollary of \Cref{thm:main} is obtained by combining the above bound with the following argument.
Let $\tilde{\theta}_n = \argmin_\Theta \widetilde{L}_n(\theta)$, we have
\begin{equation*}
L(\tilde{\theta}_n)-\inf_{\theta\in \Theta}L(\theta)\leq 2 \sup_{\theta\in \Theta}\left\vert \widetilde{L}_n(\theta)-L(\theta) \right\vert.
\end{equation*}
This immediately results in \Cref{cor1}, which reveals that minimizers of the ``debiased'' version of the empirical risk achieve exactly the same learning rate as minimizers of an (unbiased) empirical risk based on $n\geq 1$ independent observations $Z_1,\; \ldots,\; Z_n$ drawn from the test distribution $P$.
Notice that an analogous bound for the expectation of the risk excess of \Cref{eq:debiased_risk}'s minimizers can be proved using the same argument.

\begin{corollary}\label{cor1}
Suppose that the assumptions of \Cref{thm:main} are satisfied, and keep the same values for $M, M', M'', c, c', c'', \gamma, n_0$.
Let $\tilde{\theta}_n$ be any minimizer of the debiased risk $\widetilde{L}_n$ defined in \eqref{eq:debiased_risk}.
Then, for all $t > 0$ and $n \ge n_0$ it holds:
\begin{equation*}
\mathbb{P}\left\{L(\tilde{\theta}_n)-\inf_{\theta\in \Theta}L(\theta) > \frac{2\gamma}{\sqrt{n}} + 2t\right\} \le Me^{-cn} + M'e^{-c'nt^2} + (\sqrt{n}t)^r M''e^{-c''nt^2}.
\end{equation*}
\end{corollary}

\noindent Another application of particular interest is the case where $q=1$ (univariate case), and where the class $\mathcal{F}$ is the set composed of all indicator functions $z\in\mathbb{R}\mapsto \mathbb{I}\{z\leq \tau \}$, for $\tau\in \mathbb{R}$.
Recall that in the i.i.d. case, the Dvoretzky-Kiefer-Wolfowitz (DKW) inequality, see e.g., \cite{Massart90}, then yields: $\forall t\geq 0$,
\begin{equation}\label{eq:DKW}
\mathbb{P}\left\{  \sup_{z \in \mathbb{R}}\left|(\widehat{P}_n - P)((-\infty,\; z])\right|\geq t \right\}\le 2e^{-2nt^2}
\end{equation}
where $z\in\mathbb{R}\mapsto \widehat{P}_n((-\infty,\; z])=(1/n)\sum_{i=1}^n\mathbb{I}\{Z_i\leq z\}$ denotes the empirical cumulative distribution function based on i.i.d. observations $Z_1,\; \ldots,\; Z_n$ drawn from the univariate probability distribution $P$.
Analogously, under the sample biasing models, the quantity \eqref{eq:max_dev} then corresponds to the maximal deviation $\sup_{z \in \mathbb{R}}\big|(\widetilde{P}_n - P)((-\infty,\; z])\big|$.
While a functional central limit theorem for this cdf estimator is established in \cite{GVW88}, the application of \Cref{thm:main} allows to refine this statement from a nonasymptotic perspective.
However, recalling that the class composed of half-lines is of {\sc VC} dimension $2$, and thus satisfies \Cref{hyp:complex} with $r = 2V - 2 = 2$, the bound obtained contains a term of order $nt^2e^{-nt^2}$, which does not match \eqref{eq:DKW}.
A sharper analysis, leveraging the fact that $\mathcal{F}$ is a class of indicator functions, is necessary, see \Cref{apx:proof_thm_dkw}.
The refined rate thus achieved (\Cref{thm:dkw}) then matches \eqref{eq:DKW}, and provides an exact extension of the DKW inequality under biased sampling models.

\begin{theorem}\label{thm:dkw}
Suppose that Assumptions \ref{hyp:rates}, \ref{hyp:connect}, \ref{hyp:bounded_omega}, and \ref{asm:eigenvalue} are satisfied.
Then, there exist $M, M', c, c', \gamma, n_0$, depending only on $K, C_\lambda, \underline{\lambda}, \kappa, \varepsilon, \sigma$, such that for all $t > 0$ and $n \ge n_0$ it holds:
\begin{equation*}
\mathbb{P}\left\{ \sup_{z \in \mathbb{R}}\left|(\widetilde{P}_n - P)((-\infty,\; z])\right| > \frac{\gamma}{\sqrt{n}} + t \right\}\le Me^{-cn} + M'e^{-c'nt^2}.
\end{equation*}
\end{theorem}

\noindent Finally, we point out that the finite sample analysis carried out in this paper can be used in the context of $M$-estimation as well. Indeed, under the additional hypothesis that there exists a unique minimizer $\theta^*$ of the true risk $L(\cdot)$ in the state space $\Theta\subset \mathbb{R}^d$ combined with usual smoothness/coercivity assumptions related to the risk functional,  nonasymptotic  bounds for the estimation error $\vert\vert \tilde{\theta}_n -\theta \vert\vert$ can be classically deduced from the excess of risk bounds proved here.
%
%Instead of excess risk bounds, one then obtain bounds on the estimation errors of $M$-estimators under classic identifiability assumptions.}

\section{Numerical Experiments}
\label{sec:num}
In this section, we display numerical results illustrating the performance of the extension of the ERM approach we propose when training data suffer from selection bias.
First, observe that the procedure is by no means computationally expensive, the sole difference with standard methods lying in the computation of the weights involved in the risk functional.
In addition, it can be readily implemented in a plug-in manner with most machine learning libraries, using e.g., scikit-learn's \texttt{sample\_weight} option during the learning stage, see \cite{scikit}.

\begin{figure*}[!t]
\begin{center}
\begin{minipage}[l]{0.49\textwidth}
\includegraphics[height=3.8cm]{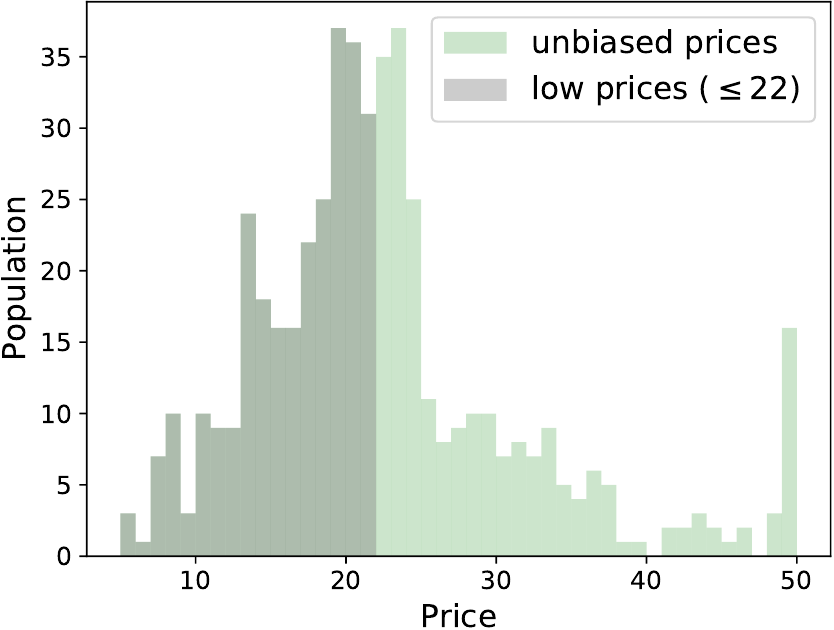}
\end{minipage}
\hfill
\begin{minipage}[l]{0.49\textwidth}
\includegraphics[height=3.8cm]{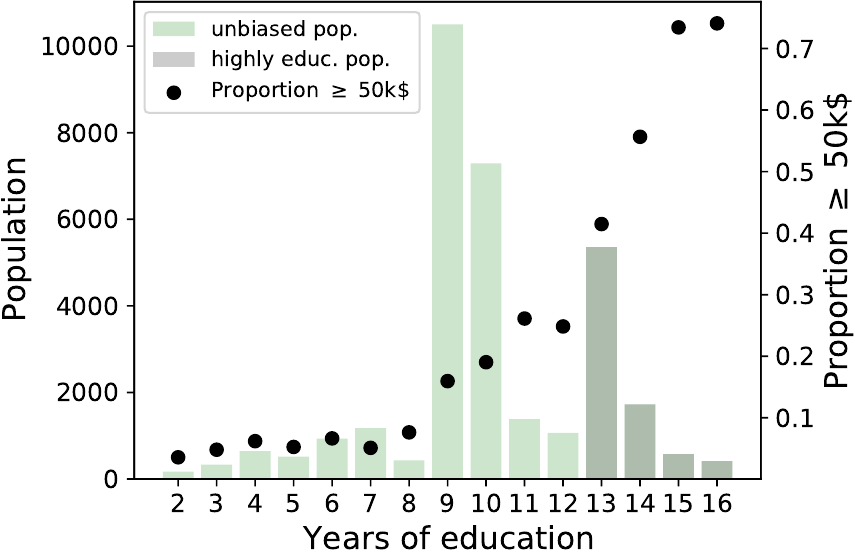}
\end{minipage}
\caption{Distribution of the house prices in the \emph{Boston} dataset (left). Distribution of the years of education in the \emph{Adult} dataset, against the proportion of people earning more than $50k\$$ yearly (right).}
\label{fig:expes}
\end{center}
\vspace{-0.5cm}
\end{figure*}

Consider first the \emph{Boston} housing dataset problem.
It is a regression problem, where one has to predict the price of a house on the range $[0, 50]$, based on $14$ attributes such as the number of rooms or statistics about the neighborhood.
One can easily imagine that such a dataset is actually composed of two samples: one dataset taken from a local estate agency, large but containing cheap houses as the neighborhood is not very trendy, and a second one, national and unbiased but smaller.
Of course, running ERM on the pooled sample without debiasing procedure should result in a global underestimation of the prices.
To replicate this framework, we have implemented the following protocol.
From the $500$ available observations, $100$ are kept for the testing phase.
From the remaining $400$ observations, two samples are extracted: a first one of size $200$, sampled among the cheapest houses, i.e., with prices lower than $22$ (see \Cref{fig:expes}), and a second unbiased of size $100$ (i.e., sampled uniformly at random).
The biasing functions are therefore $\omega_1(z) = \mathbb{I}\{y \le 22\}$, and $\omega_2(z) \equiv 1$.
Then, we have trained several ERM-based algorithms, namely Ridge Regression (RR), Support Vector Regressors (SVR), and Random Forest (RF), on the total sample of size $300$, with and without debiasing.
A third model is trained on the small unbiased sample only.
All algorithms have been run with several choices of hyperparameters around the default value.
Results in terms of Mean Square Error (MSE) on the test sample of size $100$, averaged over $100$ runs, are displayed in \Cref{tab:res} (top).
Except for SVR with very small regularization, the debiased procedure (db-ERM) outperforms standard ERM and ERM on the unbiased sample (ub-ERM).

\par Note that the previous example cannot be treated as a Covariate Shift (CS) problem, since sample bias applies to the output, breaking the CS assumption.
However, one might argue that biasing directly the output favors too much our procedure.
In the next example, we propose a binary classification problem where sample bias applies on the covariates.
Consider the machine learning problem associated to the \emph{Adult} dataset, also known as the \href{https://archive.ics.uci.edu/ml/datasets/adult}{\emph{Census Income} dataset}.
It is a binary classification task, where the goal is to predict whether a person's income exceeds $50,000$\$ a year, based on census data.
As can be seen in \Cref{fig:expes} (left), the proportion of persons having an income exceeding 50k\$ a year substantially depends on the number of years of education.
If highly educated people happen to be over-represented in the dataset (it is for instance more convenient to poll people concentrated in big cities, who have usually studied longer than people living in the countryside), it should deteriorate the predictions in absence of a debiasing procedure.
%
% Observe in addition that this setting cannot be cast as a CS problem, conditional laws being different from one education group to the other, see e.g., \Cref{fig:expes} (right) for salary with respect to age, by education group.
%
In order to highlight the interest of our debiasing procedure, we have implemented the following experimental protocol.
From the whole dataset, $1 500$ observations are kept for the testing phase.
From the rest are sampled two subgroups: one of 12+ years of education people of size $5 900$, and one unbiased (i.e., sampled uniformly from the entire population) of size $100$.
Then, logistic regression models (LogReg) and RFs are trained on the concatenation of the $6 000$ observations, with standard and debiased ERM, as well as on the small second sample of size $100$.
Numerical results are displayed in \Cref{tab:res} (bottom) in terms of test prediction scores.
Again, debiased ERM shows the best performances.
Another general comment that can be made is that the advantage brought by the debiasing decreases with the capacity of the model class considered (i.e., small $\lambda$, big $C$, or large number of trees).

\begin{table}[!t]
\begin{center}
\begin{tabular}{ll|c|c|c|}
                                 &                    & ERM                & db-ERM            & ub-ERM\\\toprule
\multirow{9}{*}{\vspace{-0.4cm}\textit{Boston}} & RR ($\lambda=0$)   & 27.41  $\pm$ 8.83  & \textbf{25.62 $\pm$ 6.63}  & 28.38 $\pm$ 7.99           \\
                                                & RR ($\lambda=0.1$) & 27.46  $\pm$ 8.90  & \textbf{25.59 $\pm$ 6.67}  & 28.11 $\pm$ 7.73           \\
                                                & RR ($\lambda=1$)   & 27.94  $\pm$ 9.25  & \textbf{25.72 $\pm$ 6.84}  & 28.05 $\pm$ 7.61           \\[0.2cm]
                                                & SVR ($C=0.1$)      & 99.63  $\pm$ 21.55 & \textbf{86.29 $\pm$ 18.79} & 86.60 $\pm$ 18.67          \\
                                                & SVR ($C=1$)        & 100.02 $\pm$ 21.87 & \textbf{85.66 $\pm$ 19.04} & 86.04 $\pm$ 18.67          \\
                                                & SVR ($C=10$)       & 97.27  $\pm$ 22.38 & 88.37 $\pm$ 21.68          & \textbf{82.35 $\pm$ 18.60} \\[0.2cm]
                                                & RF (trees=$10$)    & 19.83  $\pm$ 7.13  & \textbf{19.11 $\pm$ 6.97}  & 20.46 $\pm$ 6.50           \\
                                                & RF (trees=$100$)   & 18.20  $\pm$ 6.46  & \textbf{17.93 $\pm$ 6.58}  & 18.71 $\pm$ 6.10           \\
                                                & RF (trees=$1000$)  & 18.11  $\pm$ 6.61  & \textbf{17.69 $\pm$ 6.59}  & 18.54 $\pm$ 6.16           \\[0.1cm]\hline
&&&&\\[-0.1cm]
\multirow{9}{*}{\textit{Adult}\vspace{0.8cm}}   & LogReg ($C=0.1$)   & 63.87 $\pm$ 1.58 & \textbf{79.25 $\pm$ 1.67} & 78.24 $\pm$ 1.97 \\
                                                & LogReg ($C=1$)     & 63.81 $\pm$ 1.67 & \textbf{79.51 $\pm$ 1.80} & 77.79 $\pm$ 2.25 \\
                                                & LogReg ($C=10$)    & 63.87 $\pm$ 1.65 & \textbf{79.53 $\pm$ 1.78} & 78.01 $\pm$ 2.45 \\[0.2cm]
                                                & RF (trees=$10$)    & 39.00 $\pm$ 3.74 & \textbf{40.27 $\pm$ 4.16} & 18.48 $\pm$ 6.52 \\
                                                & RF (trees=$100$)   & 44.37 $\pm$ 3.28 & \textbf{45.36 $\pm$ 3.89} & 23.81 $\pm$ 5.71 \\
                                                & RF (trees=$1000$)  & 44.92 $\pm$ 3.24 & \textbf{46.03 $\pm$ 3.61} & 24.42 $\pm$ 5.51 \\\bottomrule
\end{tabular}
\end{center}
\caption{MSEs on \emph{Boston} and prediction scores on \emph{Adult}, averaged over 100 runs.}
\label{tab:res}
\end{table}

Hence, we have presented two learning examples, one regression task and one classification task, which cannot be tackled through ordinary CS (either bias applies to the target, or the conditional laws obviously change), empirically endorsing the soundness of our debiased ERM approach.
Additional experiments leading to similar conclusions are presented in \Cref{apx:expe}.
Notice finally that the code used to compute the debiasing weights is publicly available as a Python package at the following GitHub repository: \href{https://github.com/plaforgue/db_learn}{plaforgue/db\_learn}.

\section{Conclusion}\label{sec:conclusion}
In this article, we have provided a sound methodology to address selection bias issues in statistical learning.
We have extended the paradigmatic ERM approach to the situation where learning is based on several biased training samples.
In contrast to alternative techniques previously documented in the literature, the method proposed covers a wide range of sample bias scenarios, and applies to any ERM-like learning algorithm.
It relies on a preliminary debiasing of the raw empirical risk functional in the spirit of the procedure introduced in \cite{Vardi85} for cumulative distribution function estimation.
The nonasymptotic theoretical analysis carried out under mild assumptions shows that the rate achieved is the same as that attained in absence of any selection bias.
Numerical experiments are also documented, validating our theoretical findings.
A natural direction for future research is now to extend the statistical learning approach promoted in this article to situations were the biasing models at work are only partially known.

\appendix
%%%%%%%%%%%%%%%%%%%%%%%%%%%%%
%                           %
%     TECHNICAL DETAILS     %
%                           %
%%%%%%%%%%%%%%%%%%%%%%%%%%%%%

\section{Derivation of \Cref{eq:recover_omega}}
\label{apx:recover_omega}

Some computations, omitted in the core text for the sake of readability, are detailed below.
For all $k \le K$, we have:
\begin{align*}
\Omega_k &= \int \omega_k(z)dP(z)= \frac{\int \omega_k(z)dP(z)}{\int dP(z)}
= \frac{\int \omega_k(z) \left(\sum_{l=1}^K \frac{\lambda_l}{\Omega_l}\omega_l(z) \right)^{-1} d\bar{P}(z)}{\int \left(\sum_{l=1}^K \frac{\lambda_l}{\Omega_l}\omega_l(z) \right)^{-1} d\bar{P}(z)}\\
&= \frac{\int \omega_k(z) \left(\sum_{l=1}^K \frac{\lambda_l}{W^*_l}\omega_l(z) \right)^{-1} d\bar{P}(z)}{\int \left(\sum_{l=1}^K \frac{\lambda_l}{W^*_l}\omega_l(z) \right)^{-1} d\bar{P}(z)}\\
&= \frac{W^*_k}{\int \left(\sum_{l=1}^K \frac{\lambda_l}{W^*_l}\omega_l(z) \right)^{-1} d\bar{P}(z)},
\end{align*}
where we have successively used the fact that $\int dP = 1$, \Cref{eq:invert}, the fact that $\bW^* \propto \bm{\Omega}$ and \Cref{sys:true_omega}.

%%%%%%%%%%%%%%%%%%%%%%%%%%
%                        %
%     SIMPLE EXAMPLE     %
%                        %
%%%%%%%%%%%%%%%%%%%%%%%%%%

\section{A Simplistic Example}
\label{sec:simple_example}

Here, we exhibit a simple example where the training data samples are biased, but no system solving is required to form a debiased empirical distribution.
The flagship problem in supervised learning is multi-class classification and consists in the simplest situation, where $Z=(X,Y)$, $Y$ being a discrete random variable valued in $\{1,\; \ldots,\; Q  \}$ with $Q\geq 1$ say, and the r.v. $X$ takes its values in a measurable space $\mathcal{X}$ and models some information hopefully useful to predict $Y$.
The parameter space $\Theta$ is a set $\mathcal{G}$ of measurable mappings (i.e., classifiers) $g:\mathcal{X}\to \{1,\; \ldots,\; Q\}$ and the loss function is given by $\ell(g,\; (x,y))=\mathbb{I}\{ g(x) \neq y  \}$ for all $g$ in $\mathcal{G}$ and any $(x,y)\in \mathcal{X}\times \{1,\; \ldots,\; Q \}$.
The distribution $P$ of the random pair $(X,Y)$ can be either described by $X$'s marginal distribution $\mu$ and the posterior probability $\eta(x)=(\eta_1(x),\; \ldots,\; \eta_Q(x))$, where $\eta_q(x)=\mathbb{P}\{Y=q\mid X=x  \}$ for $q\in\{1,\; \ldots,\; Q  \}$, or else by the $((p_1, F_1),\; \ldots,\; (p_Q,F_Q))$ where $p_q=\mathbb{P}\{Y=q  \}$ and $F_{q}$ is $X$'s conditional distribution given $Y=q$ with $q\in\{1,\; \ldots,\; Q\}$.
Observe that $p_1+\ldots+p_Q=1$, we assume that $p_q\in(0,1)$ for all $q\in\{1,\; \ldots,\; Q  \}$.
It is very common that the class probabilities in the training datasets are significantly different from those in the test stage, the $p_q$'s namely.
We thus consider the case where, for all $k\in\{1,\; \ldots,\; K \}$, the distribution $P_k$ of the $k$-th training dataset $\mathcal{D}_k=\{ (X_{k,1},Y_{k,1}),\; \ldots,\, (X_{k,n_k},Y_{k,n_k})\}$ is described by $((p_{k,1}, F_1),\; \ldots,\; (p_{k,Q},F_Q))$, where the vector of class probabilities $\bm{p}_k=(p_{k,1},\; \ldots,\; p_{k,Q})\in[0,1]^Q$ (note incidentally that $p_{k,1}+ \ldots+p_{k,Q}=1$) may differ from $\bm{p}=(p_1,\; \ldots,\; p_Q)$.
We point out that it may happen that certain class probabilities $p_{k,q}$ are equal to zero, so that some labels cannot be observed among certain data samples.
The likelihood function takes the form
$$
\forall (x,y)\in \mathcal{X}\times\{1,\; \ldots,\; Q\},\;\; \frac{dP_k}{dP}(x,y)=\sum_{q=1}^Q \mathbb{I}\{ y=q \}(p_{k,q}/p_q),
$$
which reveals that it depends on the label $y$ solely.
Hence, in this very simple case, we have $\omega_k(x,y)=p_{k,y}/p_y$ and $\Omega_k=1$ for all $(y,k)\in\{1,\; \ldots,\; Q\}\times \{1,\; \ldots,\; K \}$ and there is no need for solving any system to compute a nearly debiased empirical distribution.
Observe that, for all $\kappa>0$, vertices $k$ and $l$ in $\{1,\; \ldots,\; K\}$ are connected in the graph $G_{\kappa}$ iff $\sum_{q\leq Q}p_{k,q}p_{l,q}/p_q\geq \kappa$.
%
% In addition, the upper bound condition in Assumption \ref{hyp:bounded_omega} is always fulfilled with $M=1/\min_qp_q<+\infty$ and the lower bound condition is satisfied as soon as $0<\varepsilon\leq \min_{q}p_{k_q,q}/p_q$, where $k_q\in\arg\max_{k\in\{1,\; \ldots,\; K \}}p_{k,q}$ for all $q\in\{1,\; \ldots,\; Q  \}$.
%
However, in this situation the biasing functions $\omega_k$ can be directly estimated from the data samples, replacing $p_{k,q}$ by $n_{k,q}/n_k$ with $n_{k,q}=\sum_{i=1}^{n_k}\mathbb{I}\{Y_{k,i}=q  \}$ for all $(k,q)\in \{1,\; \ldots,\; K \} \times \{1,\; \ldots,\; Q\}$ and computing
$$
\widehat{\omega}_k(x,y)=\widehat{p}_{k,y}/p_y,
$$
for all $(y,k)\in\{1,\; \ldots,\; Q\}\times \{1,\; \ldots,\; K \}$.
One may then consider the estimator
$$
\left(\sum_{k=1}^K \frac{n_k}{n}\widehat{\omega}_k(x,y) \right)^{-1} \widehat{P}_n
$$
of the distribution $P$.

\section{Technical Proofs}
\label{sec:proof_details}

We now provide the technical proofs of the results stated in the paper.
Recall that for notation simplicity we used universal constants $M, M', c, c', \gamma, n_0$, in the core text.
For the sake of clarity, we now index them by propositions, such that $M_i, M'_i, c_i, c'_i, \gamma_i, n_{0, i}$ correspond to $M, M', c, c', \gamma, n_0$ for Proposition $i$.

%%%%%%%%%%%%%%%%%%%%%%%%%%%
%                         %
%     PROOF OF PROP 1     %
%                         %
%%%%%%%%%%%%%%%%%%%%%%%%%%%

\subsection{Proof of \Cref{prop:unique_bounded}}
\label{apx:proof_unique_bounded}

First, we introduce the following notation.
Let $\bar{D}$ and $\widehat{D}_n$ be the two functions from $\mathbb{R}^K$ to $\mathbb{R}$ such that for all $\bu = (u_1, \ldots, u_K) \in \mathbb{R}^K$:
\begin{align*}
\bar{D}(\bu) &= \int \log\left[ \sum_{l=1}^K e^{u_l} \omega_l(z)\right]d\bar{P}(z) - \sum_{l=1}^K \lambda_l u_l,\\[0.2cm]
\widehat{D}_n(\bu) &= \int \log\left[ \sum_{l=1}^K e^{u_l} \omega_l(z)\right]d\widehat{P}_n(z) - \sum_{l=1}^K \hat{\lambda}_l u_l.
\end{align*}
Let $\bu^* = \argmin_{\bu} \bar{D}(\bu)$, and similarly $\hat{\bu}_n = \argmin_{\bu} \widehat{D}_n(\bu)$.
We also compute the gradients $\bar{D}'$, $\widehat{D}_n'$ and the Hessian matrices $\bar{D}''$, $\widehat{D}_n''$, of these two smooth functions. For all $ \bu = (u_1, \ldots, u_K) \in \mathbb{R}^K$ and all $k, k' \le K$, we have:
\begin{align*}
\left[\bar{D}'(\bu)\right]_k &= \int \frac{e^{u_k} \omega_k(z)}{\sum_{l=1}^K e^{u_l} \omega_l(z)}d\bar{P}(z) - \lambda_k,\\[0.2cm]
\left[\widehat{D}'_n(\bu)\right]_k &= \int \frac{e^{u_k} \omega_k(z)}{\sum_{l=1}^K e^{u_l} \omega_l(z)}d\widehat{P}_n(z) - \hat{\lambda}_k,\\[0.2cm]
\left[\bar{D}''(\bu)\right]_{k, k'} &= \int \left[\frac{e^{u_k} \omega_k(z) \delta_{kk'}}{\sum_{l=1}^K e^{u_l} \omega_l(z)} - \frac{e^{u_k} \omega_k(z) e^{u_{k'}} \omega_{k'}(z)}{\left(\sum_{l=1}^K e^{u_l} \omega_l(z)\right)^2}\right]d\bar{P}(z),\\[0.2cm]
\left[\widehat{D}''_n(\bu)\right]_{k, k'} &= \int \left[\frac{e^{u_k} \omega_k(z) \delta_{kk'}}{\sum_{l=1}^K e^{u_l} \omega_l(z)} - \frac{e^{u_k} \omega_k(z) e^{u_{k'}} \omega_{k'}(z)}{\left(\sum_{l=1}^K e^{u_l} \omega_l(z)\right)^2}\right]d\widehat{P}_n(z).
\end{align*}
Observe now that Systems \eqref{sys:true_omega} and \eqref{sys:emp_omega} are equivalent to $\bar{D}'(\bu^*) = \bm{0}$ and $\widehat{D}'_n(\hat{\bu}_n) = \bm{0}$ respectively, with $\bu^* = \log(\bm{\lambda} / \bW^*)$ and $\hat{\bu}_n = \log(\hat{\bm{\lambda}} / \widehat{\bW}_n)$, where the division and the logarithm are meant componentwise.
Recall that Systems \eqref{sys:true_omega} and \eqref{sys:emp_omega} are homogeneous of degree $0$.
Equivalently, $\bar{D}$ and $\widehat{D}_n$ are invariant under translation of vectors that are colinear with $\bm{1}$, i.e., $\bar{D}(\bu + c\bm{1}) = \bar{D}(\bu)$ and $\widehat{D}_n(\bu + c\bm{1}) = \widehat{D}_n(\bu)$ for all $\bu \in \mathbb{R}^K$ and $c \in \mathbb{R}$.
To ensure uniqueness of the solutions to Systems \eqref{sys:true_omega} and \eqref{sys:emp_omega}, we consider $\bW^*$ and $\widehat{\bW}_n$ such that $\max_{k \le K} W^*_k / \lambda_k = 1$, and $\max_{k \le K} \widehat{W}_{n, k} / \hat{\lambda}_k = 1$.
In terms of $\bu^*$ and $\hat{\bu}_n$, this normalization writes $\min_{k \le K} u^*_k = \min_{k \le K} \hat{u}_{n, k} = 0$.
We now show that there exists $\rho > 0$ such that for all $k \le K$ it holds:
\begin{equation*}
\rho \le \widehat{W}_{n, k} \le 1 \qquad \text{and} \qquad \rho \le W^*_k \le 1.
\end{equation*}
The upper bounds above are immediate, insofar as for, all $k \le K$, we have:
\begin{equation*}
\widehat{W}_{n, k} \le \hat{\lambda}_k \le 1 \qquad \text{and similarly} \qquad W^*_k \le \lambda_k \le 1.
\end{equation*}
To derive $\rho$, we show that there exist $U > 0$ such that:
\begin{equation*}
\forall k \le K, \qquad \hat{u}_{n, k} \le U \qquad \text{and} \qquad  u^*_k \le U.
\end{equation*}
The proof mechanism is as follows.
First, we derive a lower bound of $\widehat{D}_n(\bu)$, that depends linearly on $u_{k_0}$ and $u_{k_1}$, for any couple $(k_0, k_1)$ being an edge in $\widehat{G}_n$.
Next, we apply this lower bound at point $\hat{\bu}_n$, with $k_0$ such that $\hat{u}_{n, k_0} = 0$ (such an index exists by the normalization we impose).
Combining this lower bound with the observation that $\widehat{D}_n(\hat{\bu}_n) \le \widehat{D}_n(\bm{0})$, we obtain an upper bound on $\hat{u}_{n, k_1}$.
Finally, this approach is used recursively to bound the neighbors of $k_1$, and so on and so forth.
The graph $\widehat{G}_n$ being connected by the first claim of \Cref{prop:unique_bounded}, every component $\hat{u}_{n, k}$ is attained after at most $K-1$ iterations.

\noindent Let $(k_0, k_1)$ be an edge in $\widehat{G}_n$.
Using the definition of $\widehat{P}_n$ it holds:
\begin{align*}
\widehat{D}_n(\bu) &= \int \log\left[ \sum_{l=1}^K e^{u_l} \omega_l(z)\right]d\widehat{P}_n(z) - \sum_{l=1}^K \hat{\lambda}_l u_l\\
&= \sum_{k=1}^K \hat{\lambda}_k \left(\int\log\left[ \sum_{l=1}^K e^{u_l} \omega_l(z)\right]d\widehat{P}_k(z) - u_k\right).
\end{align*}
For $k \ne k_0$, it holds:
\begin{equation*}
\int\log\left[ \sum_{l=1}^K e^{u_l} \omega_l(z)\right]d\widehat{P}_k(z) - u_k \ge \int\log(e^{u_k} \varepsilon) d\widehat{P}_k(z) - u_k = \log(\varepsilon).
\end{equation*}
For $k=k_0$, we have:
\begin{align*}
\int\log\left[ \sum_{l=1}^K e^{u_l} \omega_l(z)\right]d\widehat{P}_{k_0}(z) \ge &\int\log(\varepsilon)(1 - \mathbb{I}\{\omega_{k_1}(z) > 0\})d\widehat{P}_{k_0}(z)\\
&+ \int\log(e^{u_{k_1}}\varepsilon)\mathbb{I}\{\omega_{k_1}(z) > 0\}d\widehat{P}_{k_0}(z)\\
=& ~\log(\varepsilon) + u_{k_1} \int \mathbb{I}\{\omega_{k_1}(z) > 0\}d\widehat{P}_{k_0}(z).
\end{align*}
From the proof of the first claim of \Cref{prop:unique_bounded} (see \Cref{eq:sufficient_edge}), we know that it holds:
\begin{equation}
\int \mathbb{I}\{\omega_{k_1}(z) > 0\}d\widehat{P}_{k_0}(z) \ge \frac{\kappa \varepsilon}{2}\label{eq:kappa}
\end{equation}
so that one gets:
\begin{align}
\widehat{D}_n(\bu) &\ge (1 - \hat{\lambda}_{k_0}) \log(\varepsilon) + \hat{\lambda}_{k_0}\left(\log(\varepsilon) + \frac{\kappa \varepsilon}{2} u_{k_1} - u_{k_0}\right)\nonumber\\
&\ge \log(\varepsilon) + \frac{\underline{\lambda}\kappa \varepsilon}{2}u_{k_1} - u_{k_0}.\label{eq:lower_bound}
\end{align}
Observe also that we have:
\begin{equation}\label{eq:up_bound}
\widehat{D}_n(\hat{\bu}_n) \le \widehat{D}_n(\bm{0}) = \int \log\left[ \sum_{l=1}^K \omega_l(z)\right]d\widehat{P}_n(z) \le \log(K).
\end{equation}
Combining \Cref{eq:lower_bound} evaluated at point $\hat{\bu}_n$ and \Cref{eq:up_bound}, we obtain:
\begin{equation}\label{eq:transition}
\hat{u}_{n, k_1} \le \frac{2(\log(K/\varepsilon) + \hat{u}_{n, k_0})}{\underline{\lambda}\kappa \varepsilon}.
\end{equation}
The last step consists in extending this bound to every $\hat{u}_{n, k}$.
To do so, we first set (without loss of generality) $\hat{u}_{n, k_0} = \min_{k \le K} \hat{u}_{n, k} = 0$.
Recall also the definition of graph $G_\text{min}$, as introduced in the proof of the first claim of \Cref{prop:unique_bounded}.
We can then apply \Cref{eq:transition} to all $k_1$ that are neighbors of $k_0$ in $G_\text{min}$.
Next, notice that this method can be used in a recursive fashion, with now the $k_1$ as anchor points.
Eventually, every $\hat{u}_{n, k}$ is attained, as $G_\text{min}$ is connected.
\Cref{eq:transition} becoming looser and looser as it is applied, the last question is \textit{how many recursive steps are required?}
The minimum number of recursive steps needed is the biggest (among $k \le K$) shortest path (in $G_\text{min}$) between $k_0$ and $k$, denoted $\text{diam}(G_\text{min}, k_0)$.
Combining all the arguments, we get:
\begin{align*}
\forall k \le K, \qquad \hat{u}_{n, k} &\le \frac{\left(\frac{2}{\underline{\lambda}\kappa \varepsilon}\right)^{\text{diam}(G_\text{min}, k_0) + 1} - 1}{\frac{2}{\underline{\lambda}\kappa \varepsilon} - 1}\log(K/\varepsilon)\\[0.5cm]
&\le \frac{\left(\frac{2}{\underline{\lambda}\kappa \varepsilon}\right)^K - 1}{\frac{2}{\underline{\lambda}\kappa \varepsilon} - 1}\log(K/\varepsilon).
\end{align*}
Therefore, for all $k \le K$ it holds:
\begin{equation}\label{eq:def_Wmin}
\widehat{W}_{n, k} = \hat{\lambda}_k e^{-\hat{u}_{n, k}} \ge \underline{\lambda}e^{-U} \coloneqq \rho
\end{equation}
with
\begin{equation}\label{eq:def_M}
U =  \frac{\left(\frac{2}{\underline{\lambda}\kappa \varepsilon}\right)^K - 1}{\frac{2}{\underline{\lambda}\kappa \varepsilon} - 1}\log(K/\varepsilon).
\end{equation}
Finally, note that the exact same method can be applied to $\bu^*$, by substituting $\widehat{P}_k$ with $P_k$ in the computations.
\hfill\qed

%%%%%%%%%%%%%%%%%%%%%%%%%%%
%                         %
%     PROOF OF PROP 2     %
%                         %
%%%%%%%%%%%%%%%%%%%%%%%%%%%

\subsection{Proof of \Cref{prop:dev_W}}
\label{apx:proof_dev_W}

First, we prove the following lemma, ensuring that the deviation $\big\|\widehat{\bW}_n - \bW^*\big\|_2$ is upper bounded by the deviation $\left\|\hat{\bu}_n - \bu^*\right\|_2$.

\begin{lemma}\label{lem:lipschitz}
Suppose that Assumption \ref{hyp:rates} is satisfied. Then it holds:
\begin{equation*}
\left\|\widehat{\bW}_n - \bW^*\right\|_2 \le \left\|\hat{\bu}_n - \bu^*\right\|_2 + C_\lambda\sqrt{\frac{K}{n}}.
\end{equation*}
\end{lemma}

\begin{proof}
For all $k \le K$ it holds:
\begin{align*}
\left|\widehat{W}_{n, k} - W^*_k\right| &= \left|\hat{\lambda}_k e^{- \hat{u}_{n, k} } - \lambda_k e^{-u^*_k}\right|\\
&\le \left|\hat{\lambda}_k e^{- \hat{u}_{n, k} } - \hat{\lambda}_k e^{-u^*_k}\right| + \left|\hat{\lambda}_k e^{- u^*_k } - \lambda_k e^{-u^*_k}\right|\\
&\le \left|e^{- \hat{u}_{n, k} } - e^{-u^*_k}\right| + \left|\hat{\lambda}_k - \lambda_k\right|\\
&\le |\hat{u}_{n, k} - u^*_k| + \frac{C_\lambda}{\sqrt{n}}
\end{align*}
where we have used the definition of $\bu^*$ and $\hat{\bu}_n$, the triangle inequality, the fact that $\hat{\lambda}_k \le 1$, and that $u^*_k \ge 0$, the mean value theorem on $u \mapsto e^{-u}$ with $\hat{u}_{n, k} \ge 0$, and Assumption \ref{hyp:rates}.
Applying again the triangle inequality finally yields:
\begin{align*}
\left\|\widehat{\bW}_n - \bW^*\right\|_2 &= \sqrt{\sum_{k=1}^K \left|\hat{W}_{n, k} - W^*_k\right|^2}\\[0.2cm]
&\le \sqrt{\sum_{k=1}^K \left(\left|\hat{u}_{n, k} - u^*_k\right| + \frac{C_\lambda}{\sqrt{n}}\right)^2}\\[0.2cm]
&= \left\|\left|\hat{\bu}_n - \bu^*\right| + \frac{C_\lambda}{\sqrt{n}}\bm{1}\right\|_2\\[0.2cm]
&\le \left\|\hat{\bu}_n - \bu^*\right\|_2 + C_\lambda\sqrt{\frac{K}{n}}.
\end{align*}
\end{proof}

\noindent Next, we show that \Cref{prop:unique_bounded} allows to bound the deviation $\|\hat{\bu}_n - \bu^*\|_2$ in terms of the deviation $\left\|\widehat{D}'_n(\bu^*) - \bar{D}'(\bu^*)\right\|_2$.
\begin{proposition}\label{prop:strong_cvx}
Suppose that Assumptions \ref{hyp:rates}, \ref{hyp:connect}, \ref{hyp:bounded_omega}, and \ref{asm:eigenvalue} are satisfied.
Then, there exist $M_4, c_4, n_{0, 4}, L$, depending only on $K, C_\lambda, \underline{\lambda}, \kappa, \varepsilon, \sigma$, such that for all $n \ge n_{0, 4}$ it holds with probability at least $1 - M_4\exp(-c_4n)$:
\begin{equation*}
\left\|\hat{\bu}_n - \bu^*\right\|_2 \le L \left\|\widehat{D}'_n(\bu^*) - \bar{D}'(\bu^*)\right\|_2.
\end{equation*}
\end{proposition}

\begin{proof}
Define $F: [0, 1] \rightarrow \mathbb{R}^K$ such that $F(t) = \widehat{D}'_n\left(\hat{\bu}_n + t(\bu^* - \hat{\bu}_n)\right)$.
It holds:
\begin{align}
F(1) - F(0) &= \left( \int_0^1 F'(t) dt\right)\nonumber\\[0.1cm]
\widehat{D}'_n(\bu^*) - \widehat{D}'_n(\hat{\bu}_n) &= \left( \int_0^1 \left[\widehat{D}''_n(\hat{\bu}_n + t(\bu^* - \hat{\bu}_n))\right](\bu^* - \hat{\bu}_n) dt\right)\nonumber\\[0.1cm]
\widehat{D}'_n(\bu^*) - \bar{D}'(\bu^*) &= \left( \int_0^1 \left[\widehat{D}''_n(\hat{\bu}_n + t(\bu^* - \hat{\bu}_n))\right]dt \right)(\bu^* - \hat{\bu}_n)\label{eq:integral}
\end{align}
where the integral over matrices must be understood componentwise.
The key point to relate $\big\|\widehat{D}'_n(\bu^*) - \bar{D}'(\bu^*)\big\|_2$ to $\|\hat{\bu}_n - \bu^*\|_2$ is then to study the smallest eigenvalues of $\int_0^1 \big[\widehat{D}''_n(\hat{\bu}_n + t(\bu^* - \hat{\bu}_n))\big]dt$.
From the definition of $\widehat{D}_n''$, one can see that its smallest eigenvalue is $0$, associated to $\bm{1}$.
Hopefully, thanks to the normalization, $\hat{\bu}_n - \bu^*$ is not collinear to $\bm{1}$ unless $\hat{\bu}_n = \bu^*$.
Let $\hat{\bu}_n - \bu^* = c\bm{1} + \bm{w}$ be the decomposition of $\hat{\bu}_n - \bu^*$ on $\mathrm{Span}(\bm{1}) \otimes \mathrm{Span}(\bm{1})^\perp$, such that $\bm{1}^\top \bm{w} = 0$.
One can check that $\|\bm{w}\|_\infty \ge \|c\bm{1}\|_\infty$, so that it holds
\begin{equation}\label{eq:norm_ineq}
\|\hat{\bu}_n - \bu^*\|_2^2 = \|\bw\|_2^2 + \|c\bm{1}\|_2^2 \le \|\bw\|_2^2 + K\|c\bm{1}\|_\infty^2 \le (K+1)\|\bw\|_2^2.
\end{equation}
Combining \Cref{eq:integral,eq:norm_ineq}, one gets
\begin{align*}
\left\|\widehat{D}'_n(\bu^*) - \bar{D}'(\bu^*)\right\|_2 &\ge \sigma_2\left( \int_0^1 \left[\widehat{D}''_n(\hat{\bu}_n + t(\bu^* - \hat{\bu}_n))\right]dt\right) \left\|\bw\right\|_2\\
&\ge \inf_{\bv \in [\hat{\bu}_n, \bu^*]}\sigma_2\left(\widehat{D}''_n(\bv)\right) \frac{\|\hat{\bm{u}}_n - \bu^*\|_2}{\sqrt{K + 1}}
\end{align*}
where $\sigma_2(A)$ denotes the second smallest eigenvalue of a matrix $A$.
We have now to find a lower bound of $\inf_{\bv \in [\hat{\bu}_n, \bu^*]}\sigma_2\big(\widehat{D}''_n(\bv)\big)$.
Let $\bv \in [\hat{\bu}_n, \bu^*]$.
For notation simplicity, we omit the $\bv$ in the following, and use $\widehat{D}''_n$ and $\bar{D}''$.
It holds:
\begin{align*}
\sigma_2(\widehat{D}''_n) &= \inf_{\substack{\|\bu\| = 1,\\ \bm{1}^\top \bu = 0}} \bu^\top \widehat{D}''_n \bu\\[0.2cm]
&= \inf_{\substack{\|\bu\| = 1,\\ \bm{1}^\top \bu = 0}} \bu^\top (\widehat{D}''_n - \bar{D}'')\bu + \bu^\top \bar{D}'' \bu\\[0.2cm]
&\ge -\|\widehat{D}''_n - \bar{D}''\|_{\sigma_\infty} + \sigma_2(\bar{D}'')
\end{align*}
where $\|A\|_{\sigma_\infty}$ denotes the Schatten $\infty$-norm such that $\|A\|_{\sigma_\infty} = \|\sigma(A)\|_\infty$, with $\sigma(A)$ the vector of singular values of a matrix $A$.
For $A \in \mathbb{R}^{K \times K}$ it holds:
\begin{equation*}
\|A\|_{\sigma_\infty} \le \|A\|_\text{Fr} \le K \sup_{i, j} A_{i, j}
\end{equation*}
so that we get:
\begin{equation*}
\sigma_2(\widehat{D}''_n) \ge \sigma_2(\bar{D}'') - K \sup_{i, j} \big|[\widehat{D}''_n]_{i, j} - [\bar{D}'']_{i, j}\big|.
\end{equation*}
Now, define the compact set $\mathcal{U} = [0, U]^K$, with $U$ defined in \Cref{eq:def_M}.
We know from \Cref{prop:unique_bounded} that with probability at least $1 - M_1\exp(-c_1n)$ both $\hat{\bu}_n$ and $\bu^*$ belong to $\mathcal{U}$, so that $[\hat{\bu}_n, \bu^*] \subset \mathcal{U}$.
We can then use \Cref{asm:eigenvalue} to lower bound $\sigma_2(\bar{D}''(\bv))$ by $\sigma > 0$ uniformly on $[\hat{\bu}_n, \bu^*]$.
% Let $\sigma_2^*(\mathcal{U})$ be the infimum of $\sigma_2(\bar{D}''(\bv))$ for $\bv \in \mathcal{U}$.
% %
% The second smallest eigenvalue of a matrix $A$ being a continuous function of $A$, we know that this infimum is attained over the compact $\mathcal{U}$, and that it is strictly positive.

Focus now on the term $K \sup_{i, j} \big|[\widehat{D}''_n]_{i, j} - [\bar{D}'']_{i, j}\big|$.
From the definition of $\widehat{D}''_n$ and $\bar{D}''$, we can see that their entries $(k, k')$ are the integrals of some function comprised in $[-1, 1]$, according to $\widehat{P}_n$ and $\bar{P}$ respectively.
For all $i, j \le K$, \Cref{cor:Hoeffding} gives that for all $t > 0$ and $n \ge 2\log(2K)/(\underline{\lambda}t^2)$ it holds with probability at least $1 - 2K\exp\left(-\frac{\underline{\lambda}nt^2}{2}\right)$:
\begin{equation*}
\big|[\widehat{D}''_n]_{i, j} - [\bar{D}'']_{i, j}\big| \le \frac{C_\lambda K}{\sqrt{n}} + t.
\end{equation*}
The union bound then gives that with probability $1 - 2K^3\exp\left(-\frac{\underline{\lambda}nt^2}{2}\right)$ it holds:
\begin{equation*}
K \sup_{i, j} \big|[\widehat{D}''_n]_{i, j} - [\bar{D}'']_{i, j}\big| \le \frac{C_\lambda K^2}{\sqrt{n}} + K t.
\end{equation*}
Thus, for $n \ge \max\left(\frac{16 C_\lambda^2 K^4}{{\sigma}^2}, \frac{96K^2}{\underline{\lambda}{\sigma}^2}\log(2K)\right)$, it holds with probability at least $1 - 2K^3 \exp\left(-\frac{\underline{\lambda} {\sigma}^2}{32 K^2} n\right)$:
\begin{equation*}
K \sup_{i, j} \big|[\widehat{D}''_n]_{i, j} - [\bar{D}'']_{i, j}\big| \le \frac{\sigma}{4} + \frac{\sigma}{4} = \frac{\sigma}{2}
\end{equation*}
and consequently
\begin{equation*}
\sigma_2\left(\widehat{D}''_n(\bv)\right) \ge \frac{\sigma}{2}.
\end{equation*}
The last step consists in extending this bound uniformly over the line segment $[\hat{\bu}_n, \bu^*]$.
To do so, we adopt an entropic point of view: we cover the set $\mathcal{U}$ (in which the line segment $[\hat{\bu}_n, \bu^*]$ is contained with high probability) with balls, apply the union bound for the centers of these balls, and show that within a ball, the second smallest eigenvalue is relatively stable.
%
% The line segment $[\hat{\bu}_n, \bu^*]$ being contained in $\mathcal{U}$, we know that it is of length $U$ at most, with respect to $\|\cdot\|_\infty$.
%
By definition, note that $\mathcal{U}$ can be covered with $\mathcal{N}_\epsilon = U^K/(2\epsilon)^K$ $\|\cdot\|_\infty$-balls of radius $\epsilon$.
Now, let $(\bu, \bv) \in \mathcal{U}^2$ such that $\|\bu - \bv\|_\infty \le \epsilon$.
What is the value of $\big|\sigma_2(\widehat{D}''_n(\bu)) - \sigma_2(\widehat{D}''_n(\bv))\big|$?
As noticed in \cite{GVW88}, for any $\ba \in \mathbb{R}^K$ it holds:
\begin{equation*}
\ba^\top \widehat{D}''_n(\bu) \ba = \int_z~ \sum_{k=1}^K p_k(z)a_k^2 - \left(\sum_{k=1}^K a_k p_k(z)\right)^2 d\widehat{P}_n(z)
\end{equation*}
with $p_k(z) = e^{u_k} \omega_k(z) / \sum_{l=1}^K e^{u_l} \omega_l(z)$.
Define $q_k(z) = e^{v_k} \omega_k(z) / \sum_{l=1}^K e^{v_l} \omega_l(z)$, and assume that $\|\ba\|_2 = 1$.
It holds:
\begin{align*}
&\left|\ba^\top \widehat{D}''_n(\bu) \ba - \ba^\top \widehat{D}''_n(\bv) \ba\right|\\
=&\left|\int~\sum_{k=1}^K p_k(z)a_k^2 - \left(\sum_{k=1}^K a_k p_k(z)\right)^2 - \sum_{k=1}^K q_k(z)a_k^2 + \left(\sum_{k=1}^K a_k q_k(z)\right)^2 d\widehat{P}_n(z)\right|\\[0.2cm]
\le& \int \sum_{k=1}^K \big|p_k(z) - q_k(z)\big|a_k^2~d\widehat{P}_n(z)\\
&+ \int \left|\sum_{k=1}^K(p_k(z) + q_k(z))a_k\right|\cdot \left|\sum_{k=1}^K(p_k(z) - q_k(z))a_k\right|d\widehat{P}_n(z)\\[0.2cm]
\le&~ \int \|\bm{p}(z) - \bm{q}(z)\|_\infty~d\widehat{P}_n(z) + \int \|\bm{p}(z) + \bm{q}(z)\|_2 \cdot \|\bm{p}(z) - \bm{q}(z)\|_2~d\widehat{P}_n(z)\\[0.2cm]
\le&~ 3\sqrt{K} \int\|\bm{p}(z) - \bm{q}(z)\|_2~d\widehat{P}_n(z).
\end{align*}
Furthermore, notice that $\bm{p}(z)$ is exactly the integrand in $\widehat{D}''_n(\bu)$, while $\bm{q}(z)$ is the integrand in $\widehat{D}''_n(\bv)$.
Using the same integral calculus as in the beginning of the proof, and bounding the biggest eigenvalue of the matrices by $K$ (as it is an upper bound of the trace), we get that for all $z$ it holds $\|\bm{p}(z) - \bm{q}(z)\|_2 \le K \|\bu - \bv\|_2$.
Therefore, we get for all $\ba \in \mathbb{R}^K$ such that $\|\ba\|_2 = 1$:
\begin{equation*}
\left|\ba^\top \widehat{D}''_n(\bu) \ba - \ba^\top \widehat{D}''_n(\bv) \ba\right| \le 3K^2 \|\bu - \bv\|_\infty,
\end{equation*}
and consequently
\begin{equation*}
\left|\sigma_2\left(\widehat{D}''_n(\bu)\right) - \sigma_2\left(\widehat{D}''_n(\bv)\right)\right| \le 3K^2 \|\bu - \bv\|_\infty.
\end{equation*}
Now, let $(\bu_1, \ldots, \bu_{\mathcal{N}_\epsilon})$ be an $\epsilon$-coverage of $\mathcal{U}$.
% the line segment $[\hat{\bu}_n, \bu^*]$.
%
Applying the union bound, we get that with probability at least $1 - \frac{2K^3U^K}{(2\epsilon)^K}\exp\left(-\frac{\underline{\lambda}\sigma^2}{32K^2}n\right)$ for any $i \le \mathcal{N_\epsilon}$ it holds:
\begin{equation*}
\sigma_2\left(\widehat{D}''_n(\bu_i)\right) \ge \frac{\sigma}{2}.
\end{equation*}
Let $\bv \in [\hat{\bu}_n, \bu^*] \subset \mathcal{U}$.
By definition, there exists $i \le \mathcal{N}_\epsilon$ such that $\|\bv - \bu_i\|_\infty \le \epsilon$.
Therefore, we get:
\begin{equation*}
\sigma_2\left(\widehat{D}''_n(\bv)\right) \ge \sigma_2\left(\widehat{D}''_n(\bu_i)\right) - 3K^2 \epsilon.
\end{equation*}
Taking the infimum, we have with probability $1 - \frac{2K^3U^K}{(2\epsilon)^K}\exp\left(-\frac{\underline{\lambda}\sigma^2}{32K^2}n\right)$
\begin{equation*}
\inf_{\bv \in [\hat{\bu}_n, \bu^*]}~\sigma_2\left(\widehat{D}''_n(\bv)\right) \ge \frac{\sigma}{2} - 3K^2 \epsilon.
\end{equation*}
Choosing $\epsilon = \frac{\sigma}{12 K^2}$, we have with probability $1 - 2K^{2K+3}\left(\frac{6U}{\sigma}\right)^K\exp\left(-\frac{\underline{\lambda}\sigma^2}{32K^2}n\right)$
\begin{equation*}
\inf_{\bv \in [\hat{\bu}_n, \bu^*]}~\sigma_2\left(\widehat{D}''_n(\bv)\right) \ge \frac{\sigma}{4}.
\end{equation*}
Collecting all arguments, for all $n \ge 16C_\lambda^2K^4/\sigma^2$ it holds with probability \mbox{$1 - M_1\exp(-c_1 n) - 2K^{2K+3}\left(\frac{6U}{\sigma}\right)^K\exp\left(-\frac{\underline{\lambda}\sigma^2}{32K^2}n\right)$}:
\begin{equation*}
\left\|\widehat{D}'_n(\bu^*) - \bar{D}'(\bu^*)\right\|_2 \ge \frac{\sigma}{4\sqrt{K + 1}}~\|\hat{\bm{u}}_n - \bu^*\|_2.
\end{equation*}
The proof is finally concluded by setting $M_4 = 2 \max\big(M_1;\, 2K^{2K+3}\left(\frac{6U}{\sigma}\right)^K\big)$, $c_4 = \min\big(c_1;\, \underline{\lambda} \sigma^2/(32 K^2)\big)$, $n_{0, 4} = \max\big(16C_\lambda^2K^4/\sigma^2;\, \log(M_4/c_4)\big)$, and $L = 4\sqrt{K+1}/\sigma$.
\end{proof}

\noindent The following key lemma allows to decompose the deviation $\big| \int \hat{h}_n d\widehat{P}_n - \int h d\bar{P} \big|$ into different pieces that are more easily controllable.
It is used for instance to bound $\big|\widetilde{L}_n(\theta) - L(\theta)\big|$, see \Cref{eq:decompo_text}.

\begin{lemma}\label{lem:decompo}
Let $\hat{h}_n : \mathcal{Z} \rightarrow \mathbb{R}$, $h : \mathcal{Z} \rightarrow \mathbb{R}$ be two real-valued functions.
We have
\begin{align*}
&\bigg| \int \hat{h}_n d\widehat{P}_n - \int h d\bar{P} \bigg|\\[-0.2cm]
&\hspace{1.3cm} \le \big\|\hat{h}_n - h\big\|_\infty ~+~ \|h\|_\infty \sum_{k=1}^K \left| \hat{\lambda}_k - \lambda_k \right| ~+~ \sum_{k=1}^K \hat{\lambda}_k \left| \int hd\widehat{P}_k - \int hdP_k \right|.
\end{align*}
\end{lemma}
\smallskip

\noindent\textit{Proof.}
It holds
\begin{align*}
\bigg| \int \hat{h}_n&(z) d\widehat{P}_n(z) - \int h(z) d\bar{P}(z) \bigg|\\[0.0cm]
&\le \Bigg| \int \hat{h}_n(z) d\widehat{P}_n(z) - \int  h(z) d\widehat{P}_n(z) \Bigg| + \left| \int h(z) d\widehat{P}_n(z) - \int h(z) d\bar{P}(z) \right|\\[0.0cm]
&\le \sup_z \left|\hat{h}_n(z) - h(z)\right| + \left| \sum_{k=1}^K \hat{\lambda}_k \int h(z)d\widehat{P}_k(z) - \sum_{k=1}^K \lambda_k \int h(z)dP_k(z) \right|\\[0.0cm]
&\le \sup_z \left|\hat{h}_n(z) - h(z)\right| + \left| \sum_{k=1}^K \hat{\lambda}_k \int h(z)d\widehat{P}_k(z) - \sum_{k=1}^K \hat{\lambda}_k \int h(z)dP_k(z) \right|\\
& \hspace{3.2cm} + \left| \sum_{k=1}^K \hat{\lambda}_k \int h(z)dP_k(z) - \sum_{k=1}^K \lambda_k \int h(z)dP_k(z) \right|\\[0.0cm]
&\le \sup_z \left|\hat{h}_n(z) - h(z)\right| + \sup_z |h(z)| \sum_{k=1}^K \left| \hat{\lambda}_k - \lambda_k \right|\\
& \hspace{3.2cm} + \sum_{k=1}^K \hat{\lambda}_k \left| \int h(z)d\widehat{P}_k(z) - \int h(z)dP_k(z) \right|\\[-0.2cm]
&\hspace{10.5cm}\qed
\end{align*}

\begin{corollary}\label{cor:Hoeffding}
Let $\hat{h}_n : \mathcal{Z} \rightarrow \mathbb{R}$ and $h : \mathcal{Z} \rightarrow \mathbb{R}$ be two real-valued functions.
Assume that there exist $a, b \in \mathbb{R}^2$ such that: $a \le h(z) \le b$ for all $z \in \mathcal{Z}$.
If \Cref{hyp:rates} is satisfied, then for all $t > 0$ and $n \ge (b-a)^2\log(2K)/(2\underline{\lambda}t^2)$, it holds with probability at least $1 - 2K\exp\left(-\frac{2\underline{\lambda}nt^2}{(b-a)^2}\right)$:
\begin{equation*}
\bigg| \int \hat{h}_n(z) d\widehat{P}_n(z) - \int h(z) d\bar{P}(z) \bigg| \le \sup_z \left|\hat{h}_n(z) - h(z)\right| + \frac{C_\lambda K \sup_z |h(z)|}{\sqrt{n}} + t.
\end{equation*}
\end{corollary}
\smallskip

\begin{proof}
Using \Cref{lem:decompo} and \Cref{hyp:rates}, we have
\begin{align*}
&\bigg| \int \hat{h}_n(z) d\widehat{P}_n(z) - \int h(z) d\bar{P}(z) \bigg|\\
&\hspace{1.2cm}\le \sup_z \left|\hat{h}_n(z) - h(z)\right| + \frac{C_\lambda K \sup_z |h(z)|}{\sqrt{n}} + \sum_{k=1}^K \hat{\lambda}_k\left| \int hd\widehat{P}_k - \int hdP_k \right|.
\end{align*}
Now, applying Hoeffding's inequality gives that, for all $t > 0$ and all $k \le K$,
\begin{align*}
\mathbb{P}\bigg\{\bigg| \int hd\widehat{P}_k - \int hdP_k \bigg| > t \bigg\} &\le 2 \exp\left(-\frac{2n_kt^2}{(b - a)^2}\right) \le 2 \exp\left(-\frac{2 \underline{\lambda }nt^2}{(b - a)^2}\right).
\end{align*}
The proof is concluded by applying the union bound.
\end{proof}
\smallskip

\begin{proposition}\label{prop:deviation}
Suppose that \Cref{hyp:rates} is verified.
Then, for all $t > 0$ and $n \ge \log(2K^2)/(2\underline{\lambda}t^2)$, it holds with probability at least $1 - 2K^2\exp\left(-2\underline{\lambda}nt^2\right)$:
\begin{equation*}
\left\| \widehat{D}'_n(\bu^*) - \bar{D}'(\bu^*) \right\|_2 \le \frac{2C_\lambda K^{3/2}}{\sqrt{n}} + \sqrt{K}t.
\end{equation*}
\end{proposition}

\begin{proof}
Apply \Cref{cor:Hoeffding} for every component $k$ of $\widehat{D}'_n(\bu^*) - \bar{D}'(\bu^*)$ with $\hat{h}_n = e^{u^*_k}\omega_k / (\sum_l e^{u^*_l}\omega_l) - \hat{\lambda}_k$ and $h = e^{u^*_k}\omega_k / (\sum_l e^{u^*_l}\omega_l) - \lambda_k$,
and the union bound permits to conclude.
\end{proof}
\smallskip

\noindent \textit{Proof of \Cref{prop:dev_W}.}
Combining \Cref{lem:lipschitz}, Propositions \ref{prop:strong_cvx} and \ref{prop:deviation}, we have that it holds with probability at least $1 - M_4\exp(-c_4 n) - 2K^2\exp(-2\underline{\lambda}nt^2)$:
\begin{align*}
\left\|\widehat{\bm{W}}_n - \bm{W}^* \right\|_2 &\le \|\hat{\bu}_n - \bu^*\|_2 + C_\lambda \sqrt{\frac{K}{n}}\\
&\le L \left\|\widehat{D}'_n(\bu^*) - \bar{D}'(\bu^*)\right\|_2 + C_\lambda \sqrt{\frac{K}{n}}\\
&\le L\left(\frac{2C_\lambda K^{3/2}}{\sqrt{n}} + \sqrt{K}t\right) + C_\lambda \sqrt{\frac{K}{n}}\\
&= L\sqrt{K}t + \frac{C_\lambda\sqrt{K}(2LK + 1)}{\sqrt{n}}.
\end{align*}
The proof is concluded by setting $\gamma_2 = C_\lambda\sqrt{K}(2LK + 1)$, $M_2 = M_4$, $c_2 = c_4$, $M'_2 = 2K^2$, $c'_2 = 2\underline{\lambda}/(L^2 K)$, and $n_{0, 2} = n_{0, 4}$.
\hfill\qed

%%%%%%%%%%%%%%%%%%%%%%%%%%%
%                         %
%     PROOF OF PROP 3     %
%                         %
%%%%%%%%%%%%%%%%%%%%%%%%%%%

\subsection{Proof of \Cref{prop:dev_omega}}
\label{apx:proof_dev_omega}

Recall that by \Cref{prop:unique_bounded}, it holds with probability $1 - M_1\exp(-c_1n)$:
\begin{equation*}
\forall k \le K, \qquad \rho \le \widehat{W}_{n, k} \le 1, \quad \text{and} \quad \rho \le W^*_k \le 1.
\end{equation*}
This implies for all $z \in \mathcal{Z}$:
\begin{equation*}
\rho \le \left(\sum_{l=1}^K\frac{\hat{\lambda}_l \omega_l(z)}{\widehat{W}_{n, l}}\right)^{-1} \le \frac{1}{\varepsilon\underline{\lambda}}, \quad\text{and}\quad \rho \le \left(\sum_{l=1}^K\frac{\lambda_l \omega_l(z)}{W^*_l}\right)^{-1} \le \frac{1}{\varepsilon\underline{\lambda}}.
\end{equation*}
Using the above inequalities and the mean value theorem on $t \mapsto 1/t$, we get for all $k \le K$:
\begin{align}
&\Big| \widehat{\Omega}_{n, k} - \Omega_k \Big| = \left| \frac{\widehat{W}_{n, k}}{\int \left(\sum_{l=1}^K \frac{\hat{\lambda}_l \omega_l}{\widehat{W}_{n, l}}\right)^{-1} d\widehat{P}_n}  - \frac{W^*_k}{\int \left(\sum_{l=1}^K \frac{\lambda_l \omega_l}{W^*_l}\right)^{-1}d\bar{P}} \right|\nonumber\\[0.5cm]
&\le \frac{1}{\int \left(\sum_{l=1}^K \frac{\hat{\lambda}_l \omega_l}{\widehat{W}_{n, l}}\right)^{-1} d\widehat{P}_n} \left| \widehat{W}_{n, k} - W^*_k \right|\nonumber\\[0.2cm]
&~~+W^*_k \left| \frac{1}{\int \left(\sum_{l=1}^K \frac{\hat{\lambda}_l \omega_l}{\widehat{W}_{n, l}}\right)^{-1} d\widehat{P}_n} - \frac{1}{\int \left(\sum_{l=1}^K \frac{\lambda_l \omega_l}{W^*_l}\right)^{-1}d\bar{P}} \right|\nonumber\\[0.5cm]
&\le \frac{1}{\rho}\left| \widehat{W}_{n, k} - W^*_k \right| + \frac{1}{\rho^2} \left| \int \left(\sum_{l=1}^K \frac{\hat{\lambda}_l \omega_l}{\widehat{W}_{n, l}}\right)^{-1} d\widehat{P}_n - \int \left(\sum_{l=1}^K \frac{\lambda_l \omega_l}{W^*_l}\right)^{-1}d\bar{P} \right|.\label{eq:omegas}
\end{align}

\noindent The first term in \Cref{eq:omegas} can be bounded using \Cref{prop:dev_W}.
For the second, we  can use \Cref{cor:Hoeffding}.
First we must compute:
\begin{align*}
\left| \left(\sum_{l=1}^K \frac{\hat{\lambda}_l \omega_l(z)}{\widehat{W}_{n, l}}\right)^{-1}\right. - &\left.\left(\sum_{l=1}^K \frac{\lambda_l \omega_l(z)}{W^*_l}\right)^{-1} \right| \le \left(\frac{1}{\varepsilon\underline{\lambda}}\right)^2 \left| \sum_{l=1}^K \frac{\hat{\lambda}_l \omega_l(z)}{\widehat{W}_{n, l}} - \sum_{l=1}^K \frac{\lambda_l \omega_l(z)}{W^*_l} \right|\\
&\le \left(\frac{1}{\varepsilon\underline{\lambda}}\right)^2 \sum_{l=1}^K \frac{\left| \hat{\lambda}_l - \lambda_l\right|}{\widehat{W}_{n, l}} + \lambda_l \left| \frac{1}{\widehat{W}_{n, l}} - \frac{1}{W^*_l} \right|\\
&\le \left(\frac{1}{\varepsilon\underline{\lambda}}\right)^2 \left(\frac{C_\lambda K}{\rho \sqrt{n}} + \frac{1}{\rho^2} \sum_{l=1}^K \lambda_l \left| \widehat{W}_{n, l} - W^*_l \right|\right).
\end{align*}
\Cref{prop:dev_W} then allows to bound the last term with overwhelming probability.
Next, applying \Cref{cor:Hoeffding} with $\hat{h}_n = \left(\sum_l \frac{\hat{\lambda}_l \omega_l}{\widehat{W}_{n, l}}\right)^{-1}$ and $h = \left(\sum_l \frac{\lambda_l \omega_l}{W^*_l}\right)^{-1}$, we obtain that for all $t_1, t_2 > 0$ with probability at least $1 - M_2\exp(-c_2n) - M'_2\exp(-c'_2nt_1^2) - 2K\exp(-2\varepsilon^2\underline{\lambda}^3nt_2^2)$ it holds for all $k\le K$:
\begin{align*}
\Big| \widehat{\Omega}_{n, k}& - \Omega_k \Big|\\
&\le \frac{1}{\rho}\left(t_1 + \frac{\gamma_2}{\sqrt{n}}\right) + \frac{1}{\rho^2}\left(\frac{C_\lambda K}{\varepsilon^2\underline{\lambda}^2\rho\sqrt{n}} + \frac{1}{\varepsilon^2\underline{\lambda}^2\rho^2}\left(t_1 + \frac{\gamma_2}{\sqrt{n}}\right) + \frac{C_\lambda K}{\varepsilon\underline{\lambda}\sqrt{n}} + t_2\right)\\[0.5cm]
&= t_1\left(\frac{1}{\rho} + \frac{1}{\varepsilon^2\underline{\lambda}^2 \rho^4}\right) + \frac{t_2}{\rho^2}  + \left(\frac{\gamma_2}{\rho} + \frac{C_\lambda K}{\varepsilon^2\underline{\lambda}^2\rho^3} + \frac{\gamma_2}{\varepsilon^2\underline{\lambda}^2\rho^4} + \frac{C_\lambda K}{\varepsilon\underline{\lambda}\rho^2}\right)\frac{1}{\sqrt{n}}.
\end{align*}
The proof is concluded by setting $M_3 = M_2$, $c_3 = c_2$, $M'_3 = 2\max\big(M'_2; 2K\big)$,
\begin{align*}
c'_3 &= \max\left(\frac{c'_2}{4\left(\frac{1}{\rho} + \frac{1}{\varepsilon^2\underline{\lambda}^2\rho^4}\right)^2}; \frac{\varepsilon^2 \underline{\lambda}^3 \rho^4}{2}\right),\\[0.2cm]
\gamma_3 &= \frac{\gamma_2}{\rho} + \frac{C_\lambda K}{\varepsilon^2\underline{\lambda}^2\rho^3} + \frac{\gamma_2}{\varepsilon^2\underline{\lambda}^2\rho^4} + \frac{C_\lambda K}{\varepsilon\underline{\lambda}\rho^2},
\end{align*}
and $n_{0, 3} = n_{0, 2}$.
\hfill\qed

%%%%%%%%%%%%%%%%%%%%%%%%%%
%                        %
%     PROOF OF THM 4     %
%                        %
%%%%%%%%%%%%%%%%%%%%%%%%%%

\subsection{Proof of \Cref{thm:main}}
\label{apx:proof_thm}

Let $\theta \in \Theta$.
The first step of the proof consists in using \Cref{lem:decompo} with the choices $\hat{h}_{n, \theta}(z) = \psi(z, \theta)\left(\sum_{k=1}^K \frac{\hat{\lambda}_k \omega_k(z)}{\widehat{\Omega}_{n, k}}\right)^{-1}$, and $h_\theta(z) = \psi(z, \theta)\left(\sum_{k=1}^K \frac{\lambda_k \omega_k(z)}{\Omega_k}\right)^{-1}$.
We obtain
\begin{align}
\Big|&\widetilde{L}_n(\theta) - L(\theta)\Big|\nonumber\\
& = \left|\int \psi(z,\theta) \left( \sum_{k=1}^K \frac{\hat{\lambda}_k \omega_k(z)}{\widehat{\Omega}_{n, k}} \right)^{-1} d\widehat{P}_n(z) - \int \psi(z,\theta)\left(\sum_{k=1}^K \frac{\lambda_k \omega_k(z)}{\Omega_k} \right)^{-1} d\bar{P}(z)\right|\nonumber\\
& = \left|\int \hat{h}_n(z, \theta) d\widehat{P}_n(z) - \int h(z, \theta) d\bar{P}(z)\right|\nonumber\\
&\le \sup_z \left|\hat{h}_{n, \theta}(z) - h_\theta(z)\right| \,+\, \sup_z |h_\theta(z)| \sum_{k=1}^K |\hat{\lambda}_k - \lambda_k| \,+\, \sum_{k=1}^K \hat{\lambda}_k \left|\int h_\theta \,d(\widehat{P}_k - P)\right|\nonumber\\
&\le \sup_z \left|\hat{h}_{n, \theta}(z) - h_\theta(z)\right| ~+~ \frac{C_\lambda K \sup_z |h_\theta(z)|}{\sqrt{n}} ~+~ \max_{k \le K} ~ \left|\int h_\theta \,d(\widehat{P}_k - P)\right|.\label{eq:decompo}
\end{align}
Now, by the definitions of $\widehat{\bm{\Omega}}_n$ and $\bm{\Omega}$, for all $k \le K$, we have
\begin{equation*}
\varepsilon\underline{\lambda} \rho \le \widehat{\Omega}_{n, k} \le \frac{1}{\rho}, \qquad \text{and} \qquad \varepsilon\underline{\lambda} \rho \le \Omega_k \le 1.
\end{equation*}
Hence, for all $z \in \mathcal{Z}$ it holds
\begin{equation*}
0 \le \left(\sum_{k=1}^K \frac{\hat{\lambda}_k \omega_k(z)}{\widehat{\Omega}_{n, k}}\right)^{-1} \le \frac{1}{\varepsilon\underline{\lambda} \rho}, \quad \text{and} \quad 0 \le \left(\sum_{k=1}^K \frac{\lambda_k \omega_k(z)}{\Omega_k}\right)^{-1}\le \frac{1}{\varepsilon\underline{\lambda}}.
\end{equation*}
As $|\psi(z, \theta)| \le 1$, this implies that $\sup_{z, \theta} |h_\theta(z)| \le 1/(\varepsilon\underline{\lambda})$.
And we also have
\begin{align*}
\left|\hat{h}_{n, \theta}(z) - h_\theta(z)\right| &= \left| \psi(z, \theta)\left(\sum_{k=1}^K \frac{\hat{\lambda}_k \omega_k(z)}{\widehat{\Omega}_{n, k}}\right)^{-1} - \psi(z, \theta)\left(\sum_{k=1}^K \frac{\lambda_k \omega_k(z)}{\Omega_k}\right)^{-1}\right|\\
&\le \left(\frac{1}{\varepsilon\underline{\lambda}\rho}\right)^2 \left|\sum_{k=1}^K \frac{\hat{\lambda}_k \omega_k(z)}{\widehat{\Omega}_{n, k}} - \sum_{k=1}^K \frac{\lambda_k \omega_k(z)}{\Omega_k}\right|,\\
&\le \left(\frac{1}{\varepsilon\underline{\lambda}\rho}\right)^3 \frac{C_\lambda K}{\sqrt{n}} + \left(\frac{1}{\varepsilon\underline{\lambda} \rho}\right)^4 \sum_{k=1}^K \lambda_k \left| \widehat{\Omega}_{n, k} - \Omega_k \right|.
\end{align*}
Plugging into \Cref{eq:decompo}, and taking the supremum over $\theta \in \Theta$, we obtain
\begin{align}\label{eq:decompo_sup}
\sup_{\theta \in \Theta} \Big|\widetilde{L}_n(\theta) - L(\theta)\Big| \le &\left(\frac{1}{\varepsilon\underline{\lambda}\rho}\right)^3 \frac{C_\lambda K}{\sqrt{n}} + \left(\frac{1}{\varepsilon\underline{\lambda} \rho}\right)^4 \sum_{k=1}^K \lambda_k \left| \widehat{\Omega}_{n, k} - \Omega_k \right|\nonumber\\
&+ \frac{C_\lambda K}{\varepsilon\underline{\lambda}\sqrt{n}}  + \max_k \sup_{\theta \in \Theta}\left|\int h_\theta(z) d(\widehat{P}_k - P)(z)\right|
\end{align}
Thus, we have bounded $\sup_{\theta \in \Theta} |\widetilde{L}_n(\theta) - L(\theta)|$ by a sum involving: $(1)$ non-random terms scaling as $\mathcal{O}(1/\sqrt{n})$, $(2)$ random terms independent from $\theta$ which can be controlled using \Cref{prop:dev_omega}, and $(3)$ the supremum of an empirical process.
We can now use standard arguments such as chaining \cite{vdG00,VanderVaart1996} to bound this last term.
Let $k \le K$, and $t > 0$, we have
\begin{align}
\mathbb{P}\left\{\sup_{\theta \in \Theta}\left|\int h_\theta(z) d(\widehat{P}_k - P)(z)\right| > t\right\}\label{eq:emp_process}\\
&\hspace{-2cm}= \mathbb{P}\left\{\sup_{\theta \in \Theta}\left|\int \varepsilon\underline{\lambda}h(z, \theta) d(\hat{P}_k - P)(z)\right| > \varepsilon\underline{\lambda}t\right\}.\nonumber
\end{align}
By applying Theorem 2.14.9 in \cite{VanderVaart1996} to the class
\[
\mathcal{G}_\Theta = \Big\{\varepsilon\underline{\lambda}h(\cdot, \theta) \colon \theta \in \Theta \Big\} = \left\{\varepsilon\underline{\lambda}\left(\sum_{k=1}^K \frac{\lambda_k \omega_k(\cdot)}{\Omega_k}\right)^{-1} \psi(\cdot, \theta) \colon \theta \in \Theta\right\}
\]
which also satisfies \Cref{hyp:complex}, since it is a pointwise multiplication of $\mathcal{F}_\Theta = \big\{\psi(\cdot, \theta) \colon \theta \in \Theta \big\}$ by a function with values in $[0, 1]$, we obtain that \eqref{eq:emp_process} is upper bounded by
\begin{equation}\label{eq:foo}
\left(\frac{\Delta \varepsilon\underline{\lambda}\sqrt{n_k}t}{\sqrt{r}}\right)^r e^{-2(\varepsilon\underline{\lambda})^2 n_k t^2} \le \left(\frac{\Delta \varepsilon\underline{\lambda}}{\sqrt{r}}\right)^r n^\frac{r}{2}t^r e^{-2\varepsilon^2\underline{\lambda}^3 n t^2}
\end{equation}
where $\Delta$ is a constant that depends only on $C_\Theta$.
Finally, plugging \eqref{eq:foo} with the union bound and \Cref{prop:dev_omega} into \eqref{eq:decompo_sup}, we get that with probability at least $1 - M_3e^{-c_3n} - M'_3e^{-c'_3nt_1^2} - K\left(\frac{\Delta \varepsilon\underline{\lambda}}{\sqrt{r}}\right)^r n^\frac{r}{2}t_2^r~e^{-2\varepsilon^2\underline{\lambda}^3 n t_2^2}$
\begin{align*}
\sup_{\theta \in \Theta} \Big|\widetilde{L}_n(\theta) - L(\theta)\Big| \le \left(\frac{1}{\varepsilon\underline{\lambda}\rho}\right)^3 \frac{C_\lambda K}{\sqrt{n}} + \left(\frac{1}{\varepsilon\underline{\lambda} \rho}\right)^4 \left(\frac{\gamma_3}{\sqrt{n}} + t_1\right) + \frac{C_\lambda K}{\varepsilon\underline{\lambda}\sqrt{n}}  + t_2\,,
\end{align*}
or again, we have with probability at least $1 - M_3e^{-c_3n} - M'_3\exp\left(-\frac{c'_3(\varepsilon\underline{\lambda}\rho)^8 nt^2}{4}\right) - K\left(\frac{\Delta \varepsilon\underline{\lambda}}{2\sqrt{r}}\right)^r n^\frac{r}{2}t^r \exp\left(-\frac{\varepsilon^2\underline{\lambda}^3 n t^2}{2}\right)$
\begin{align*}
\sup_{\theta \in \Theta} \Big|\widetilde{L}_n(\theta) - L(\theta)\Big| \le \frac{\gamma}{\sqrt{n}} + t\,,
\end{align*}
with
\[
\gamma = \frac{C_\lambda K}{(\varepsilon\underline{\lambda}\rho)^3} + \frac{\gamma_3}{(\varepsilon\underline{\lambda} \rho)^4} + \frac{C_\lambda K}{\varepsilon\underline{\lambda}}\,.
\]
The proof is concluded by setting $M = M_3$, $c = c_3$, $M' = M'_3$, $M'' = K\left(\frac{\Delta \varepsilon\underline{\lambda}}{2\sqrt{r}}\right)^r$, $c' = c'_3(\varepsilon\underline{\lambda}\rho)^8/4$, $c'' = (\varepsilon^2\underline{\lambda}^3)/2$, and $n_0 = n_{0, 3}$.\hfill\qed

%%%%%%%%%%%%%%%%%%%%%%%
%                     %
%     NEW THM DKW     %
%                     %
%%%%%%%%%%%%%%%%%%%%%%%

\subsection{Proof of \Cref{thm:dkw}}
\label{apx:proof_thm_dkw}

Let $\mathcal{F} = \big\{z \mapsto \mathbb{I}\{z \le \tau\} \colon \tau \in \mathbb{R}\big\}$.
As discussed in the main body of the paper, applying \Cref{thm:main} to $\mathcal{F}$ in a straightforward fashion yields a bound that does not match the standard DKW inequality.
To match the rate of the DKW inequality, we have to develop a refined analysis, specifically tailored to classes which are composed of indicator functions.
We introduce the following complexity assumption \cite[Chapter 14]{VanderVaart1996}, that strengthens \Cref{hyp:complex}.

\begin{assumption}\label{hyp:complex_indicator}
The class $\mathcal{F} = \mathcal{F}_\mathcal{C}$ is composed of indicator functions of sets, i.e., $\mathcal{F}_\mathcal{C} = \{z \mapsto \mathbb{I}\{z \in C\} \colon C \in \mathcal{C}\}$, with $\mathcal{C}$ a collection of sets of satisfying for some constants $C_\mathcal{C} > 0$ and $r\ge 1$
\[
\sup_{Q}\mathcal{N}(\zeta,\; \mathcal{C},\; L_1(Q) )\leq (C_\mathcal{C}/\zeta)^r.
\]
Furthermore, for $k \le K$ and $\delta > 0$, let $\mathcal{C}_{k, \delta} = \{C \in \mathcal{C} \colon |P_k(C) - 1/2| < \delta \}$.
We also assume the existence of $C'_\mathcal{C}, r', r''$ such that for every $\delta \ge \zeta > 0$
\[
\sup_{k \le K} \mathcal{N}(\zeta,\; \mathcal{C}_{k, \delta},\; L_1(P_k) )\leq C'_\mathcal{C}~\delta^{r'}~\zeta^{-r''}.
\]
\end{assumption}

\noindent Note that a class $\mathcal{F}_\mathcal{C}$ of finite {\sc VC} dimension $V<+\infty$ verifies the first inequality with $r = V - 1$, and $C_\mathcal{C}$ that depends only on $V$, see e.g., Theorem 2.6.4 in \cite{VanderVaart1996}.
Under \Cref{hyp:complex_indicator}, a tighter control of the empirical processes in decomposition \eqref{eq:decompo_text} is possible, yielding the following theorem.

\begin{theorem}\label{thm:new_thm}
Suppose that Assumptions \ref{hyp:rates}, \ref{hyp:connect}, \ref{hyp:bounded_omega}, \ref{asm:eigenvalue}, and \ref{hyp:complex_indicator} are satisfied.
Then, there exist $M, M', M'', c, c', c'', \gamma, n_0$, depending only on $K, C_\lambda, \underline{\lambda}, \kappa, \varepsilon, \sigma, C_\mathcal{C}, C'_\mathcal{C}$, $r$, $r'$, and $r''$ such that for all $t > 0$ and $n \ge n_0$ it holds:
\begin{equation*}
\mathbb{P}\left\{\sup_{\theta \in \Theta} \left|\widetilde{L}_n(\theta) - L(\theta)\right| > \frac{\gamma}{\sqrt{n}} + t\right\} \le Me^{-cn} + M'e^{-c'nt^2} + (nt^2)^{r''-r'} M''e^{-c''nt^2}.
\end{equation*}
\end{theorem}

\begin{proof}
The proof follows the same path as that of \Cref{thm:main}.
In particular, we start from the same decomposition \eqref{eq:decompo_sup}, but \Cref{hyp:complex_indicator} now allows a better control on the empirical processes that compose the last term.
Specifically, for every $k \le K$, Theorem 2.14.14 in \cite{VanderVaart1996} gives that
\begin{align*}
\mathbb{P}\left\{\sup_{C \in \mathcal{C}}\left|\int h_C(z) d(\widehat{P}_k - P)(z)\right| > t\right\} &\le \Delta (\varepsilon\underline{\lambda}\sqrt{n_k}t)^{2r'' - 2r'}e^{-2\varepsilon^2\underline{\lambda}^2n_k t^2}\\
&\le \Delta (\varepsilon\underline{\lambda}\sqrt{n}t)^{2r'' - 2r'}e^{-2\varepsilon^2\underline{\lambda}^3n t^2}
\end{align*}
where $\Delta$ is a constant that depends only on $C_\mathcal{C}, C'_\mathcal{C}, r, r'$, and $r''$.
Plugging into \Cref{eq:decompo_sup}, we get that $\sup_{C \in \mathcal{C}} \Big|\widetilde{L}_n(C) - L(C)\Big| \le \frac{\gamma}{\sqrt{n}} + t$ with probability at least $1 - M_3e^{-c_3n} - M'_3\exp\left(-\frac{c'_3(\varepsilon\underline{\lambda}\rho)^8 nt^2}{4}\right) - K\Delta (\varepsilon\underline{\lambda}\sqrt{n}t)^{2r'' - 2r'}~e^{-\frac{\varepsilon^2\underline{\lambda}^3 n t^2}{2}}$.
We conclude by setting $M=M_3$, $c=c_3$, $M'=M'_3$, $c' = c'_3(\varepsilon\underline{\lambda}\rho)^8/4$, $M'' = K\Delta (\varepsilon\underline{\lambda})^{2r'' - 2r'}$, $c'' = (\varepsilon^2\underline{\lambda}^3)/2$, and $n_0 = n_{0, 3}$.
\end{proof}

\Cref{thm:dkw} is actually a corollary of \Cref{thm:new_thm}, applied to the class $\mathcal{F} = \big\{z \mapsto \mathbb{I}\{z \le \tau\} \colon \tau \in \mathbb{R}\big\}$.\\

\noindent \textit{Proof of \Cref{thm:dkw}.}
The class $\mathcal{F} = \big\{z \mapsto \mathbb{I}\{z \le \tau\} \colon \tau \in \mathbb{R}\big\}$ satisfies \Cref{hyp:complex_indicator} with $r = r' = r'' = 1$, see \cite[page 247]{VanderVaart1996}.
Applying \Cref{thm:new_thm}, we obtain
\begin{align*}
\mathbb{P}\left\{ \sup_{z \in \mathbb{R}}\left|(\widetilde{P}_n - P)((-\infty,\; z])\right| > \frac{\gamma}{\sqrt{n}} + t \right\}\\
&\hspace{-1cm}\le Me^{-cn} + M'e^{-c'nt^2} + M''e^{-c''nt^2}\\
&\hspace{-1cm}\le Me^{-cn} + 2\max(M'; M'')e^{-\min(c'; c'')nt^2}\,.
\end{align*}
\hfill\qed

%%%%%%%%%%%%%%%%%%%%%%%%
%                      %
%   ADDITIONAL EXPES   %
%                      %
%%%%%%%%%%%%%%%%%%%%%%%%

\section{Additional Experiments}
\label{apx:expe}

In this section we provide additional experimental results, both on a synthetic estimation problem (\Cref{apx:gauss_expes}) and real data learning applications, see \Cref{apx:adult_expes}.

\subsection{Estimation Experiments}
\label{apx:gauss_expes}

Recall that the synthetic data here consist of 1000 train and 300 test realizations of a 3-dimensional Gaussian random vector.
The goal is to predict the norm of the realizations through different learning algorithms: Linear Regression (LR), Kernel Ridge Regression (KRR), Support Vector Regression (SVR) and Random Forest (RF).
They are implemented with default hyperparameters, as focus is not on the performances \textit{per se}, but rather on the impact of the debiasing for a given model.
The biasing functions $\omega_k$ used here are indicator functions of subspaces of $\mathbb{R}^3$.
These functions (or equivalently the subsets) are chosen according to twelve different scenarios, so as to contrast the debiasing effects.
When one biasing function is the identity (i.e., one subspace is $\mathbb{R}^3$), the algorithm is also trained on the sole unbiased sample.
However, this approach does not benefit from the whole dataset, and performances reported compare unfavorably to debiased ERM.
Numerical results are gathered in \Cref{tab:complete_norm,tab:complete_dim}.
For scenarios in which no subspace is $\mathbb{R}^3$, two lines are displayed: the upper one corresponds to the standard ERM (ERM), while the second one is achieved through the debiased approach we promote (db-ERM).
When one subspace is $\mathbb{R}^3$, a third line is displayed, which corresponds to the result obtained with training on the sole unbiased sample (ub-ERM).

\begin{figure*}[!t]
\centering
\subfigure[Scenario a)\label{fig:sc_a}]{\includegraphics[width=0.45\textwidth]{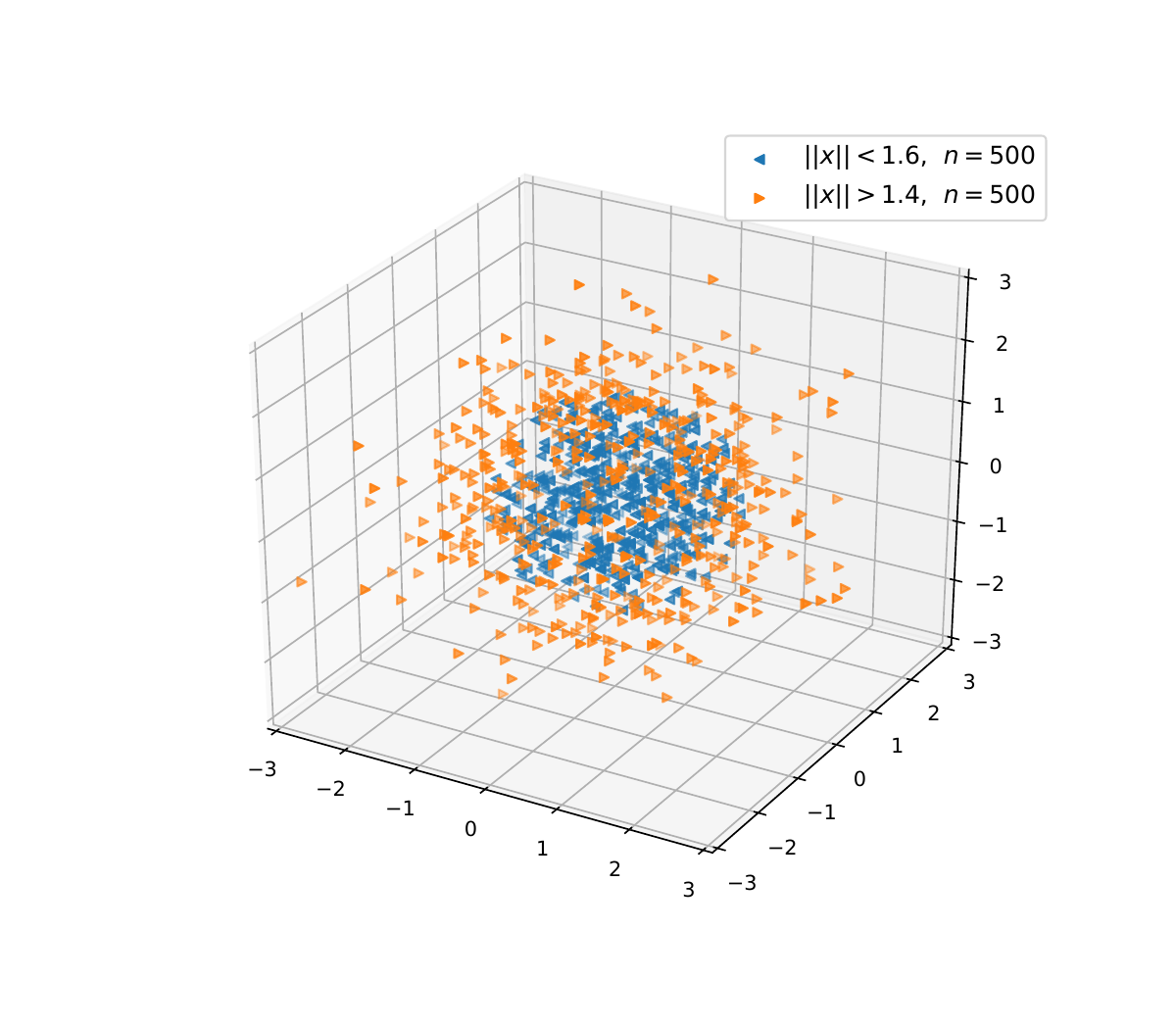}}
\subfigure[Scenario b)\label{fig:sc_b}]{\includegraphics[width=0.45\textwidth]{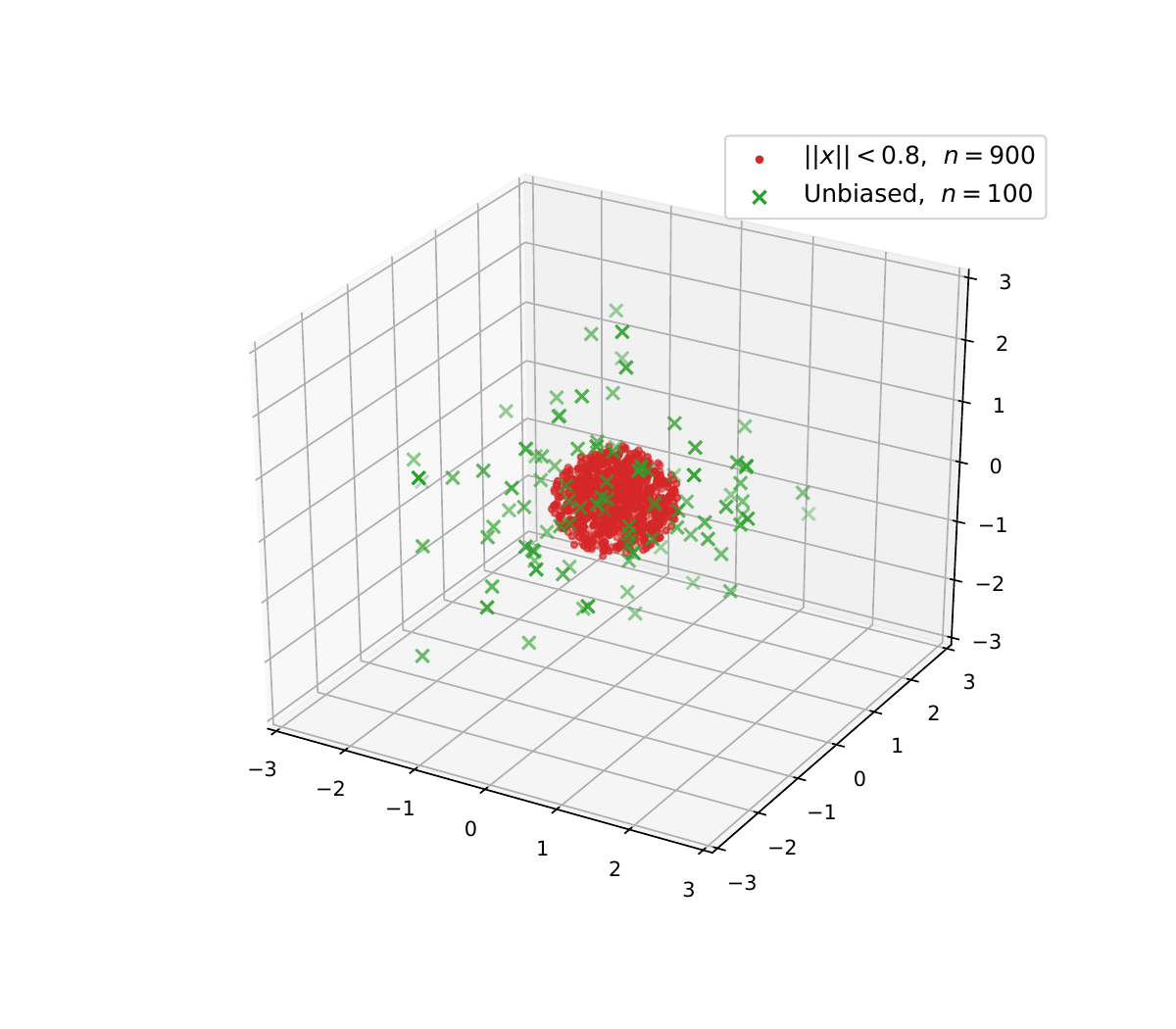}}\\
\subfigure[Scenario c)\label{fig:sc_c}]{\includegraphics[width=0.45\textwidth]{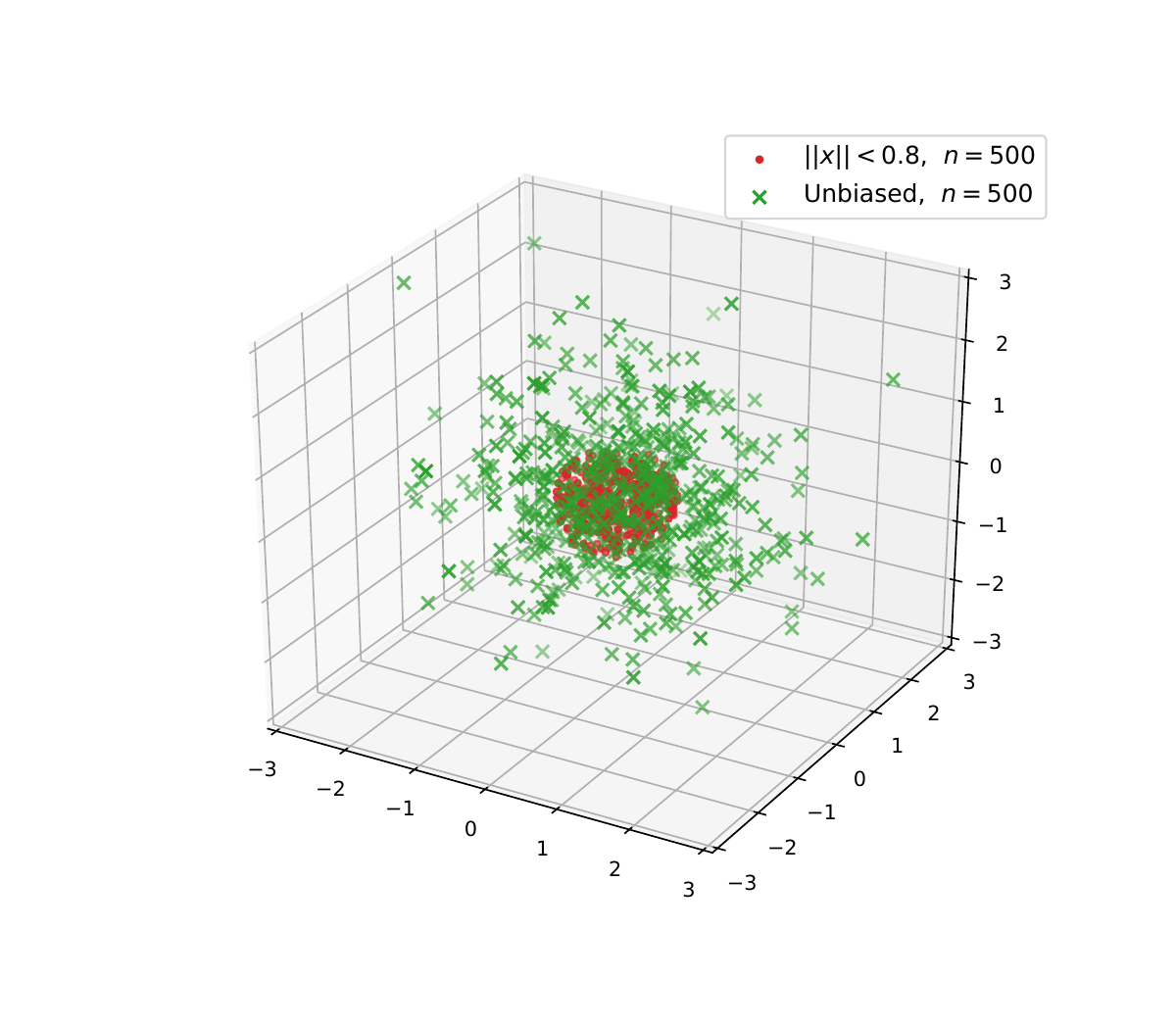}}
\subfigure[Scenario d)\label{fig:sc_d}]{\includegraphics[width=0.45\textwidth]{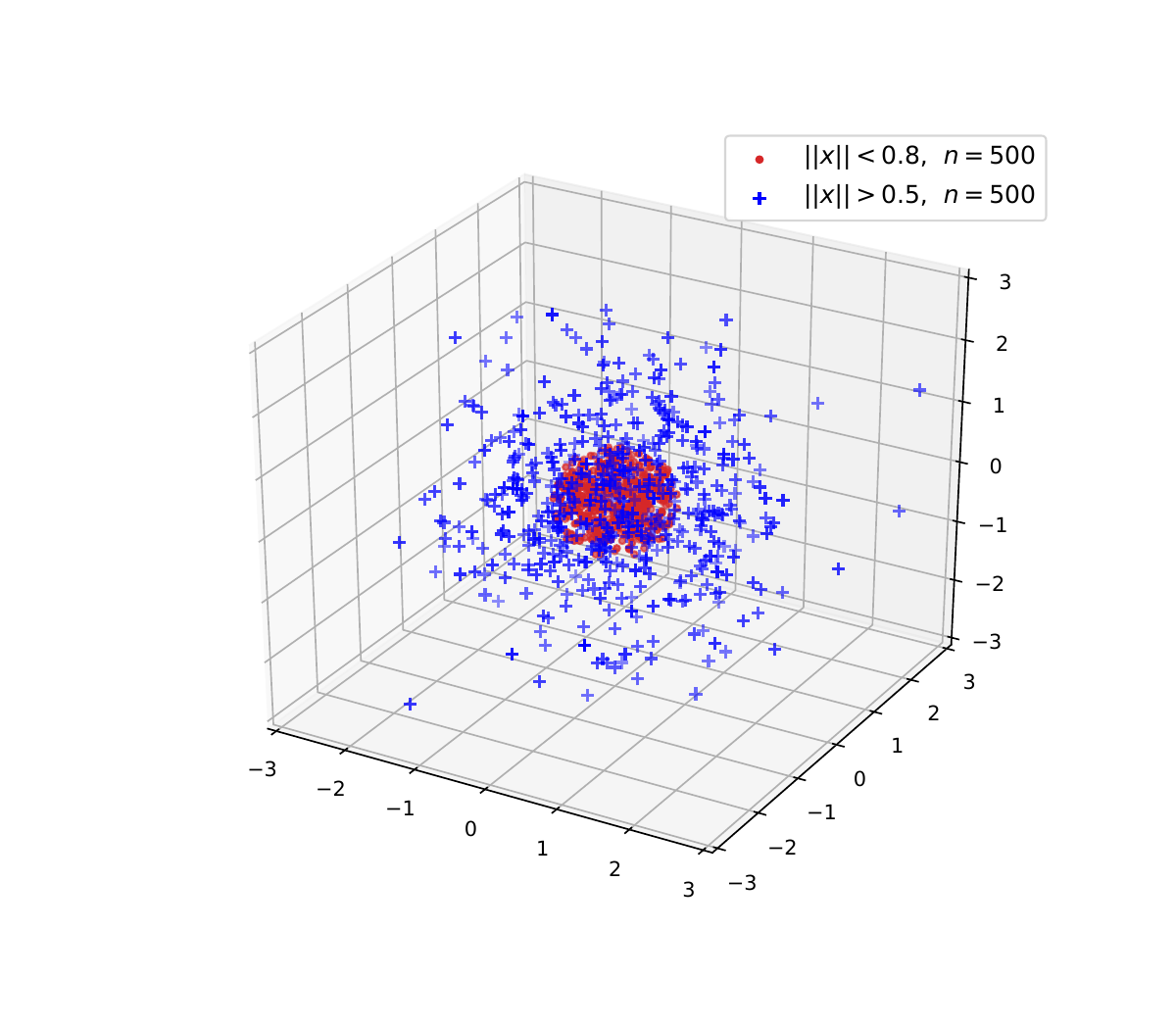}}\\
\subfigure[Scenario e)\label{fig:sc_e}]{\includegraphics[width=0.45\textwidth]{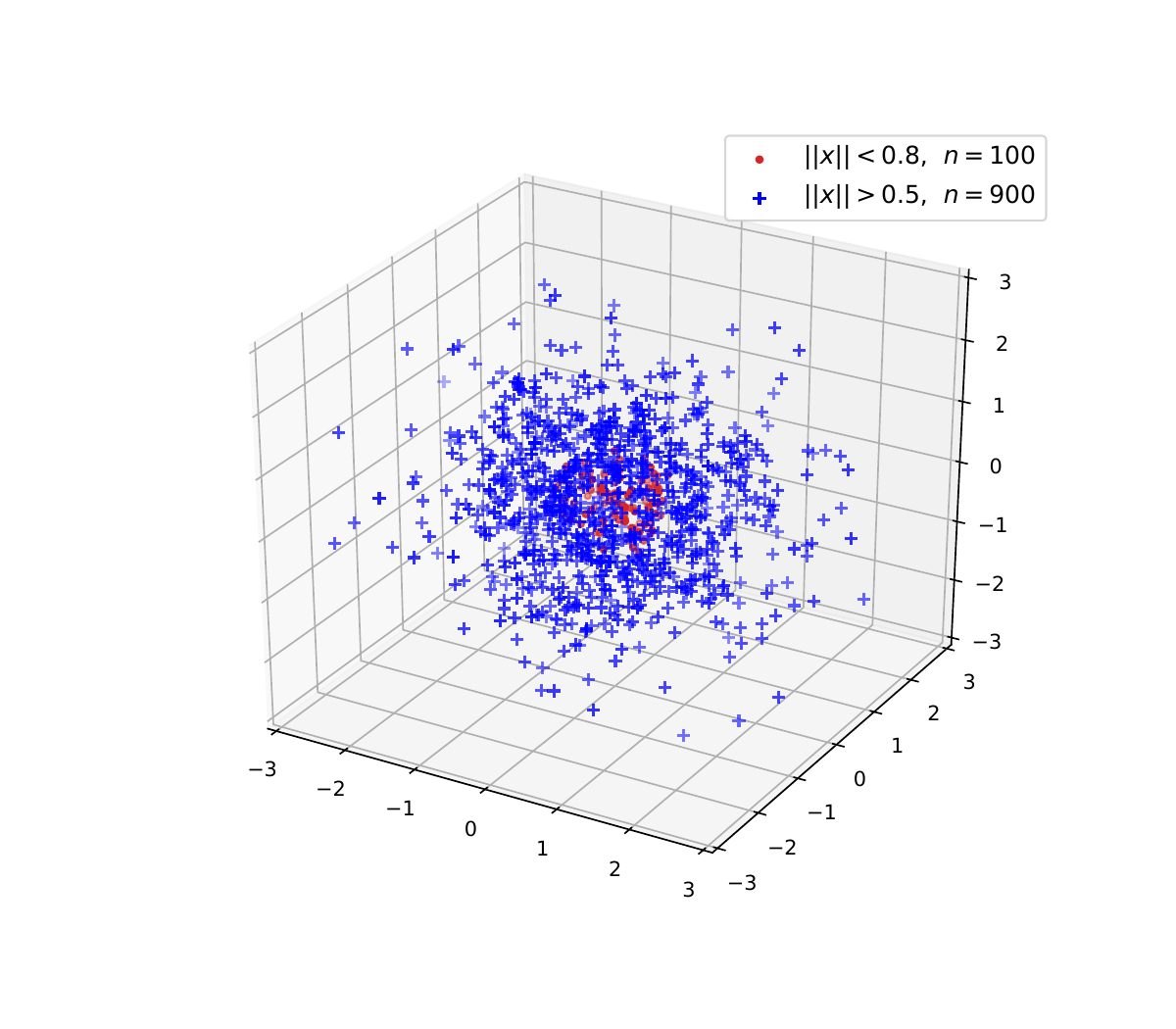}}
\subfigure[Scenario f)\label{fig:sc_f}]{\includegraphics[width=0.45\textwidth]{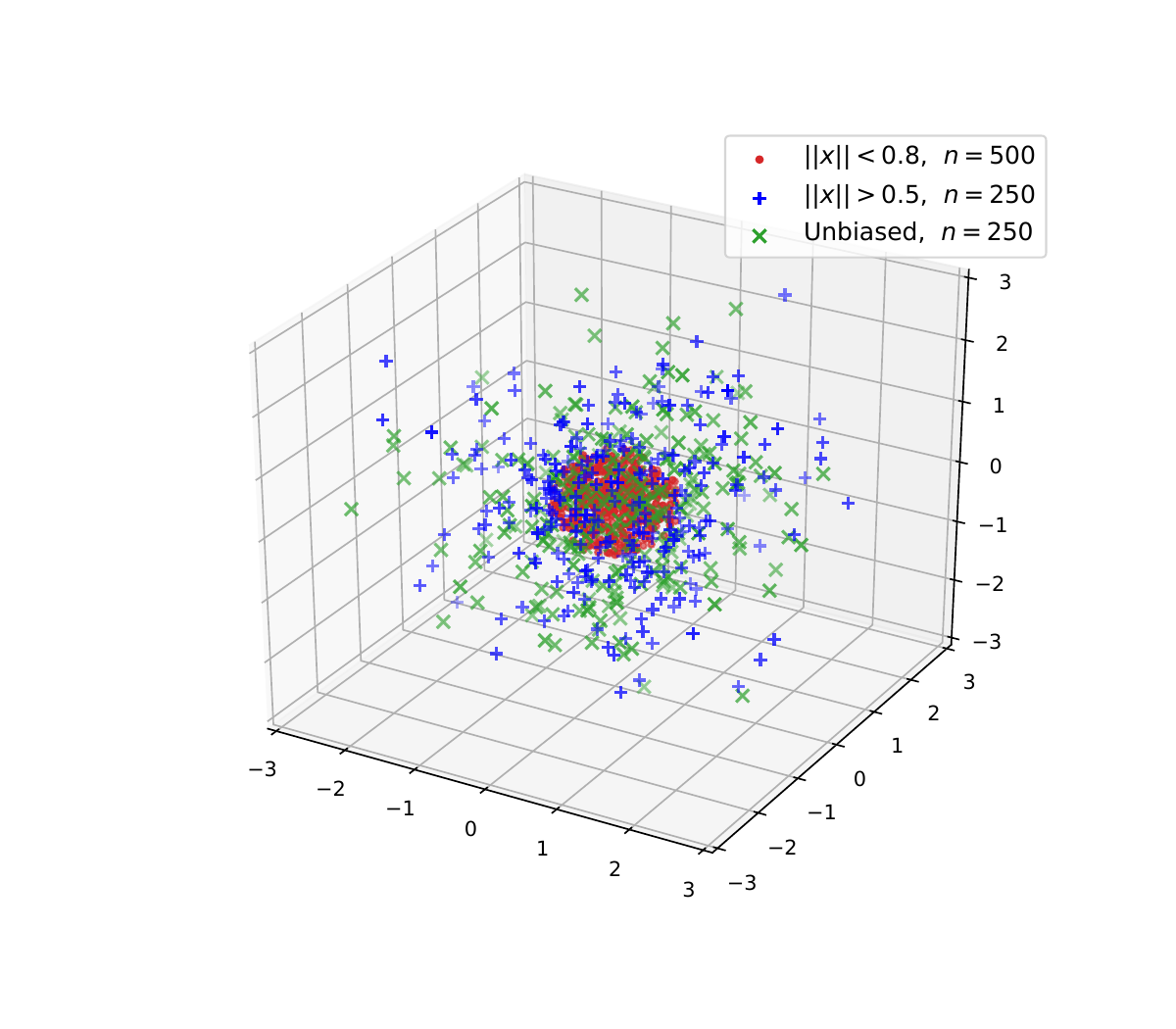}}\\
\caption{Different scenarios when selection bias applies to the vector's norm}
\label{fig:norm_scenarios}
\vspace{0.8cm}
\end{figure*}

\begin{table*}[!ht]
\begin{center}
\Rotatebox{90}{
\begin{tabular}{lc|cccc}\toprule
                        &              & LR                             & KRR                            & SVR                            & RF                             \\\midrule
\multirow{2}{*}{a)} & ERM          & \textbf{4.6e-1 $\pm$ 4.0e-2} & 6.8e-2 $\pm$ 2.9e-2          & 6.6e-3 $\pm$ 2.7e-3          & \textbf{3.4e-2 $\pm$ 6.7e-3} \\
                        & db-ERM & 4.6e-1 $\pm$ 4.0e-2          & \textbf{6.3e-2 $\pm$ 2.8e-2} & \textbf{6.5e-3 $\pm$ 2.6e-3} & 3.4e-2 $\pm$ 6.6e-3          \\\midrule
                        & ERM          & 1.3e+0 $\pm$ 9.8e-2          & 3.2e-1 $\pm$ 7.5e-2          & 3.8e-2 $\pm$ 1.2e-2          & 1.5e-1 $\pm$ 3.2e-2          \\
b)                  & db-ERM & \textbf{4.8e-1 $\pm$ 4.8e-2} & \textbf{1.8e-1 $\pm$ 5.6e-2} & 4.4e-2 $\pm$ 1.3e-2          & \textbf{1.2e-1 $\pm$ 2.8e-2} \\
                        & ub-ERM          & 4.8e-1 $\pm$ 4.9e-2          & 3.4e-1 $\pm$ 7.8e-2          & \textbf{3.0e-2 $\pm$ 9.7e-3} & 1.3e-1 $\pm$ 2.8e-2          \\\midrule
                        & ERM          & 7.2e-1 $\pm$ 6.6e-2          & 1.1e-1 $\pm$ 3.7e-2          & \textbf{1.0e-2 $\pm$ 4.0e-3} & 5.2e-2 $\pm$ 1.1e-2          \\
c)                  & db-ERM & \textbf{4.6e-1 $\pm$ 3.8e-2} & \textbf{7.7e-2 $\pm$ 3.1e-2} & 1.0e-2 $\pm$ 3.7e-3          & \textbf{4.5e-2 $\pm$ 9.0e-3} \\
                        & ub-ERM          & \textbf{4.6e-1 $\pm$ 3.8e-2} & 1.0e-1 $\pm$ 3.7e-2          & 1.1e-2 $\pm$ 4.1e-3          & 4.6e-2 $\pm$ 8.9e-3          \\\midrule
\multirow{2}{*}{d)} & ERM          & 7.0e-1 $\pm$ 6.6e-2          & 1.0e-1 $\pm$ 3.6e-2          & \textbf{9.8e-3 $\pm$ 3.8e-3} & 5.1e-2 $\pm$ 1.0e-2          \\
                        & db-ERM & \textbf{4.6e-1 $\pm$ 3.8e-2} & \textbf{7.5e-2 $\pm$ 3.1e-2} & 9.9e-3 $\pm$ 3.6e-3          & \textbf{4.4e-2 $\pm$ 8.5e-3} \\\midrule
\multirow{2}{*}{e)} & ERM          & 4.6e-1 $\pm$ 4.0e-2          & 6.2e-2 $\pm$ 2.7e-2          & 6.2e-3 $\pm$ 2.5e-3          & 3.4e-2 $\pm$ 6.7e-3          \\
                        & db-ERM & \textbf{4.6e-1 $\pm$ 3.8e-2} & \textbf{6.0e-2 $\pm$ 2.7e-2} & \textbf{6.2e-3 $\pm$ 2.4e-3} & \textbf{3.3e-2 $\pm$ 6.3e-3} \\\midrule
                        & ERM          & 7.1e-1 $\pm$ 6.8e-2          & 1.0e-1 $\pm$ 3.6e-2          & \textbf{9.7e-3 $\pm$ 3.6e-3} & 5.1e-2 $\pm$ 1.1e-2          \\
f)                  & db-ERM & \textbf{4.6e-1 $\pm$ 3.9e-2} & \textbf{7.4e-2 $\pm$ 3.0e-2} & 9.9e-3 $\pm$ 3.4e-3          & \textbf{4.4e-2 $\pm$ 8.8e-3} \\
                        & ub-ERM          & 4.7e-1 $\pm$ 4.1e-2          & 1.7e-1 $\pm$ 5.1e-2          & 1.7e-2 $\pm$ 5.8e-3          & 6.9e-2 $\pm$ 1.5e-2          \\\bottomrule
\end{tabular}
}
\end{center}
\caption{Mean Squared Errors by $4$ Algorithms on the $6$ \textit{Norm Biased} Scenarios.}
\label{tab:complete_norm}
\end{table*}

We now thoroughly describe the first six scenarios, that depict situations where selection bias applies directly to the norm of the realizations, and whose visualizations are available in \Cref{fig:norm_scenarios}.
To understand scenario a), one must have in mind that $1.5$ is approximately the median value of $\|x\|$ when $x \sim \mathcal{N}(\mathbf{0}_3, \mathbf{I}_3)$ (see the $\chi^2(3)$ law).
Hence, partitioning the whole space using $\mathbb{I}\{\|x\| \le 1.6\}$ and $\mathbb{I}\{\|x\| \ge 1.4\}$ (the two subspaces must intersect) divides $\mathbb{R}^3$ into parts of roughly equal importance.
Considering two samples of equal size, each associated to one of these biasing functions, should therefore be almost equivalent to considering blindly the concatenated sample.
Consequently, debiasing ERM in this scenario should not lead to any particular improvement, what is verified empirically.
As no subset is the full space, no third line is provided.
On the contrary, if the samples were of different sizes, one should expect an improvement when using debiasing ERM.
In order to emphasize this effect, scenario b) considers even strongly concentrated points around $0$, with $\mathbb{I}\{\|x\| \le 0.8\}$.
A sample of size $900$ is drawn from this part of the space, which usually represents $10\%$ of the distribution, while a $100$ long unbiased sample completes the scenario.
As expected, the debiasing ERM appears to be less fooled by the outnumbered examples with small norm, and induces a significant improvement compared to the naive ERM.
Furthermore, ERM based the sole unbiased sample is also globally outperformed.
Scenario c) is similar to scenario b), with less imbalanced samples.
Debiasing ERM remains the most successful approach, but by expected lower margins.
What happens if one attempts to fight the selection bias towards $\mathbf{0}_3$ by considering a second sample biased towards great norms, rather than an unbiased one ?
It is the purpose of scenarios d) and e) to investigate this option, using $\mathbb{I}\{\|x\| \ge 0.5\}$ as a second biasing function.
Almost no change can be acknowledged when the sample sizes are the same as in scenario c) (see scenario d)).
However, the advantage of debiasing ERM decreases with the proportion of small norm points, as illustrated by scenario e).
Finally, scenario f) illustrates that the  number of samples is of low importance.
If the sample biased towards small norms is large enough, debiasing ERM outperforms all other methods, even if two additional samples are considered, one biased towards large norms, and one unbiased.
All numerical results can be found in \Cref{tab:complete_norm} and attest that: 1) ignoring selection bias may have dramatic consequences 2) discarding some data and learning only on the unbiased sample -- when it exists -- is not a viable solution either, thus endorsing the debiased approach we promote.\\

\begin{figure*}[!ht]
\centering
\subfigure[Scenario g)\label{fig:sc_g}]{\includegraphics[width=0.45\textwidth]{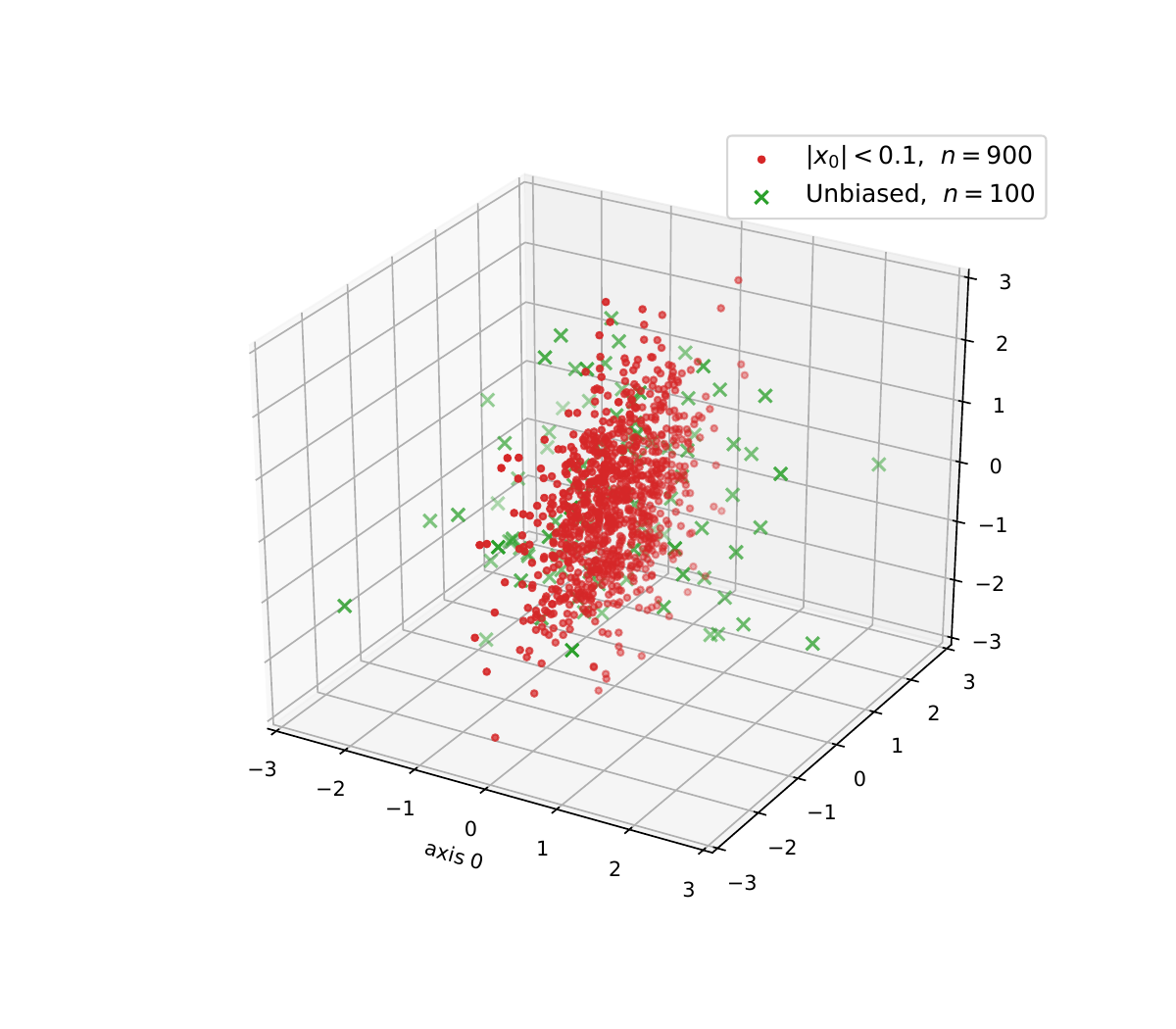}}
\subfigure[Scenario h)\label{fig:sc_h}]{\includegraphics[width=0.45\textwidth]{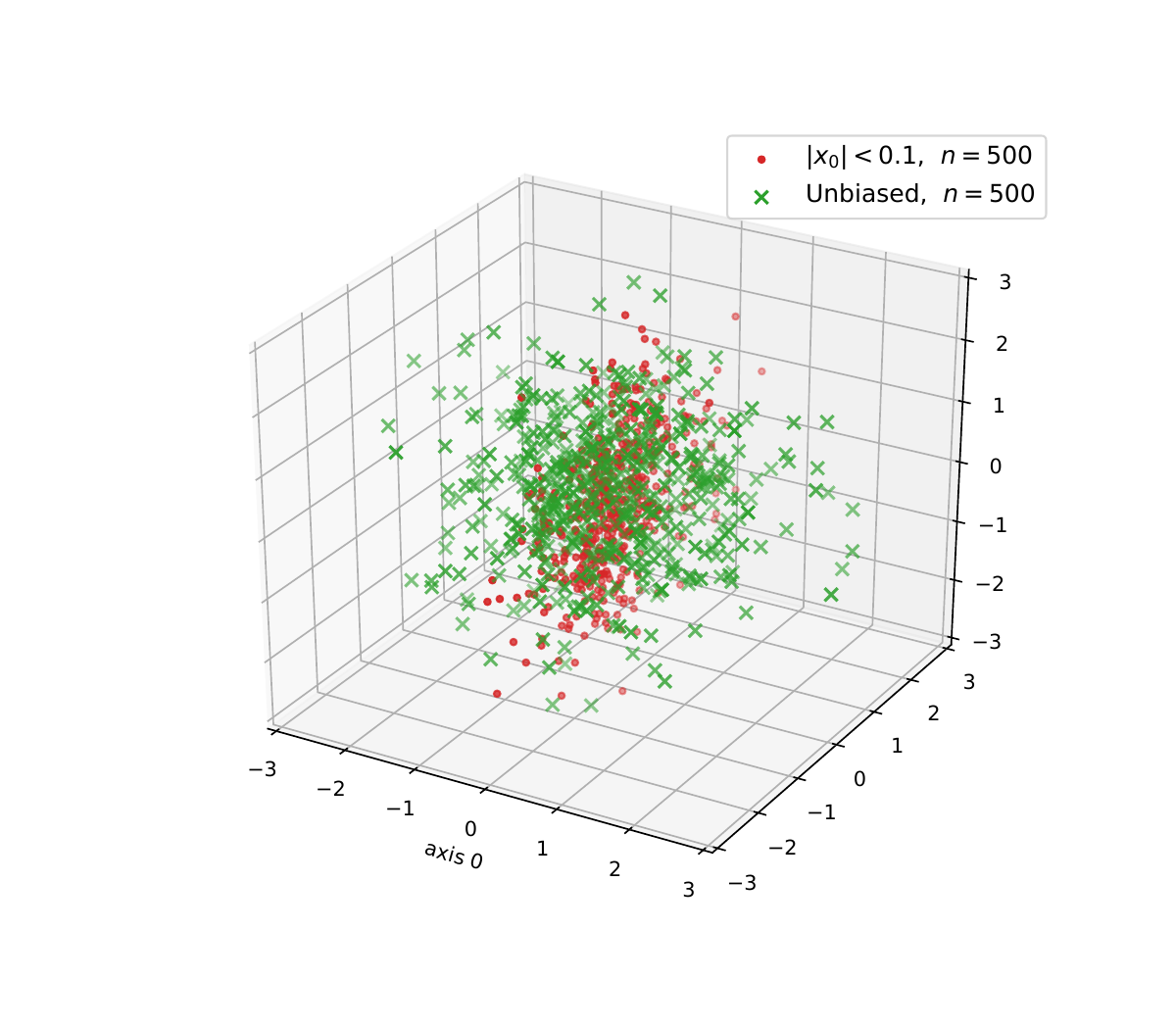}}\\
\subfigure[Scenario i)\label{fig:sc_i}]{\includegraphics[width=0.45\textwidth]{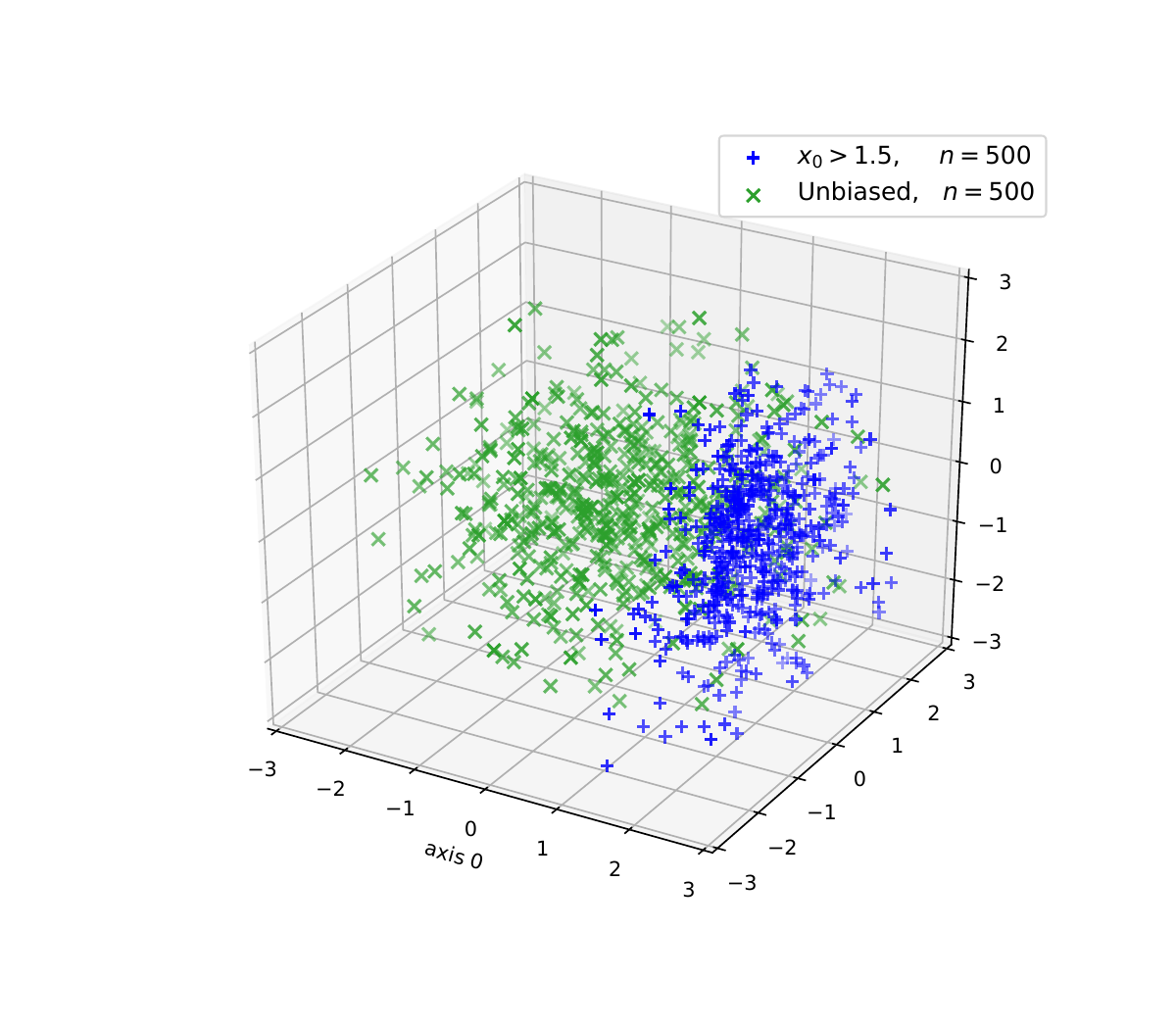}}
\subfigure[Scenario j)\label{fig:sc_j}]{\includegraphics[width=0.45\textwidth]{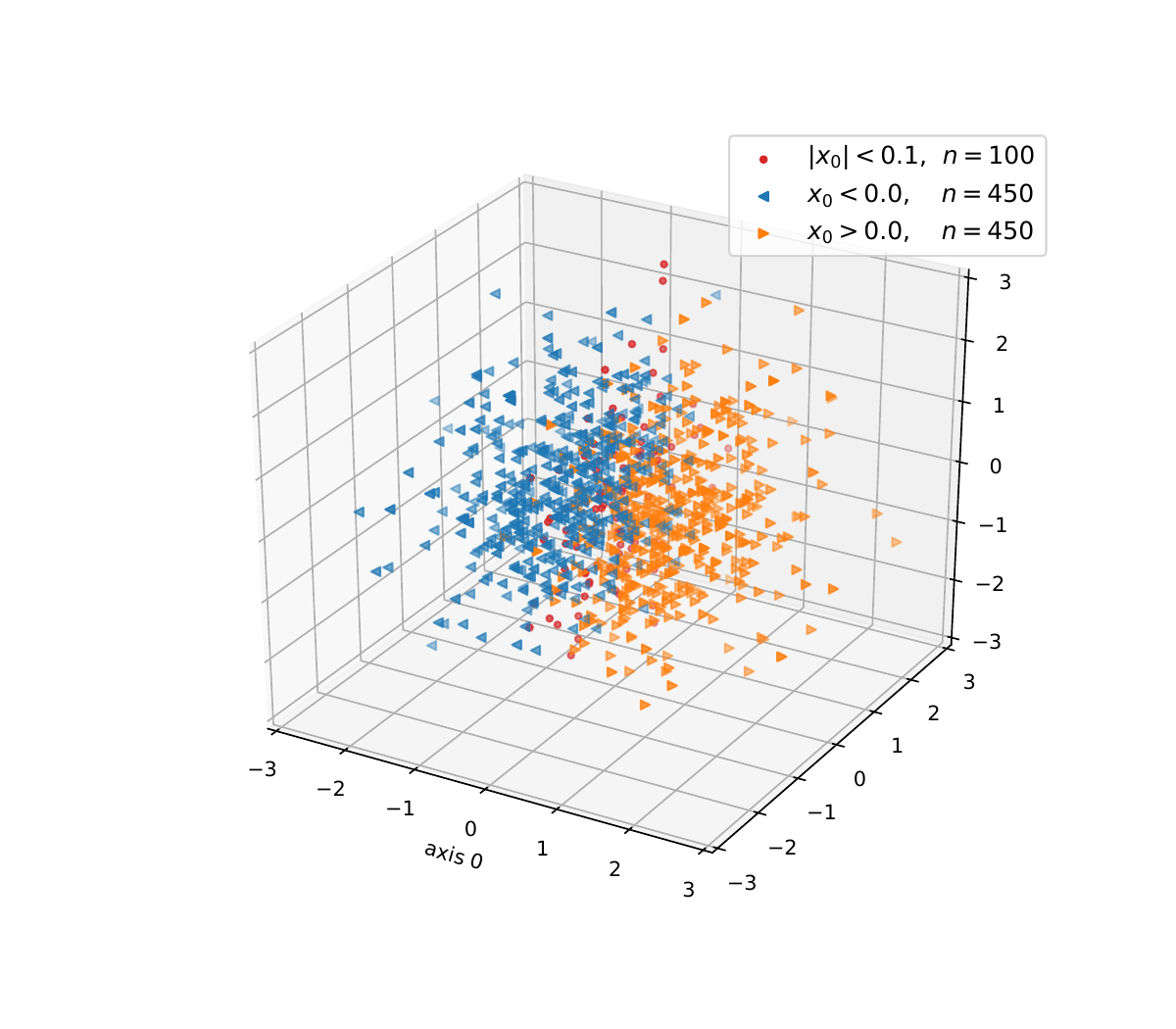}}\\
\subfigure[Scenario k)\label{fig:sc_k}]{\includegraphics[width=0.45\textwidth]{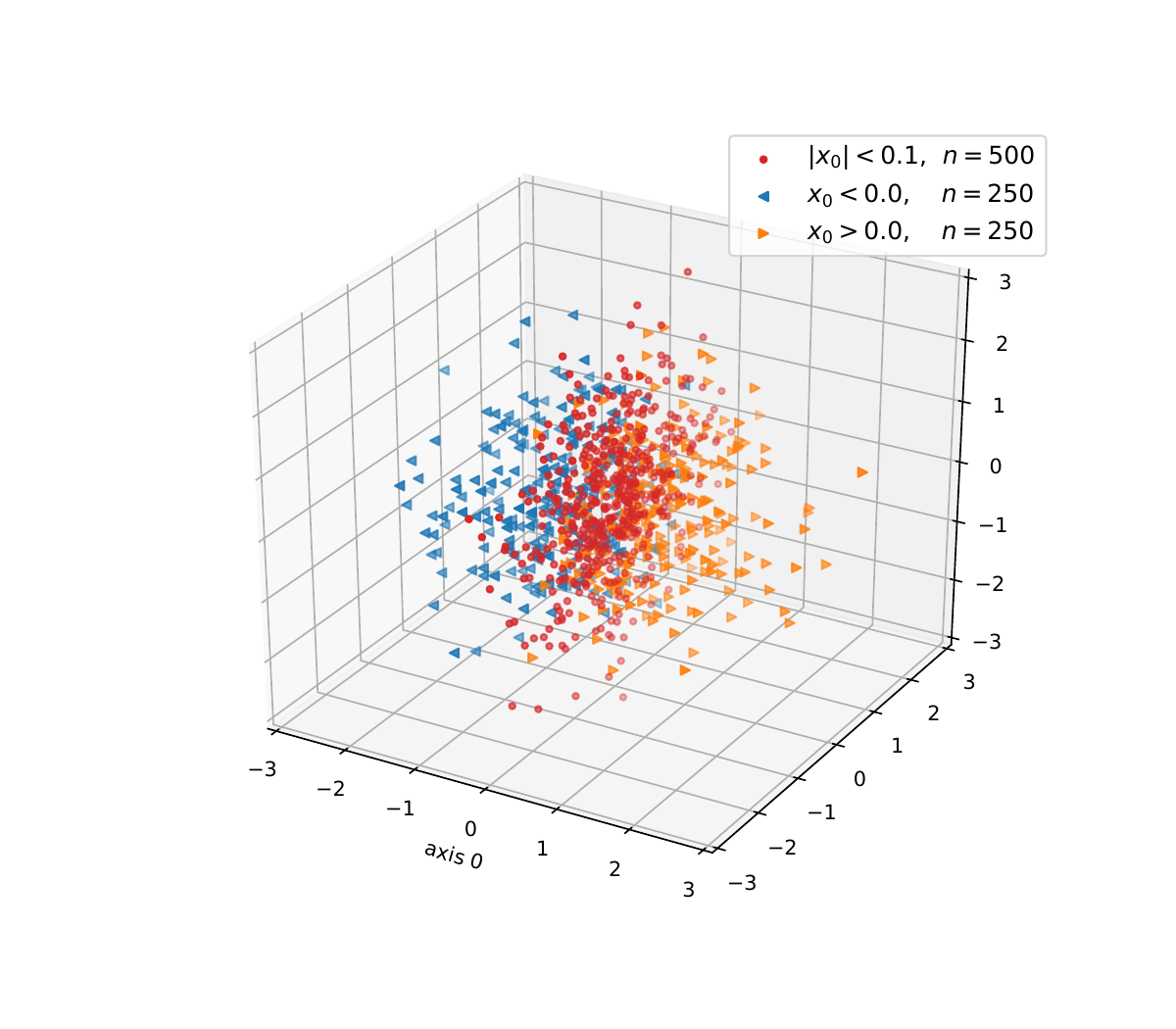}}
\subfigure[Scenario l)\label{fig:sc_l}]{\includegraphics[width=0.45\textwidth]{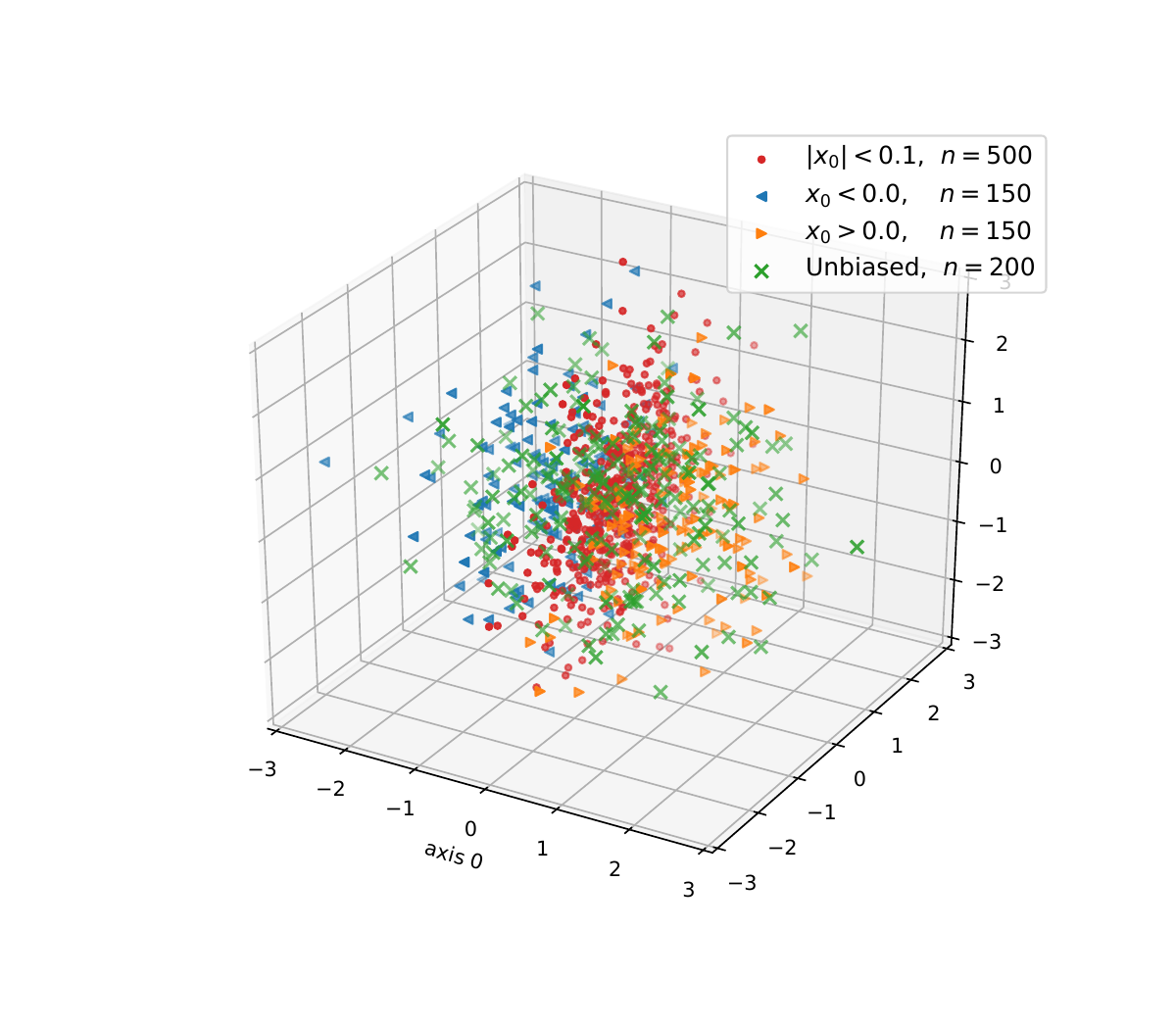}}\\
\caption{Different scenarios when selection bias applies to vector's first dimension}
\label{fig:dim_scenarios}
\vspace{1cm}
\end{figure*}

One may however argue that results presented in \Cref{tab:complete_norm} overestimate the debiasing effect, as bias occur precisely on the problem's target.
We now present similar results obtained when selection bias applies on on one component of the Gaussian only, and not on the norm itself.
Again, six different scenarios have been investigated, and depicted in \Cref{fig:dim_scenarios}, while complete numerical results are gathered in \Cref{tab:complete_dim}.
Scenarios g) and h) are analogous to scenarios b) and c), except that only one component, $x_0$, is now biased towards small values using $\mathbb{I}\{|x_0| < 0.1\}$.
The improvements induced by debiasing ERM remains substantial, and decrease expectedly as the unbiased sample becomes larger (scenario h)).
Scenario i) illustrates that debiasing ERM may improve the results even if a bias applies on large values, using $\mathbb{I}\{x_0 > 1.5\}$ for instance.
However, this bias does not distort the predictions towards small norm values, inducing smaller squared norm errors, hence the smaller benefit of debiasing.
Scenario j) is analogous to scenario a), but with $3$ samples, and leads to similar conclusions: when the blind concatenated sample is very similar to an unbiased sample (the interval $|x_0| < 0.1$ indeed represents $10\%$ of the distribution), debiased ERM is of lower interest.
But when the proportions are not respected anymore, as in scenario k), it significantly increases the performances.
Finally, scenario l) involves $4$ samples, with similar conclusions as above.
Again, and although bias does not apply on the target itself, but rather on one simple covariate, the debiasing approach naturally yields improvements, both upon the standard and the unbiased methods.

\begin{table*}[!ht]
\begin{center}
\small
\Rotatebox{90}{
\begin{tabular}{lc|cccc}\toprule
                        &              & LR                             & KRR                            & SVR                            & RF                             \\\midrule
                        & ERM          & 5.6e-1 $\pm$ 5.7e-2          & 2.0e-1 $\pm$ 5.8e-2          & \textbf{1.5e-2 $\pm$ 5.4e-3} & 1.4e-1 $\pm$ 3.2e-2          \\
Sc. g)                  & db-ERM & \textbf{4.8e-1 $\pm$ 4.5e-2} & \textbf{1.6e-1 $\pm$ 5.3e-2} & 3.8e-2 $\pm$ 1.3e-2          & \textbf{8.6e-2 $\pm$ 2.1e-2} \\
                        & ub-ERM & 4.8e-1 $\pm$ 4.6e-2          & 3.4e-1 $\pm$ 8.1e-2          & 3.0e-2 $\pm$ 1.0e-2          & 1.3e-1 $\pm$ 3.0e-2          \\\midrule
                        & ERM          & 4.9e-1 $\pm$ 4.7e-2          & 8.7e-2 $\pm$ 3.4e-2          & \textbf{8.3e-3 $\pm$ 3.2e-3} & 4.4e-2 $\pm$ 9.1e-3          \\
Sc. h)                  & db-ERM & \textbf{4.6e-1 $\pm$ 4.0e-2} & \textbf{7.6e-2 $\pm$ 3.1e-2} & 1.0e-2 $\pm$ 3.5e-3          & \textbf{4.1e-2 $\pm$ 8.1e-3} \\
                        & ub-ERM & 4.6e-1 $\pm$ 4.0e-2          & 1.0e-1 $\pm$ 3.7e-2          & 1.1e-2 $\pm$ 3.9e-3          & 4.6e-2 $\pm$ 9.3e-3          \\\midrule
                        & ERM          & 5.5e-1 $\pm$ 4.8e-2          & 6.7e-2 $\pm$ 2.9e-2          & \textbf{6.7e-3 $\pm$ 2.3e-3} & 3.9e-2 $\pm$ 7.9e-3          \\
Sc. i)                  & db-ERM & \textbf{4.6e-1 $\pm$ 3.8e-2} & \textbf{6.7e-2 $\pm$ 2.9e-2} & 8.7e-3 $\pm$ 3.0e-3          & \textbf{3.8e-2 $\pm$ 7.8e-3} \\
                        & ub-ERM & 4.6e-1 $\pm$ 3.9e-2          & 1.0e-1 $\pm$ 3.7e-2          & 1.1e-2 $\pm$ 3.9e-3          & 4.6e-2 $\pm$ 9.0e-3          \\\midrule
\multirow{2}{*}{Sc. j)} & ERM          & \textbf{4.6e-1 $\pm$ 4.0e-2} & 6.4e-2 $\pm$ 2.9e-2          & \textbf{6.4e-3 $\pm$ 2.6e-3} & 3.3e-2 $\pm$ 6.9e-3          \\
                        & db-ERM & 4.6e-1 $\pm$ 4.0e-2          & \textbf{6.3e-2 $\pm$ 2.9e-2} & 6.5e-3 $\pm$ 2.6e-3          & \textbf{3.3e-2 $\pm$ 6.8e-3} \\\midrule
\multirow{2}{*}{Sc. k)} & ERM          & 4.9e-1 $\pm$ 4.6e-2          & 8.7e-2 $\pm$ 3.5e-2          & \textbf{8.3e-3 $\pm$ 3.4e-3} & 4.4e-2 $\pm$ 9.2e-3          \\
                        & db-ERM & \textbf{4.6e-1 $\pm$ 4.0e-2} & \textbf{7.6e-2 $\pm$ 3.3e-2} & 1.0e-2 $\pm$ 3.7e-3          & \textbf{4.1e-2 $\pm$ 8.6e-3} \\\midrule
                        & ERM          & 4.9e-1 $\pm$ 4.7e-2          & 8.6e-2 $\pm$ 3.3e-2          & \textbf{8.2e-3 $\pm$ 3.2e-3} & 4.4e-2 $\pm$ 8.8e-3          \\
Sc. l)                  & db-ERM & \textbf{4.6e-1 $\pm$ 4.0e-2} & \textbf{7.5e-2 $\pm$ 3.1e-2} & 9.9e-3 $\pm$ 3.5e-3          & \textbf{4.1e-2 $\pm$ 8.3e-3} \\
                        & ub-ERM & 4.7e-1 $\pm$ 4.2e-2          & 2.0e-1 $\pm$ 5.8e-2          & 2.0e-2 $\pm$ 7.0e-3          & 8.1e-2 $\pm$ 1.7e-2          \\\bottomrule
\end{tabular}}
\end{center}
\caption{Mean Squared Errors by $4$ Algorithms on the $6$ \textit{First Component Biased} Scenarios}
\label{tab:complete_dim}
\end{table*}

% IN THE CORE TEXT

% \subsection{Experiments on the \emph{Boston} dataset}
% \label{apx:boston_expes}

% As an illustration of the benefits of our debiasing approach on learning tasks, we consider here the \emph{Boston} housing dataset problem, where the price of a house is to be predicted based on $14$ attributes, such as the number of rooms or neighborhood statistics.
% %
% One may easily conceive that the dataset at disposal is actually composed of two samples: one large open dataset, in which the most expensive houses do not appear for privacy purposes, and a second one, unbiased but smaller, taken e.g., from a local estate agency.
% %
% This setting can be simulated the following way: from the $500$ observations available, $400$ are kept as a first training sample.
% %
% Two samples are then derived from it: a biased one with the cheapest houses of size $250$, and an unbiased one of size $50$.
% %
% Models are trained on the $300$ selected observations, and tested on the other $100$ ones first set aside.
% %
% Numerical results are displayed in \Cref{tab:boston} in terms of Mean Squared Errors (MSEs), validating the soundness of the debiasing approach.

\subsection{Second Experiments on the \emph{Adult} dataset}
\label{apx:adult_expes}

\begin{figure*}[!t]
\begin{center}
\subfigure[Persons earning more than 50k\$ w.r.t. age]{\label{fig:age}\includegraphics[width=0.45\textwidth]{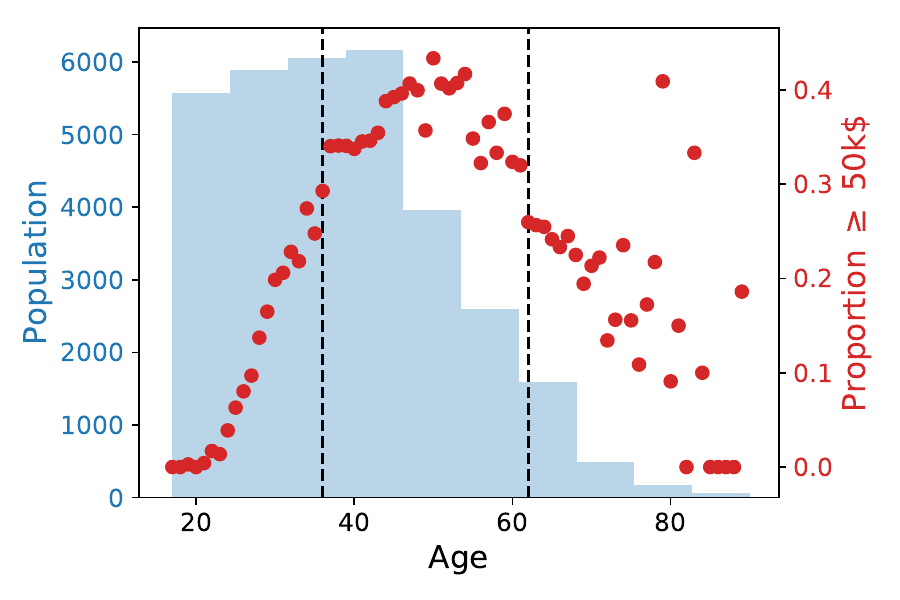}}
\hfill
\subfigure[Persons earning more than 50k\$ w.r.t. education by age group]{\label{fig:edu_by_age}\includegraphics[width=0.45\textwidth]{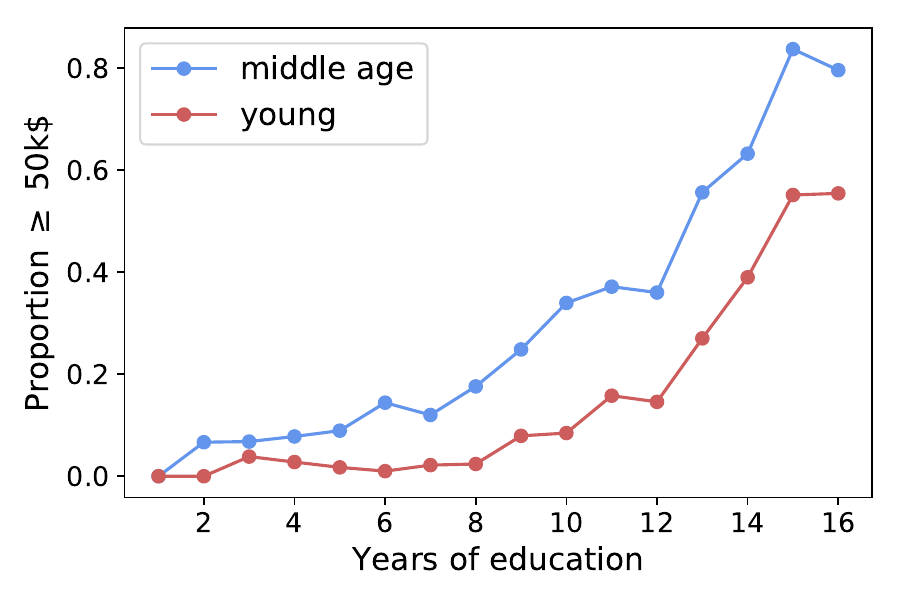}}
\caption{Proportion of people earning more than 50k with respect to age (left),
and with respect to years of education by age group (right)}
\end{center}
\end{figure*}

In this subsection, we present another experiment on the \emph{Adult} dataset showing the benefit of the debiasing approach we promote.
Following a similar reasoning as that of \Cref{sec:num}, first notice that the age of the subject has a strong impact on his/her probability to earn more than 50k\$ a year (see \Cref{fig:age}).
Moreover, and as for the example based on years of education, this scenario cannot be cast as a Covariate Shift problem.
Indeed, the conditional laws cannot be assumed to remain identical.
\Cref{fig:edu_by_age} illustrates this phenomenon by showing the dependence of the income with respect to the years of education by age group.
Clearly, middle age people take more advantage of their education than younger people, which is totally normal as they are working for a longer period.
This observation makes simple covariate shift impossible to consider here.
If middle age people happen to be over-represented in the training dataset, it should induce a general over-estimation of the probability, unless our general debiasing procedure is used.
This setting has been simulated as follows.
From the initial observations, 5 000 are kept for the testing phase.
From the rest are sampled two subgroups: one of middle age people of size 9 900, and one unbiased (i.e., sampled from the entire population) of size 100.
A Logistic Regression (LogReg) and a Random Forest (RF) are then trained on the concatenation of the 10 000 observations, with and without debiasing procedure, as well as on the small second sample of size 100 only.
Numerical results are summarized in \Cref{tab:age} in terms of prediction errors.
Again, the debiased version of the ERM yields the best performances, and for both algorithms.
The gaps are however less spectacular than that presented in \Cref{sec:num}.
It is probably due to a softer biasing effect than the one achieved when it applies to the years of education.
The less striking difference between conditional laws (\Cref{fig:expes} and \Cref{fig:edu_by_age}) is another marker that the debiasing effect expected in this latter example is less important.

\begin{table*}[!t]
\begin{center}
\begin{tabular}{c|cc}\toprule
                      & LogReg                    & RF                        \\\midrule
Standard ERM          & 21.26 $\pm$ 1.24          & 16.48 $\pm$ 0.52          \\
\textbf{Debiased ERM} & \textbf{19.10 $\pm$ 1.09} & \textbf{15.91 $\pm$ 0.62} \\
Unbiased Sample       & 22.04 $\pm$ 1.96          & 19.54 $\pm$ 1.17          \\\bottomrule
\end{tabular}
\end{center}
\caption{Prediction errors on the \emph{Adult} dataset, bias on age, averaged over 100 runs.}
\label{tab:age}
\end{table*}

%%%%%%%%%%%%%%%%%%%%%%%%%%%%%%%%%%%%%%%%%
%                                       %
%     SUITE ARITHMETICO-GEOMETRIQUE     %
%                                       %
%%%%%%%%%%%%%%%%%%%%%%%%%%%%%%%%%%%%%%%%%

\iffalse
Define the following sequence
%
$$x_k \le \lambda(x_{k-1} + b).$$
%
%
It can be shown by induction that
%
$$x_k \le \lambda^kx_0 + \left(\sum_{t=1}^k\lambda^t\right)b = \lambda^kx_0 + \left(\frac{\lambda^{k+1} - 1}{\lambda - 1} - 1\right)b \le \lambda^kx_0 + \frac{\lambda^{k+1} - 1}{\lambda - 1}b.$$
%
In our case $x_0 = 0$, $\lambda = 2/(\underline{\lambda}\kappa \varepsilon)$, $b = \log(K) + \log(1/\varepsilon)$, and we apply the inequality $K-1$ times, so we get:
%
$$\forall k \le K, \qquad \hat{u}_{n, k} \le \frac{\left(\frac{2}{\underline{\lambda}\kappa \varepsilon}\right)^K - 1}{\frac{2}{\underline{\lambda}\kappa \varepsilon} - 1}\big(\log(K) + \log(1/\varepsilon)\big).$$
\fi

\bibliographystyle{abbrv}
\bibliography{Ref_Ranking}

\end{document}